%% file: mhrc.tex
\PassOptionsToPackage{table}{xcolor}
\documentclass[single-column]{fairmeta}

\usepackage{amsmath,amsfonts,amssymb}
\usepackage{array}
\usepackage{textcomp}
\usepackage{stfloats}
\usepackage{url}
\usepackage{verbatim}
\usepackage{enumitem}
\usepackage{longtable}
\usepackage{makecell}
\usepackage{algorithm}
\usepackage{algpseudocode}
\usepackage{xspace}
\usepackage{listings}
\usepackage{tabularx}
\usepackage{ragged2e}
\usepackage{wrapfig}
\newcommand{\fitintolinewidth}[1]{%
  \resizebox{\ifdim\width>\linewidth\linewidth\else\width\fi}{!}{#1}%
}

\newtcolorbox{PromptBlock}{
  enhanced,
  breakable,                % ⭐ 允许跨页/跨栏断开
  colback=bg-gray,          % 背景色
  colframe=light-gray,      % 边框色
  boxrule=0.5pt,            % 边框线宽
  arc=2pt,                  % 圆角
  left=6pt, right=6pt, top=6pt, bottom=6pt,
  before skip=6pt, after skip=6pt,
  segmentation style={solid, dashed, draw=light-gray},    % ← 让分段的上下边框连续
}

\newcommand{\promptblock}[1]{%
  \begin{PromptBlock}\small
  \textcolor{black}{#1}
  \end{PromptBlock}%
}

\newcommand{\commenttext}[1]{\textcolor{light-gray}{#1}}

\definecolor{dark-gray}{HTML}{A9A9A9} % 
\definecolor{light-gray}{HTML}{b7b7b7} %  
\definecolor{light-green}{HTML}{dcf8c6}
\definecolor{light-blue}{HTML}{a5cdfb}
\definecolor{bg-gray}{HTML}{F8F8F8} % light gray 1
	
\definecolor{dark-green}{HTML}{6aa84f} % dark green 1
\definecolor{highlight}{HTML}{cfe2f3} % light blue 3
\definecolor{blue}{HTML}{2B60DE} % light blue 3 
\definecolor{brown}{HTML}{9F8C76}
\definecolor{pink}{HTML}{D88782}
\definecolor{purple}{HTML}{9370DB}
\definecolor{dark-yellow}{HTML}{F6AE2D}
\definecolor{light-yellow}{HTML}{ebdc7d}

\newcommand{\sysname}{\textit{DynaHMRC}\xspace}
\newcommand{\sysnameo}{\textit{DynaHMRC-GPT4o}\xspace}
\newcommand{\sysnameq}{\textit{DynaHMRC-Qwen3-4B}\xspace}

\title{DynaHMRC: Decentralized Heterogeneous Multi-Robot Collaboration for Dynamic Tasks with Large Language Models}

\author[1,*,\dagger]{Wenhao Yu}
\author[1,*]{Yu'ang Xie}
\author[1]{Yifan Duan}
\author[1]{Jie Peng}
\author[2]{Guanting Ye}
\author[2]{Ka-Veng Yuen}
\author[1]{Yanyong Zhang}
\author[1,\ddagger]{Jianmin Ji}

\small{
\affiliation[1]{University of Science and Technology of China (USTC)}
\affiliation[2]{University of Macau (UM)}
}

\footnotesize{
\contribution[*]{Equal contribution}
\contribution[\dagger]{Project leader}
\contribution[\ddagger]{Corresponding author}
}

\abstract{
Large language models (LLMs) offer robots richer task understanding and stronger adaptability, showing substantial potential for coordinating multi-robot systems in long-horizon, heterogeneous tasks. Despite this promise, several key challenges remain underexplored:
(1) Centralized LLM schedulers struggle to scale with increasing team sizes and environmental complexity. Even with extended context windows, models cannot reliably attend to all salient information, and long-context approximations degrade reasoning quality;
(2) Existing task formulations insufficiently consider dynamic settings, whereas robust adaptation to evolving task conditions is essential for real-world deployment;
(3) Domain-specific data scarcity limits specialized reasoning ability, making proprietary general-purpose models inefficient when applied to expert robotic tasks.
To address these limitations, we propose \sysname, a decentralized framework in which each robot operates as a role-aware LLM agent. This design alleviates the single-model context bottleneck and supports flexible collaboration across different heterogeneous team configurations. \sysname organizes collaboration as a four-stage closed-loop process, including self-description, task allocation with leadership bidding, leader election, and reflective execution. Executable robot interfaces support this process.
We further develop a comprehensive benchmark spanning three task families, four dynamic variations, and six team configurations to systematically study dynamic task modeling. Additionally, we conduct an empirical analysis to guide the construction of domain-specific expert datasets and fine-tune pretrained LLMs to enhance specialized competence. Experiments show that, compared with strong baselines, \sysname achieves higher success rates with fewer action and communication steps while showing promising scalability trends as team size grows within the evaluated settings.
}

\correspondence{Jianmin Ji at \email{jianmin@ustc.edu.cn}; Wenhao Yu at \email{wenhaoyu@mail.ustc.edu.cn}.}
\metadata[Keywords]{Decentralized Heterogeneous Multi-Robot Collaboration, LLMs, and Dynamic Task Conditions}
\metadata[Project Page]{\url{https://dynahmrc.github.io}}

\begin{document}

\maketitle

\input{chapters/1.introduction}
\input{chapters/2.relatedwork}
\input{chapters/3.method}
\input{chapters/4.experiment}

\input{chapters/5.conclusion}

% \balance

% Biography text here without a photo.

% \begin{IEEEbiography}[{\includegraphics[width=1in,height=1.25in,clip,keepaspectratio]{fig1.png}}]{IEEE Publications Technology Team}
% In this paragraph you can place your educational, professional background and research and other interests.\end{IEEEbiography}

\bibliographystyle{assets/plainnat}
\bibliography{reference}

\clearpage
\beginappendix
\input{chapters/6.appendix}

\end{document}

%% file: chapters/1.introduction.tex
\section{Introduction}
As robotic systems transition from controlled laboratory settings to real-world deployment, they increasingly face dynamic, communication-constrained environments that demand adaptability, robustness, and generalization.
Consider a disaster-response scenario in which a mobile manipulator navigates damaged indoor areas to retrieve objects or clear obstacles, a drone inspects elevated or obstructed regions from the air, and a ground robot flexibly explores accessible areas to gather local spatial information and support team collaboration. In such settings, communication bandwidth may be limited, and centralized cloud connectivity may be intermittent or unavailable. What's more, task conditions may evolve unexpectedly as goals change, access becomes restricted, and team members fail or join during execution. These characteristics make heterogeneous multi-robot collaboration both essential and highly challenging.

Rule-based pipelines and tightly coupled models often fail to scale under uncertain goals, diverse constraints, and evolving teams. In this context, LLMs present a transformative opportunity. Their rich semantic understanding, strong logical reasoning, and powerful generalization capabilities~\cite{wei2023cot} allow robots to comprehend high-level task intent and infer implicit constraints. These capabilities also help robots adapt to novel tasks, unfamiliar environments, and previously unseen teammates with minimal prior knowledge.

In heterogeneous multi-robot systems, individual robots are specialized for manipulation, navigation, or perception. Collaborative execution is therefore essential for solving complex, multi-stage tasks. While heterogeneity brings complementary capabilities, it also introduces challenges in collaboration, role allocation, and behavior synchronization, especially when subtasks are temporally or causally interdependent~\cite{gerkey2004formal}. Centralized planning methods can provide globally optimized solutions. However, they often struggle to scale and become brittle under dynamic task conditions, agent failures, or the unexpected addition of new team members~\cite{choi2009consensus, lagoudakis2005auction}.

\begin{wrapfigure}{r}{0.5\textwidth}
    \centering
    \includegraphics[width=\linewidth]{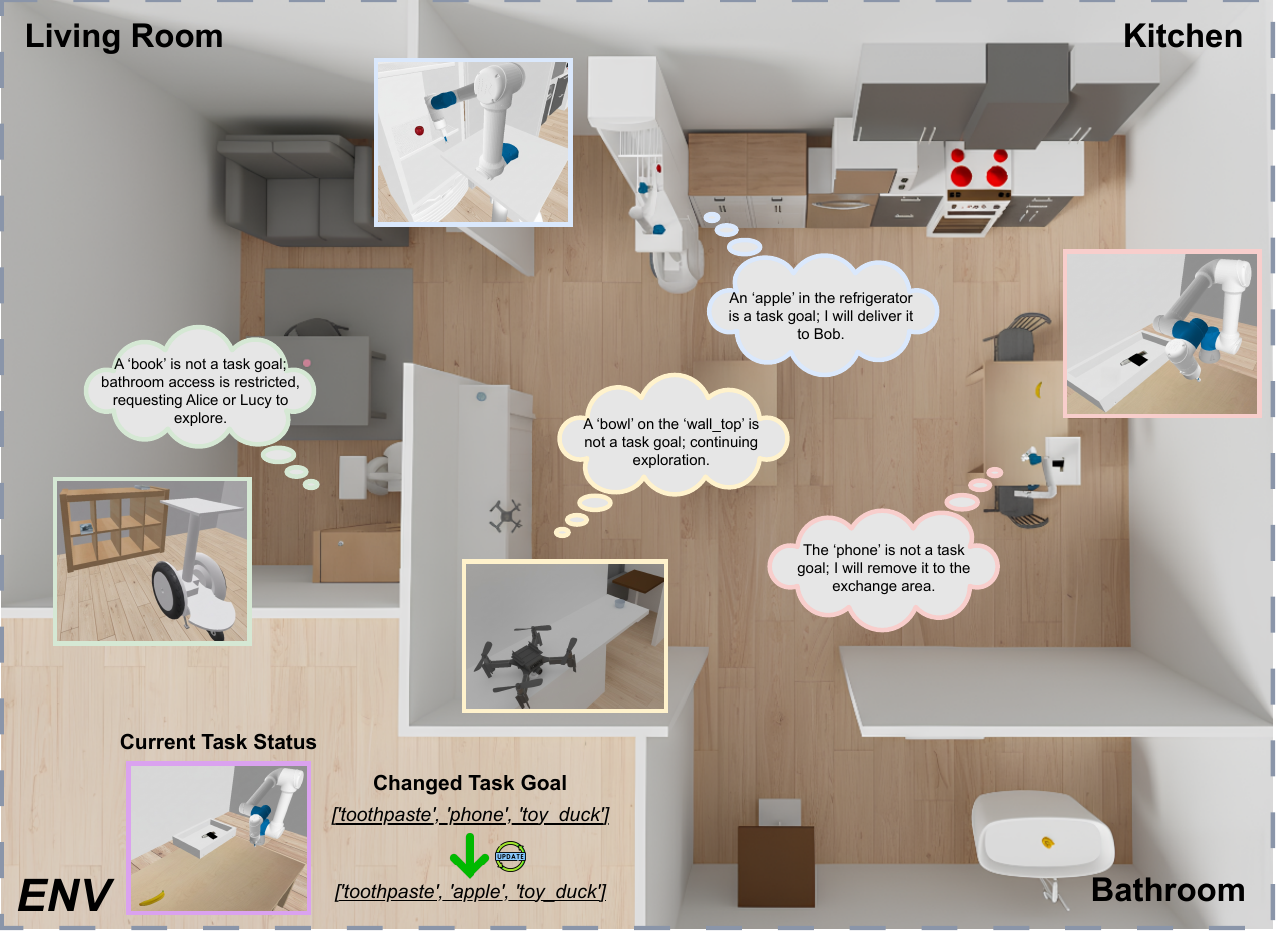}
    \caption{The figure illustrates a heterogeneous multi-robot collaboration scenario under evolving task conditions. At this point, the task objective has shifted: the original target, a phone, has been replaced by an apple. Each robot updates its behavior accordingly. The manipulation robot (Bob) removes the phone from the tray because it is no longer relevant. The mobile manipulation robot (Alice) locates an apple in the refrigerator, which is the new target, and should transport it to the table. The drone robot (Lucy), upon reaching the wall\_top area, identifies a bowl that is not a target and must continue searching. The mobile robot (David) discovers a book on the bookshelf, also a non-target, and, being unable to enter the bathroom, requests assistance from Alice or Lucy to proceed with the exploration.}
    \label{fig:first}
\end{wrapfigure}

% Recent efforts leveraging LLMs for heterogeneous multi-robot collaboration have shown promising potential, but they typically focus on static tasks with fixed constraints. In contrast, real-world deployments require the ability to handle evolving task goals, shifting environmental conditions, and dynamically changing team compositions.

Recent LLM-robotics pipelines have demonstrated impressive single-robot planning and skill composition, primarily under largely static assumptions~\cite{ahn2022can, 10160591, huang2022innermonologueembodiedreasoning}. 
In heterogeneous multi-robot systems, LLM-based frameworks have demonstrated the capability to coordinate robots with diverse physical embodiments. However, most existing frameworks rely on centralized allocators that must aggregate global information from all robots, substantially increasing the LLM's input context length. This often causes the model to lose focus on salient information, leading to performance degradation as the number of robots increases~\cite{liu2023lost}. Moreover, such centralized approaches are typically evaluated on static benchmarks and lack adaptability to dynamic variations, such as shifting task goals, restricted zones, or changing teammates~\cite{liu2025coherent, mandi2024roco, chen2025emos, chen2024scalable, yu2024mhrc, wan2025embodiedagent}.

To bridge this gap, we propose a decentralized framework for heterogeneous multi-robot collaboration that harnesses the semantic and reasoning capabilities of LLMs to enable flexible, robust, and generalizable collaboration. Our framework is designed to accommodate varying numbers and types of robots and dynamically adapts to unknown object locations, changing task requirements, and evolving environmental constraints. 
Inspired by Kolb’s experiential learning theory~\cite{kolb2014experiential}, our system follows a similar four-stage closed-loop process. This structure supports decentralized negotiation, planning, and reflection in heterogeneous multi-robot collaboration through four stages: (1) self-description stage, (2) task allocation and leadership bidding stage, (3) leader election stage, and (4) closed-loop execution \& self-reflection stage. 
This architecture employs periodic group discussions to achieve lightweight global consistency without compromising decentralization. These mechanisms help to synchronize task execution with environmental dynamics, thus improving collaboration efficiency and system resilience. We model the collaborative task as a decentralized partially observable Markov decision process (DEC-POMDP)~\cite{bernstein2002complexity, amato2013decentralized, oliehoek2016concise}. Under this formulation, each robot makes decisions based on its local observations and communications with other robots. 
We use role-aware prompts with in-context learning to improve long-horizon reasoning while respecting bounded context windows.

To systematically study heterogeneous multi-robot collaboration under dynamic task conditions, we evaluate the proposed framework on a benchmark with diverse tasks, dynamic variations, and team configurations, and further validate it in real-world experiments.
The main contributions of our work are summarized as follows:
\begin{itemize}
\item We introduce \sysname, a novel framework that leverages LLMs to enable decentralized collaboration among heterogeneous multi-robot systems. The framework constrains the LLM context for fine-grained attention, with low-frequency group discussions providing lightweight global consistency and mid-horizon correction.
\item The framework is designed to support flexible heterogeneous team compositions, generalize to unseen object locations and novel constraints with minimal task-specific engineering, and maintain robust collaboration across temporally and causally interdependent subtasks amid diverse dynamic variations such as changing task goals, team compositions, no-go zones, and action constraints.
\item We establish a benchmark encompassing three representative tasks, four types of execution-time dynamic variations, and six heterogeneous team configurations. This benchmark is coupled with a unified evaluation protocol and an empirical study that surfaces failure modes and guides expert data collection.
\item Extensive simulations and real-robot experiments show that our approach consistently achieves higher success rates with fewer steps compared to baselines, demonstrating its efficiency, robustness, and practical applicability in dynamic and unstructured environments.
\end{itemize}

%% file: chapters/2.relatedwork.tex
\section{Related work}
\subsection{Semantic Instruction Grounding for Robotics Planning and Control}

Recent efforts have integrated LLMs into robotic task and motion planning by leveraging their rich semantic knowledge and strong reasoning capabilities. This integration has demonstrated notable improvements in robotic manipulation, autonomous navigation, and human–robot interaction~\cite{jansen2020visuallygroundedplanningvisionlanguage,sharma2022skillinductionplanninglatent,huang2022language,li2022pre,ahn2022can,zeng2022socraticmodelscomposingzeroshot}.
LLMs are increasingly employed to interpret high-level instructions, perform hierarchical task decomposition and planning, and enable end-to-end generation of low-level actions~\cite{huang2022language,ahn2022can,10161317,10160591}.
SayCan~\cite{ahn2022can} combines pre-trained skills with value functions to select appropriate actions for embodied tasks. Code as Policies (CaP)~\cite{10160591} uses structured code generation for robot control, while VoxPoser~\cite{huang2023voxposercomposable3dvalue} employs 3D Value Maps and LLM-generated code for manipulation tasks, with re-planning capabilities in case of failure. The methods discussed above highlight the substantial potential of LLMs in advancing robot task planning. 

Furthermore, integrating LLMs with other modality encoders, such as those for images and point clouds, opens up new opportunities for advancing robotic task execution. RoboBrain2.0~\cite{team2025robobrain} develops a multimodal embodied foundation model that emphasizes perception-driven cognition for question answering, enabling robots to integrate vision and language for reasoning about affordances, spatial relations, and interactive tasks. Video foundation models, such as Veo~3~\cite{wiedemer2025video}, demonstrate strong zero-shot capabilities in segmentation, affordance recognition, tool use simulation, and visual reasoning, which provide promising foundations for enhancing robots’ multimodal perception and embodied cognition.

Building on the synergy between LLMs and multimodal perception, recent works have further enhanced robotic autonomy through closed-loop feedback mechanisms and dynamic re-planning strategies. TREE-PLANNER~\cite{hu2024treeplanner} optimizes the closed-loop feedback process through plan sampling, action tree construction, and grounded decision-making. Huang et al.~\cite{huang2022innermonologueembodiedreasoning} presented a general formulation of Inner Monologue that combines different sources of environmental feedback with methods fusing LLM planning with robotic control policies. HiCRISP~\cite{ming2023hicrisp} allows robots to identify and correct errors at each step of the task. The re-planning for closed-loop feedback in COME-robot systems~\cite{zhi2024closed} for mobile robots has been primarily focused on the robotic arm, without fully exploiting the inherent mobility of the robot. This oversight limits the robot's effectiveness in tasks such as grasping and transporting objects. Compared to these methods, \sysname emphasizes decision-making and planning tasks for multiple heterogeneous robots by integrating closed-loop feedback specifically associated with robotic capabilities with LLMs. As a result, \sysname is able to better leverage the complementary strengths of different robots, thereby accomplishing complex tasks that cannot be completed by a single robot alone.

\subsection{Multi-Agent Organization and Heterogeneous Robot Collaboration}

The study of multi-agent systems (MAS) has a rich history~\cite{stone2000multiagent,351231}. MAS can be classified into two types: homogeneous, where agents share similar characteristics, and heterogeneous, where agents have diverse capabilities~\cite{dorri2018multi}.

\subsubsection{Organization and role learning in MARL.}
Beyond LLM-based robotic task execution, organization and role assignment have been widely studied in MARL for cooperative Dec-POMDPs. A representative line of work learns latent or explicit organizational structures to improve coordination and scalability. ROMA~\cite{wang2020roma} learns emergent latent roles through stochastic role embeddings, encouraging agents with similar roles to share behaviors while specializing in different subtasks. RODE~\cite{wang2020rode} further decomposes the joint action space into restricted role action spaces and learns a role selector, forming a bi-level hierarchy that improves scalability and transfer. MAVEN~\cite{mahajan2019maven} introduces a shared latent variable for hierarchical control, enabling temporally extended and committed exploration for long-horizon coordination. More closely related to our leader mechanism, SOG~\cite{shao2022self} proposes self-organized grouping with conductor election and message summary to improve generalization under dynamic team composition and partial observability.

These methods are structurally related to our role allocation and leader election mechanisms. However, they typically operate within the MARL policy-learning paradigm, requiring predefined state/action spaces, reward functions, and extensive environment interaction or centralized training. In contrast, \sysname operates at the semantic task-planning level, using LLMs to reason over natural-language task descriptions, robot self-descriptions, textual feedback, and dynamic task constraints. The roles and leader in \sysname are therefore not latent variables optimized through policy learning, but semantic and embodiment-aware coordination structures generated through LLM reasoning and prompt conditioning. In this sense, our work is complementary to MARL organization methods: it brings role-aware and leader-mediated organization into LLM-based heterogeneous robot collaboration without training a new low-level MARL policy for each task.

\subsubsection{Homogeneous Agents}

Sampling-based methods and their various algorithmic improvements have been proposed as a widely used approach in homogeneous agents~\cite{karaman2011sampling,7354289}. Li et al.~\cite{7902130} tackled the optimal synchronization in homogeneous multi-agent systems using an actor-critic neural network and least squares to approximate the control policy and value function. Zhang et al.~\cite{8458223} use a non-policy reinforcement learning algorithm to compute each agent's optimal control policy. Guo et al.~\cite{guo2024embodied} places greater emphasis on the organizational framework of LLM agents, investigating how to harness the potential of LLMs to develop more effective collaboration strategies. CoELA~\cite{zhang2023building} leverages LLMs to establish a collaborative framework for homogeneous agents, facilitating cooperative exploration, transport, and communication between the two agents. Although CoELA adopts a decentralized architecture, it does not explicitly address heterogeneous embodiment grounding, dynamic task specifications, or global consistency in long-horizon tasks. Consequently, when task goals, access constraints, or team composition change during execution, such designs are prone to communication redundancy and limited adaptability.
\subsubsection{Heterogeneous Agents}

Compared to homogeneous agents, heterogeneous agents present different challenges, including the heterogeneity of agents, limited view of the environment, and the dynamicity of the multi-agent system (MAS) or environment~\cite{1545539}. Several heterogeneous robot combinations have already been proposed, including aerial-ground collaboration, main-picket collaboration, and humanoid-quadruped collaboration~\cite{9341023,9562042,6907527,9340688}. VIKI-R~\cite{kang2025viki} fine-tunes vision–language models with reinforcement learning to frame heterogeneous multi-robot collaboration as a visual–language reasoning task. IRoT-LLM~\cite{liu2024heterogeneous} applied LLMs to ad hoc and original heterogeneous robots to generate reasonable collaboration strategies. Centralized LLM frameworks for heterogeneous robot collaboration include EMOS~\cite{chen2025emos}, which employs embodiment-aware descriptions with LLM-based agents to orchestrate perception, navigation, and manipulation, and COHERENT~\cite{liu2025coherent}, which decomposes and allocates complex tasks for efficient execution. COHERENT relies on aggregating diverse information sources from multiple robots. As the scale of the robot team increases and the LLM input context grows, the overall collaboration process becomes increasingly fragile, which may result in redundant, inconsistent, or infeasible action proposals. This issue is further exacerbated in dynamic task settings, where changes in task conditions amplify these limitations.

By contrast, our work leverages large language models to construct a decentralized framework for heterogeneous multi-robot collaboration. The framework enables efficient closed-loop collaboration through role allocation and reflective reasoning, allowing scalable support for varying numbers of robots within the evaluated settings and flexible adaptation to diverse task variations. Furthermore, to address the lack of domain-specific data and the limited specialization of existing open-source models, we collect task-relevant data on the benchmark and fine-tune open-source models. This enables us to achieve performance that surpasses state-of-the-art (SOTA) closed-source models on domain-specific tasks, while incurring only modest adaptation costs.

%% file: chapters/3.method.tex
\section{Method}

\begin{figure*}[h]
    \centering
    \includegraphics[width=0.95\textwidth, height=0.45\textwidth]{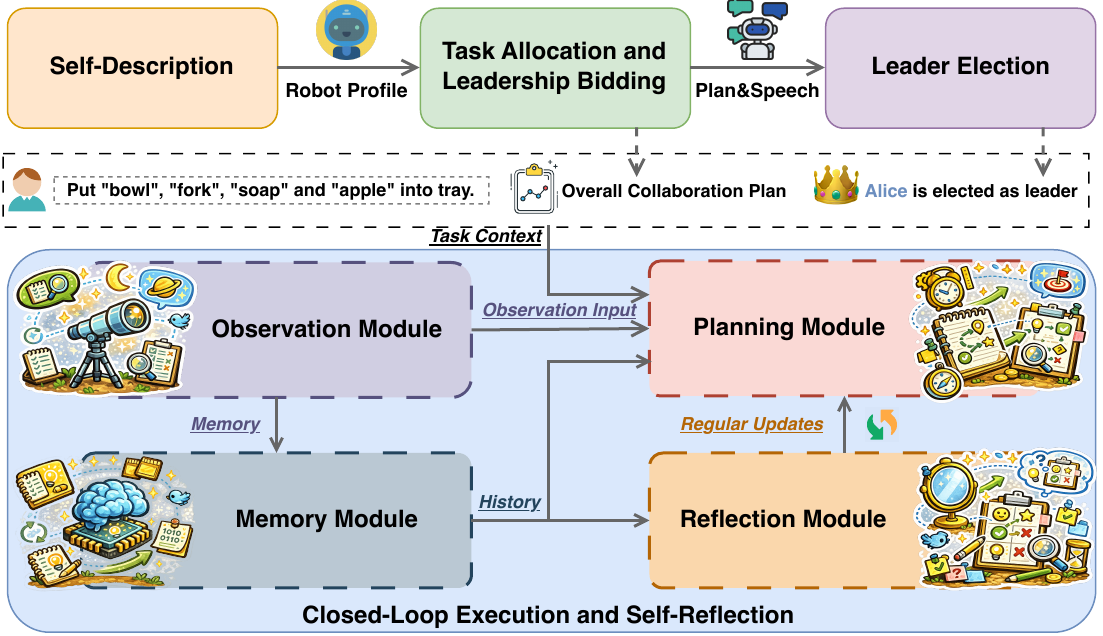}
    \caption{Overview of the proposed decentralized multi-robot collaboration framework. In this system, each robot functions as a role-aware LLM agent equipped with its own perception, memory, and action interfaces. During self-description and task allocation, robots announce their capabilities and bid for leadership; a leader is then elected through distributed voting and proposes a high-level division-of-labor plan. During execution, robots continuously perceive the environment, update their memories, and carry out atomic actions (e.g., navigation, picking, placing, communication) while exchanging status messages. Feedback from action outcomes drives dynamic adaptation, allowing robots to correct errors, resolve conflicts, and adjust their collaboration strategies. Finally, through self-reflection, each robot summarizes its progress and formulates future task plans, completing the iterative loop that enables robust collaboration across heterogeneous mobile, manipulation, and aerial robots. A formal description of the process is provided in Algo.~\ref{algo:collaboration}. It is important to emphasize that, although the overall collaboration loop is organized into sequential phases, decentralization in our framework pertains to the decision-making architecture rather than a fully asynchronous execution schedule. Please refer to App\ref{lab:detailed_arch} for the detailed initialization process and the detailed architecture of the functional modules.}
    \label{fig:overview}
\end{figure*}

\subsection{Preliminary Information}

We formulate this problem as a DEC-POMDP~\cite{bernstein2002complexity} to facilitate collaborative tasks among heterogeneous multi-robots. Specifically, a DEC-POMDP is defined as $(n, \mathcal{S}, \{\mathcal{A}_i\}, \{\mathcal{O}_i\}, \mathcal{P}, \mathcal{T}, \mathcal{R}, \gamma, \mathcal{H})$, where $n$ is the number of heterogeneous robots in the task; $\mathcal{S}$ denotes a finite set of states; $\mathcal{A}_i = A_i^F \times A_i^M$ represents the action set for robot $i$, including a finite action set $ A_i^F$ determined by the robot's individual capabilities and a communication action $A_i^M$ used to exchange messages with other robots; $\mathcal{O}_i = O_i^F \times O_i^M$ signifies the observation set for robot $i$, encompassing both the observation set $O_i^F$ derived from the robot's perceptions and the message set $O_i^M$ received from other robots; 
$\mathcal{P}: \mathcal{S} \times \mathcal{A}_1 \times \dots \times \mathcal{A}_n \times \mathcal{S} \rightarrow [0,1]$ denotes the state transition function, representing the probability $\mathcal{P}(s' \mid s,\mathbf{a})$ of transitioning to state $s'$ given the joint action $ \mathbf{a} = (a_1, \dots, a_n)$;
$\mathcal{T}: \mathcal{S} \times \mathcal{O}_1 \times \dots \times \mathcal{O}_n \rightarrow [0,1]$ denotes the observation probability function, i.e., $\mathcal{T}(\mathbf{o} \mid s')$, which represents the probability of receiving the joint observation $\mathbf{o} = (o_1, \dots, o_n)$ given the underlying state $s'$.
$\mathcal{R}: \mathcal{S} \times \mathcal{A}_1 \times \dots \times \mathcal{A}_n \rightarrow \mathbb{R}$ is the global reward function, which assigns a scalar reward based on the current state and the joint action taken by all robots.
$\gamma$ is the discount factor in $(0, 1)$; $\mathcal{H}$ is a finite planning horizon. 

% It is important to emphasize that our research centers on optimizing the utilization of LLMs for executing heterogeneous multi-robot collaborative tasks, with the goal of enabling generalization across diverse scenarios, varying numbers and configurations of robots, as well as changes in task goals and constraints. This research falls within the domain of robot task and motion planning (TAMP). Consequently, our approach incorporates a description function that translates the semantics and observations related to the robotic tasks into natural language prompts $f_i: \mathcal{O}_i \rightarrow \mathcal{L}, l_i^t = f_i(o_i^t)$, where $\mathcal{L}$ represents the natural language space.

Our research focuses on how to effectively use LLMs to execute collaborative tasks in heterogeneous multi-robot systems. The goal is to enable generalization across diverse scenarios, varying numbers and configurations of robots, as well as changes in task goals and constraints. This research falls within the domain of robot task and motion planning (TAMP). In our implementation, the DEC-POMDP variables are instantiated through textual prompts, feedback, and executable robot interfaces. Each robot's local observation $o_i^t \in O_i$ is translated into natural language by a description function $f_i: O_i \rightarrow L$, $l_i^t=f_i(o_i^t)$, where $L$ denotes the natural language space. The observation includes task-relevant scene information, robot status, task progress, and messages received from other robots. The physical action component $A_i^F$ corresponds to the robot-specific atomic action set, while the communication component $A_i^M$ is implemented by the \texttt{communicate} action. After each action, textual feedback reports the execution outcome, including success, failure causes, constraint violations, and task-progress updates. The feedback is stored together with action and message histories in the bounded memory $h_i^t$. The LLM policy $\pi_i$ then maps the textual observation, memory, feedback, and current leader plan to the next executable atomic action, thereby grounding the formal DEC-POMDP abstraction in the prompt-based planning loop of \sysname.

\input{table/feedback}
\input{table/action}

\begin{figure*}[t]
    \centering
    \includegraphics[width=\textwidth, height=0.17\textheight]{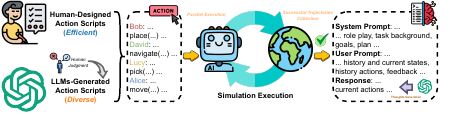}
    \caption{The complete data acquisition pipeline. Action scripts from two distinct sources are first collected, after which simulation execution validates their effectiveness while simultaneously recording observations, feedback, etc. Successful task trajectories are subsequently filtered, and additional reasoning traces are generated with the assistance of \textit{GPT-4o}. A formal description of the process is provided in Algo.~\ref{algo:data_collection}.}
    \label{fig:data}
\end{figure*}

\subsection{Role Assignment in Heterogeneous Robot Teams}
\subsubsection{Self-Description Stage}
At the beginning of the task, the collaborating robots have no prior knowledge about the other participants. Therefore, we introduce a self-description mechanism that allows each robot to describe itself based on its own capabilities, the overall task objective, and the task context. These pieces of information are organized into textual form as part of the prompt design for this stage. App\ref{lab:selfDP} details the prompt design. The output of this process is defined as: 
\begin{equation}
    D=\{d_i = s_i(f_i(o_i^0), \theta_i, \tau_i) \mid i = 1, \dots, n\},
\end{equation}
where $s_i$ denotes the LLM-based self-description function of robot $i$, which depends on its initial observation $o_i^0$, capability configuration $\theta_i$, and the task context $\tau_i$.

This self-description mechanism can adapt to different numbers and compositions of heterogeneous robots. It is crucial for enabling mutual awareness among robots and for facilitating more effective collaboration of task roles.
\subsubsection{Task Allocation and Leader Election Stage}
After completing the self-description phase, the robots enter a leaderless group discussion stage. In this stage, each participating robot is required to accomplish two tasks:
\begin{enumerate}[label=\roman*.]
    \item After fully understanding the information shared by other collaborators, each robot proposes a task allocation plan aimed at achieving the overall task objective  $P$ and delivers a leader campaign speech $G$ that highlights its unique strengths,
    \begin{equation}
        (P, G) = \{(p_i, g_i) = q_i(d_i, \tau_i) \mid i = 1, \dots, n\}.
    \end{equation}
    \item Each robot analyzes the proposed task allocation plans and campaign speeches of others carefully, and conducts a fair and impartial election to select the leader,
    \begin{equation}
        E = \arg\max_{r \in \{1,\dots,n\}} \sum_{i=1}^n v_i(p_r, g_r).
        \label{eq:select}
    \end{equation}
\end{enumerate}

Eq.~\eqref{eq:select} represents the high-level leader selection objective, where each robot evaluates the collaboration plan $p_r$ and campaign speech $g_r$ of candidate $r$. The candidate with the highest overall support is selected as the leader.

In implementation, leader election is carried out iteratively for at most $K_{\mathrm{sel}}$ rounds. At the $k$-th round, each robot outputs one discrete vote $v_i^{(k)}\in\{1,\dots,n\}$, and the vote count received by candidate $r$ is computed as $s_r^{(k)}=\sum_{i=1}^n \mathbf{1}[v_i^{(k)}=r]$. If there exists a unique candidate with the highest vote count, i.e., $|\arg\max_r s_r^{(k)}| = 1$, that candidate is selected as the final leader. Otherwise, the current vote statistics are shared with all robots and appended to the election context for the next round. If a unique leader still does not emerge after $K_{\mathrm{sel}}$ rounds, a referee LLM is invoked as a final arbitration module to select one leader based on the candidate plans and the accumulated voting history. The final elected leader is then broadcast to all robots before task execution begins, ensuring a unique team-wide leader and avoiding split-brain behavior. The leader election is conducted as a one-time warm-up procedure during the task initialization stage. The leader remains fixed throughout the task; periodic discussion updates the high-level division of labor but does not trigger re-election. We further provide additional analysis in App\ref{lab:leader_vote} on the stability of the election process and evaluate the performance of the algorithm under degraded election conditions.

Both $q_i$ and $v_i$ are LLM-based generation functions. After leader election, the elected leader’s proposal $P_E$ serves as shared high-level collaboration guidance, while each robot still generates its executable actions locally based on its own observation, memory, feedback, and received messages.

These two tasks play a vital role in enhancing collaboration performance. A task allocation proposal that fully considers the differences in capabilities among collaborators can help the team complete more complex tasks efficiently, while the introduction of a leadership role within the team can improve collaboration efficiency and reduce redundant actions. App\ref{lab:TALBP} and App\ref{lab:LEP} provide the corresponding prompt designs.

\subsection{Closed-Loop Execution and Self-Reflection in Heterogeneous Robot Teams}
We propose a closed-loop execution and reflection framework based on LLMs to enable collaboration among heterogeneous multi-robot systems. The framework consists of four functional modules: the Observation Module, the Memory Module, the Planning Module, and the Reflection Module.
The overall workflow is as follows: the Observation Module and the Memory Module respectively provide perceptual inputs $o_i^t$ and historical context $h_i^t$. Based on this information, the Planning Module generates the next action in a next-action prediction manner. This action is then executed in the environment, producing new feedback $fb_i^t$. The Planning Module continuously incorporates the updated perception, historical data, and feedback in a closed-loop fashion to plan subsequent actions,
\begin{equation}
    \mathcal{A}^t = \{a_i^t = \pi_i(f_i(o_i^t), h_i^t, fb_i^t, p_E^t) \mid i = 1, \dots, n\}.
\end{equation}

Throughout the planning cycle, the Reflection Module is periodically triggered to summarize and analyze past interactions, enabling updates to future task allocation strategies.
\subsubsection{Heterogeneous Multi-Robot Collaboration Stage}\hfill\\
\indent\textbf{Observation Module.} \label{section:obsmi}

This module encompasses all the state information necessary for robot tasks and motion planning. It mainly includes the following three parts:
\begin{enumerate}[label=\roman*.]
    \item \textbf{\textit{Scene Graph}}: This part presents structured information about the environment. The scene graph contains the positions and orientations of key furniture items. For each piece of furniture, at least one navigation target point is defined for ``\textit{navigate}" action. This setting is similar to~\cite{habitatrearrangechallenge2022, gu2022multi, makhal2018reuleaux}. In addition to this static information, the scene graph incorporates dynamic updates, including the open/closed status of manipulable furniture, newly discovered objects, and changes in their positions and orientations.
    \item \textbf{\textit{Messages Information}}: To achieve effective collaboration in completing tasks, the robots must share information and request assistance as needed. All communication behaviors and message contents are autonomously initiated and structured by the robots themselves, guided through prompts. For example, a manipulation robot (Bob) may request assistance from robots equipped with mobile manipulation capabilities to transport objects on the table that are beyond its operational reach; a mobile manipulation robot (Alice) may receive specialized requests to operate on articulated objects; a drone robot (Lucy) may be asked to explore elevated or hard-to-reach areas; and a mobile robot (David) often proactively shares newly discovered information with other collaborators in a timely manner. The communication paradigm among robots is inherently flexible, allowing for unicast, multicast, or broadcast. Each robot independently selects the appropriate mode based on its operational context.
    \item \textbf{\textit{Robot Information}}: This part includes the different capabilities of the robot and the corresponding status information. For robots equipped with navigation capabilities, the status includes the robot's pose information. For robots with manipulation capabilities, the status encompasses the gripper's open or closed state, the name of the grasped object, and the maximum grasping range.
\end{enumerate}

\indent\textbf{Memory Module.}

In the context of our work, the tasks are designed as long sequences. As the LLMs are queried multiple times, they may lose track of prior tasks, making it essential to retain past decision contexts. The memory module is designed to capture three key components:

\begin{enumerate}[label=\roman*.]
    \item \textbf{\textit{Feedback History} ($fh_i^t$)}: This section primarily stores limited feedback generated from the robot's past interactions with the environment. We have designed various types of textual feedback for our tasks, covering both success and failure cases. For a more comprehensive description of the feedback, please refer to Tab.~\ref{tab:feeb_description}.
    \item \textbf{\textit{Received Message History} ($mh_i^t$)}: This part primarily archives messages previously received by the robot from other robots. %For detailed information regarding message types, please refer to the message information in the section~\ref{section:obsmi}. %and for message formats, please refer to Fig.~\ref{fig:prompts}
    \item \textbf{\textit{Action History} ($ah_i^t$)}: This component primarily archives the sequences of decision-making actions previously executed by the robot. The action types and formats strictly follow the designed atomic action set, as shown in Tab.~\ref{tab:atomic_action}. 
\end{enumerate}

In addition, multiple queues are employed to store different types of historical information, each preserving the latest $K$ steps of its respective interaction history. The historical context is formally defined as follows:
\begin{equation}
    h_i^t = \{(fh_i^{t-k}, mh_i^{t-k}, ah_i^{t-k}) \mid k = K, \dots, 1\}.
\end{equation}

\indent\textbf{Planning Module.}

We employ LLMs to simulate role-playing for various types of robots, guiding task planning through a Chain-of-Thought (CoT) framework. This process involves selecting and executing one action from a predefined list at a time. After each action, feedback is obtained, and based on this feedback, along with historical data, the robot autonomously determines the next subtask or adjusts the plan. This multi-turn interaction enables robots to collaborate effectively in completing long-horizon tasks.

In mobile manipulation, identifying an optimal base position is critical, as it significantly influences the success of grasping actions~\cite{shao2025moma, duan2025stdarm}. To address this, we propose a strategy that combines the ``\textit{navigate}'' and ``\textit{move}'' actions to accomplish the task. For instance, when grasping an apple from a table, the robot first selects a navigation target from the scene graph and moves near the table. It then evaluates the validity of the robotic arm’s initial and target configurations relative to the apple’s position and attempts to grasp the object. If the attempt fails, the robot selects alternative target points closer to the apple, navigates to those positions, and reattempts the action. Should the task remain incomplete, the robot refines its position using the ``\textit{move}'' action, adjusting its base location according to the relative x and y coordinates between the base and the apple. 

Compared to other actions, executing a  ``\textit{move}'' action requires two separate LLM calls. In the first call, the LLM selects the move action. In the second call, we provide the LLM with a robot-centered local costmap $CM_i^t = \{X_{\text{free}}^t, X_{\text{obs}}^t, X_{\text{goal}}^t, X_{\text{base}}^t\}$, which includes annotations for the target object location $X_{\text{goal}}^t$, the robot's current position $X_{\text{base}}^t$, obstacle regions $X_{\text{obs}}^t$, and navigable areas $X_{\text{free}}^t$. The LLM is then asked to select a position $x_i^{\text{next}}$ within the navigable area. 
The selected position should allow the robot to move from its current location while remaining as close as possible to the target object, i.e., 
% such that the robot can navigate from its current location to the selected point, while positioning itself as close as possible to the target object 
$x_i^{\text{next}} = \arg \min_{x \in X_{\text{free}}^t} cost(x,X_{\text{goal}}^t)$.
The relative displacement $(\Delta x, \Delta y)$ is then computed from $x_i^{\text{next}}$ and the current base position.
This design leverages the spatial reasoning capabilities of LLMs, an ability that has also been the subject of recent investigations in related work~\cite{mandi2024roco}. App\ref{lab:HMRCP} provides the detailed prompt design.

% For tabletop manipulation tasks, it is essential to evaluate the success of pick-and-place actions and ensure that the outcomes align with the task's overall objectives following the execution of the planned motions.

\subsubsection{Reflection and Group Discussion Stage}\hfill\\
\indent\textbf{Reflection Module.}

In the planning module, the strategy of next-action prediction based on feedback in a closed-loop manner often leads to short-sighted behavior, especially in long-horizon or dynamically changing tasks. In such tasks, robots need to reflect on past experiences and lessons to improve the accuracy and efficiency of task planning. To address the above issues, we design a Reflection Module, which operates in two main steps:
\begin{enumerate}
    \item At fixed intervals, a group discussion is initiated among the collaborating robots. Each robot is required to compare the current task state with the target objectives, analyze the current environmental observations along with various historical data, and summarize successful experiences and lessons to inform its subsequent task planning.
    \begin{equation}
        (RS^t, RP^t) = \{(rs_i^t, rp_i^t) = u_i(f_i(o_i^t), \overline{h}_i^t) \mid i = 1, \dots, n\},
    \end{equation}
    where $rs_i^t$ and $rp_i^t$ denote the task summary and the task plan generated by robot $i$ during the reflection phase, respectively. $\overline{h}_i^t$ represents the historical context spanning a longer temporal horizon, formally defined as:
    \begin{equation}
        \overline{h}_i^t = \{(fh_i^{t-\overline{k}}, mh_i^{t-\overline{k}}, ah_i^{t-\overline{k}}) \mid \overline{k} = \overline{K}, \dots, 1\},\  \overline{K} > K.
    \end{equation}
    \item Based on both its own reflection and the summaries and plans provided by other team members, the team leader integrates this information and proposes an updated task plan,
    \begin{equation}
        P_E = \mathrm{Update}_E(RS^t, RP^t).
    \end{equation}
\end{enumerate}

This design enables the team to accomplish long-horizon tasks more efficiently while learning from past mistakes to avoid repeated failures. App\ref{lab:RGDP} details the prompt design.

\subsection{Complexity Discussion}
We briefly analyze the framework in terms of context growth and communication complexity. Let $n$ denote the number of robots, $H$ the task horizon, $\Delta t$ the discussion interval, and $K_{\mathrm{sel}}$ the maximum number of voting rounds. With bounded history queues, increasing the number of task objects mainly enlarges the task specification and execution horizon, so the total planning workload grows approximately linearly with $H$.

In the one-time initialization stages, each robot reasons over information from the other agents, so the per-agent context grows as $O(n)$ and the total team-wide processing cost grows as $O(n^2)$. Leader election requires at most $K_{\mathrm{sel}}$ rounds, with $O(K_{\mathrm{sel}}n)$ logical messages in total, or $O(K_{\mathrm{sel}}n^2)$ message deliveries under broadcast semantics. During ordinary execution, planning remains local to each robot, while periodic group discussion introduces only $O(n)$ communication every $\Delta t$ steps. Thus, compared with centralized planners whose single-model context continuously grows with global team information, our framework keeps most computation local and only performs low-frequency synchronization.

\subsection{Data Construction and Model Adaptation}

Our framework leverages the rich semantic knowledge and powerful reasoning capabilities of LLMs to achieve strong few-shot performance in dynamic tasks. However, current LLMs are predominantly trained on internet-scale corpora, which fall short of the specialized requirements of heterogeneous multi-robot collaboration. Due to inherent hallucinations and the absence of domain-specific training data, their action outputs often remain unstable, inaccurate, and inefficient for task execution. This limitation persists even with mitigation strategies such as CoT prompting and other prompt engineering techniques. Sec.~\ref {sec:empirical_study} provides a detailed discussion. Moreover, in real-world robotic scenarios, the limited computational resources of edge devices impose additional constraints on model size and inference speed. To address these challenges, we collected a domain-specific dataset for heterogeneous multi-robot collaboration in simulation environments and fine-tuned an open-source LLM.

\subsubsection{Data Generation Pipeline}

To efficiently gather diverse data, we propose a dual-source action generation framework that integrates both human-designed and LLM-generated scripts. These action sequences are executed across a variety of environments and tasks within the proposed collaboration framework, where environmental constraints enforce the validity of actions. Trajectories that successfully complete the tasks are retained and constitute the foundation of our dataset. The overall data collection pipeline is illustrated in Fig.~\ref{fig:data}.

\begin{enumerate}[label=\roman*.]
    \item \textbf{\textit{Human-Designed Action Scripts}}: Human experts carefully design action sequences based on task requirements. These scripts prioritize efficiency by minimizing redundant steps and yielding shorter trajectories. As reliable expert demonstrations, they provide consistent baselines for effective multi-robot collaboration. Nevertheless, the reliance on predefined patterns limits their diversity, which constrains the model’s adaptability to varied environments and unexpected conditions.
    \item \textbf{\textit{LLM-Generated Action Scripts}}: We employ \textit{GPT-4o} to automatically generate action scripts under the same task specifications. Compared with human-designed scripts, these sequences offer greater diversity, including variations in action ordering, division of labor, and execution strategies, thus broadening the action distribution. However, such diversity may introduce suboptimal or infeasible actions, requiring human judgment and subsequent cleaning. This human-in-the-loop process ensures, while enriching diversity, the resulting dataset also preserves efficiency and reliability for multi-robot collaboration.
\end{enumerate}

Beyond collecting trajectories, we transform them into step-wise decision-making instances. At every step, we record the system prompt and user prompt, which encapsulate task background, goals, historical and current states, executed actions, and feedback from the environment. Along with the ground-truth action, these prompts are fed to \textit{GPT-4o} to generate the corresponding reasoning process. The dataset we collected for model training consists of 30,390 samples. We provide the validation loss learning curves under different training data sizes in App\ref{lab:dataset_size_curves}.

This design serves two purposes: (1) it enriches the dataset with explicit decision rationales, enabling training that combines action supervision with reasoning supervision, and (2) it aligns with the principle of the test-time scaling law. By structuring each step with progressively richer context and generating detailed reasoning traces, the model is encouraged to leverage longer chains of thought and deeper contextual information at inference time, thereby improving performance without additional parameter updates. To demonstrate the importance of the two supervision signals, we conduct an ablation study in App\ref{lab:fine_tuning}, comparing reasoning-only supervision, action-only supervision, and their joint supervision.

%% file: table/feedback.tex
\begin{table}[!htbp]
\centering
\caption{Feedback Description}
\label{tab:feeb_description}
\renewcommand{\arraystretch}{0.6}
\small
\setlength{\tabcolsep}{4pt}
\fitintolinewidth{
\begin{tabular}{>{\centering\arraybackslash}m{2.5cm} 
                >{\centering\arraybackslash}m{14.0cm} 
                }
\toprule
\textbf{Type} & \textbf{Description} \\
\midrule
\rowcolor[gray]{.9} \multicolumn{2}{c}{\textbf{Feedback on successful actions}} \\
\midrule
Navigation Success & Provide feedback confirming successful arrival at the designated target point. Additionally, for furniture items such as tables that do not necessitate opening, include information on the types of objects placed on their surfaces. \\
\midrule
Open Success & Provide feedback confirming the successful opening of the target object, along with detailed information regarding the types and positions of items contained within it. \\
\midrule
Move Success & Provide feedback indicating successful displacement by the specified distances along both the x-axis and y-axis. \\
\midrule
Pick Success & Return feedback that the target object has been successfully picked. \\
\midrule
Place Success & Provide feedback on the successful placement of the target object at the specified location. \\
\midrule
\rowcolor[gray]{.9} \multicolumn{2}{c}{\textbf{Feedback on failed actions}} \\
\midrule
Navigation Failed & (1) The starting or ending point of global path planning is deemed invalid if it falls on an obstacle or exceeds the map boundary; (2) The target object for navigation is considered invalid if it does not exist in the scene graph or does not support navigation; (3) A discrepancy greater than an acceptable threshold between the robot's current pose and the target pose can lead to failure; (4) Navigation failures caused by limitations in robot capabilities (e.g., inability to explore elevated areas) or environmental constraints (e.g., restricted or inaccessible zones).\\
\midrule
Open Failed & (1) The target object is either already in an open state or cannot be opened; (2) The target object is positioned beyond the operational range of the robot. \\
\midrule
Move Failed & The feedback type aligns with "navigation failed" due to the invocation of the API responsible for the \textit{navigate} command. \\
\midrule
Pick Failed & (1) The gripper is grabbing with other objects; (2) The scene graph lacks information about the object to be grasped; (3) The initial or target state is invalid during the verification of the robot arm planning algorithm; (4) The distance between the robot arm’s end effector and the target object exceeds the allowable threshold. Return this distance, and for mobile manipulation robots, also provide the relative distance between the base and the target object along the x and y axes. \\
\midrule
Place Failed & (1) The gripper is empty; (2) The object to be placed does not match the object currently being grasped; (3) The object to be placed has not been placed at the target location; (4) The placement violates task constraints. \\
\midrule
\rowcolor[gray]{.9} \multicolumn{2}{c}{\textbf{Common feedback}} \\
\midrule
Target Task Status & (1) In the "pack objects" task, the feedback refers to the types of objects present in the tray; (2) In the "sort solids" task, it provides information regarding the shapes and colors of solids placed on panels of different colors; (3) In the "make sandwich" task, the feedback describes the types of ingredients arranged on the cutting board from bottom to top. \\
\bottomrule
\end{tabular}}
\end{table}

%% file: table/action.tex
\begin{table}[!htbp]
\centering
\caption{The Atomic Action Set of Different Types of Robots}
\label{tab:atomic_action}
\renewcommand{\arraystretch}{1.08}
\small
\setlength{\tabcolsep}{5pt}
\begin{tabularx}{\linewidth}{
>{\centering\arraybackslash}p{0.26\linewidth} 
>{\raggedright\arraybackslash}X
}
\toprule
\textbf{Type} & \multicolumn{1}{c}{\textbf{Atomic Action Set}} \\
\midrule
Mobile Manipulation Robot & 
1. \texttt{[navigate] to <stand\_pose\_id> of <object>};
2. \texttt{[open] <container>};
3. \texttt{[pick] up <object>};
4. \texttt{[place] <object> on/into <platform>};
5. \texttt{[move] <delta\_x> and <delta\_y>};
6. \texttt{[communicate] <content> to <role>};
7. \texttt{[wait]} \\
\midrule
Manipulation Robot & 
1. \texttt{[pick] up <object>};
2. \texttt{[place] <object> on/into <platform>};
3. \texttt{[communicate] <content> to <role>};
4. \texttt{[wait]} \\
\midrule
Mobile Robot & 
1. \texttt{[navigate] to <stand\_pose\_id> of <object>};
2. \texttt{[communicate] <content> to <role>};
3. \texttt{[wait]} \\
\midrule
Drone Robot & 
1. \texttt{[navigate] to <stand\_pose\_id> of <object>};
2. \texttt{[pick] up <object>};
3. \texttt{[place] <object> on/into <platform>};
4. \texttt{[communicate] <content> to <role>};
5. \texttt{[wait]} \\
\bottomrule
\end{tabularx}
\end{table}

%% file: chapters/4.experiment.tex
\section{experiment}

\input{table/constant}

\subsection{Benchmark}

\begin{wrapfigure}{r}{0.5\textwidth}
    \centering
    \includegraphics[width=\linewidth]{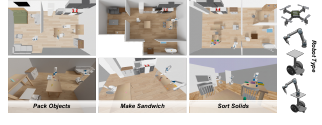}
    \caption{This figure describes the experimental scenarios, tasks, and robot types involved in our research. The furniture types and layouts differ across various settings, with distinct room configurations, such as kitchens, bathrooms, and bedrooms. The task design is inspired by RoCo~\cite{mandi2024roco}, encompassing activities such as sandwich making, sorting solid objects, and packing items.}
    \label{fig:task}
\end{wrapfigure}

In this study, we constructed several experimental environments using the BestMan~\cite{ding2023task, ding2023integrating} simulation platform and developed additional functionalities to support the specific tasks of our research. Each environment features a variety of room types and layouts, diverse furniture types and arrangements, and an array of different objects. The first row of Fig.~\ref{fig:task} illustrates the environments corresponding to these three different scenarios, respectively. For the robot navigation algorithm, we use the A*~\cite{Kurzer1057261} algorithm for global path planning without considering local obstacle avoidance~\cite{fox1997dynamic, yu2024pathrl, LDP}, as it is not the focus of our task. For the robotic arm planning algorithm, we use the sample-based BIT*~\cite{gammell2015batch} algorithm, considering path planning in a static state.

We also designed the following three different tasks:

\begin{itemize}
    \item \textit{Pack Objects}: The objective of this task is to evaluate the robot's fundamental picking and placing capabilities. The robot is provided with a list of objects and must accurately place each object into a designated tray.
    \item \textit{Sort solids}: In addition to evaluating the robot's basic pick-and-place capabilities, this task also requires the robot to perform color matching. The robot must accurately place solids, each of a different color, onto the corresponding colored panels on the table.
    \item \textit{Make Sandwich}: This task further tests the robot's ability to stack objects in a specific order. The robot is tasked with assembling sandwiches of varying flavors based on a given menu, requiring it to sequentially stack the sandwich ingredients.
\end{itemize}

\input{table/hyperparameters}

\begin{figure*}[h]
    \centering
    \includegraphics[width=\textwidth, height=0.4\textwidth]{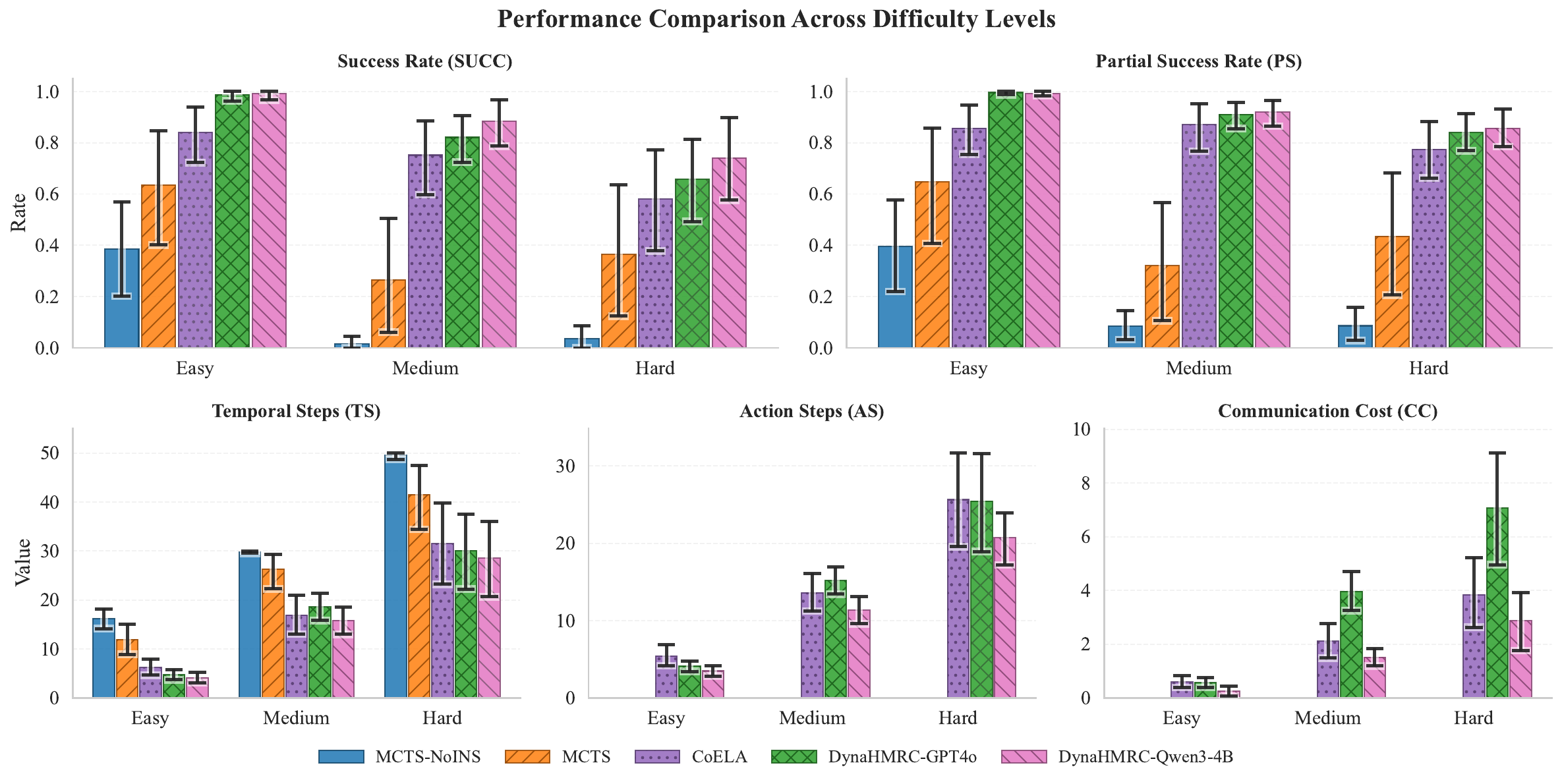}
    \caption{Performance of different methods under different task difficulties. The five subplots, arranged from top to bottom and left to right, present the overall task success rate (SUCC), partial success rate (PS), average temporal steps (TS), average action steps (AS), and average communication cost (CC), respectively. The error bars represent 95\% confidence intervals computed via bootstrap resampling with 10,000 iterations. Each task difficulty level is associated with a maximum step limit, set to 20, 30, and 50 for the easy, medium, and hard tasks, respectively. App\ref{lab:box_plots} presents the box plots for the Pack Objects task under different difficulty levels.}
    \label{fig:robodiff}
\end{figure*}

\begin{figure*}[h]
    \centering
    \includegraphics[width=\textwidth]{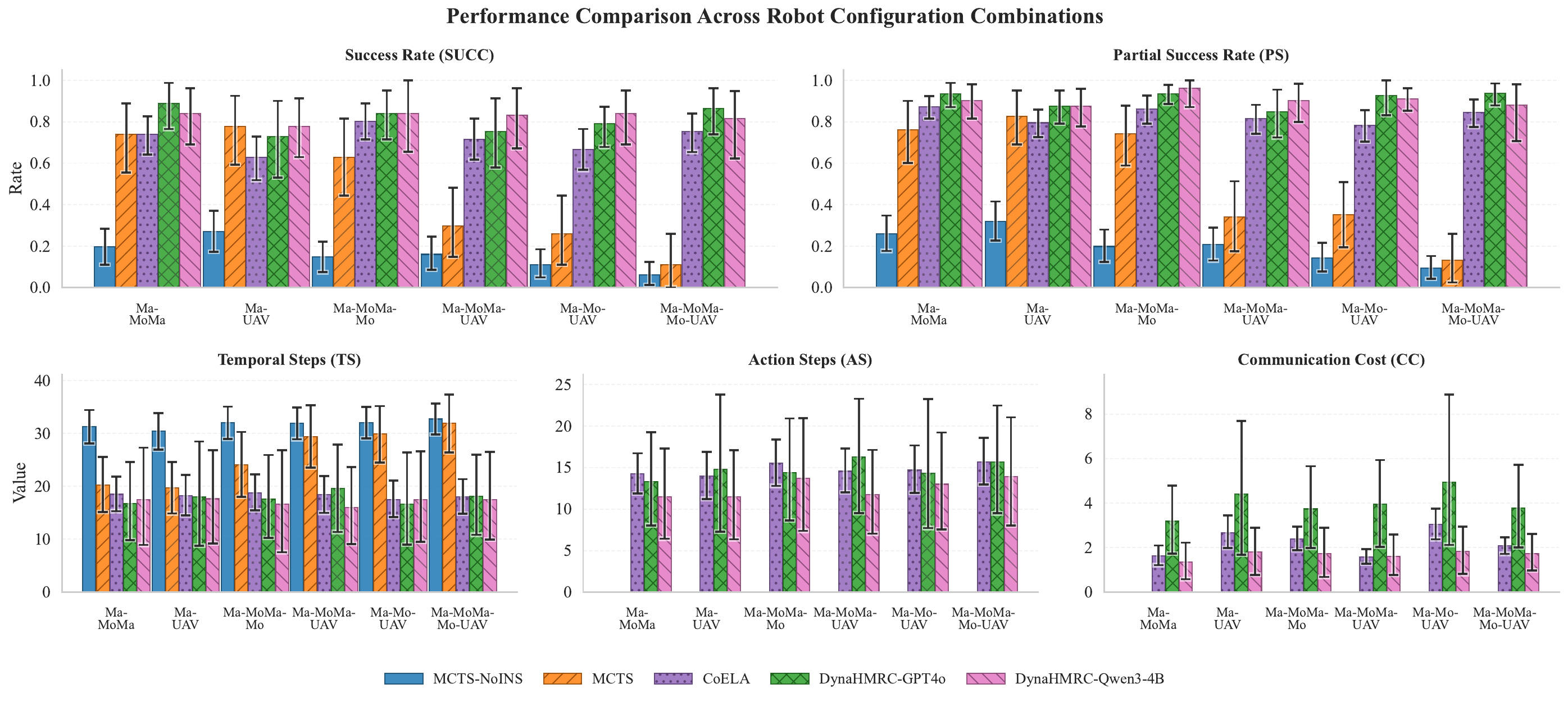}
    \caption{Performance of different methods under different robot configuration combinations. The arrangement of the subplots follows the same display configuration as in Fig.~\ref{fig:robodiff}.}
    \label{fig:robocomb}
\end{figure*}

The objects involved in the aforementioned tasks are randomly distributed within the environment. As a result, in addition to executing the required manipulation actions, robots must also explore the unknown environment to locate target objects. This requirement significantly extends the task sequence and increases the overall complexity of the task.
The design of these tasks presents a significant challenge to robots’ mobile manipulation capabilities in unknown environments, aligning with a key research focus in the field~\cite{rulebook_2025, melnik2023uniteam, gan2021threedworld}.

To facilitate a more comprehensive evaluation of task difficulty, we propose a unified task difficulty metric that integrates multiple contributing factors, including object location complexity, object quantity, and collaboration intensity.
Let $J$ denote the number of target objects in the task. The overall task difficulty is defined as:
\begin{equation}
    D=\alpha L + \beta N + \gamma Y,\ D \in [0,1],
    \label{equ:difficulty_metric}
\end{equation}
where each term corresponds to a specific aspect of task complexity and $\alpha + \beta + \gamma = 1, \alpha, \beta, \gamma \ge 0$. Given the importance of exploration and collaboration in successful task execution, we assign the weights $\alpha=0.4$, $\beta=0.2$, and $\gamma=0.4$, thereby placing greater emphasis on location difficulty and collaboration intensity. 
These weights are heuristic design choices introduced for coarse-grained task categorization rather than learned parameters. We assign a lower weight to the object quantity factor because, in our benchmark, increasing the number of target objects primarily extends the task duration, whereas location difficulty and collaboration intensity more directly determine the intrinsic complexity of execution. In particular, even a single object can significantly increase task difficulty if it lies outside the manipulator’s reachable workspace or requires additional exploration, transportation, or inter-robot coordination.
Based on the overall difficulty factor $D$, we categorize our tasks into three levels: easy,  medium, and hard. Specifically, tasks with $D \in [0, 0.3]$ are considered easy, those with $D \in (0.3, 0.6)$ are classified as medium, and tasks with $D \in [0.6, 1]$ are defined as hard.

Specifically, $L$ denotes the location difficulty factor, which incorporates both the maximum and average difficulty levels associated with object locations.
In our task setting, the location difficulty of an object is categorized into three levels:
\begin{itemize}
    \item $\ell_i = 0$: The object is placed on the tabletop where the manipulation robot’s base is located and lies within its operational range.
    \item $\ell_i = 1$: The object is also placed on the tabletop but lies outside the robot's operational range, necessitating at least one instance of inter-robot collaboration.
    \item $\ell_i = 2$: The object is not on the tabletop, requiring additional exploration steps and transportation efforts to move it onto the working surface.
\end{itemize}

Accordingly, the location difficulty factor $L$ is formally defined as follows:
\begin{equation}
    L = \lambda \frac{max_j \ell_j}{2} + (1 - \lambda) \frac{1}{2J} \sum_{j=1}^{J}\ell_j,\ L \in [0, 1],
\end{equation}
where $\lambda$ is set to 0.8 to place greater emphasis on maximum location difficulty among all objects. This weighted formulation prevents the overall location difficulty from being underestimated when low-difficulty objects are added.

% Among the three tasks we designed, Make-Sandwich and Sort-Solids involve additional requirements related to sequencing and matching, which are not present in the Pack-Objects task. To reflect these added complexities, we introduce the constraint complexity factor $C$, which captures the increased difficulty arising from such task-specific constraints.
% \begin{equation}
%     C = \begin{cases}
%     0, & \text{pack objects.} \\
%     1, & \text{make sandwich \& sort solids .} 
%     \end{cases}
% \end{equation}

In general, tasks involving a larger number of target objects tend to require longer execution times. To capture this aspect of difficulty, we define the object quantity factor $N$, which quantifies the impact of object count on overall task difficulty.
\begin{equation}
    N = \frac{J - J_{min}}{J_{max} - J_{min}},\ N \in [0, 1],
\end{equation}
where $J_{min}$ and $J_{max}$ denote the minimum and maximum numbers of target objects observed across all experimental tasks, which are set to 1 and 5, respectively.

Finally, we evaluate task difficulty from the perspective of robot collaboration. For a given task, we define $y$ as the minimum number of additional robots required, beyond a single manipulation robot, to complete the task successfully. In our experiments, all tasks can be accomplished using no more than three heterogeneous robots, hence $y \in \{0, 1, 2\}$. Based on this definition, we introduce the collaboration intensity factor as follows:
\begin{equation}
    Y = \frac{y}{Y_{max}},\ Y_{max} = 2,\ Y \in [0, 1].
\end{equation}

Fig.~\ref{fig:robodiff} shows that as task difficulty increases, the success rate and partial success rate of all methods gradually decrease, while the number of steps steadily increases. This trend validates the rationality of our difficulty settings. We further provide a sensitivity analysis of Eq.~\eqref{equ:difficulty_metric} in App\ref{lab:sensitivity}, showing that the empirical alignment of the difficulty score remains stable under reasonable weight perturbations, along with correlation plots between object quantity and the task metrics PS and TS.

\subsection{Baselines}
In this work, we establish two main categories of baselines: traditional heuristic planning algorithms (\textit{MCTS}, \textit{MCTS-NoINS}) and generative model–based approaches (\textit{COHERENT}, \textit{CoELA}, \sysnameo, \sysnameq). We do not include classical decentralized coordination methods as direct baselines because they rely on substantially different problem formulations from ours. In particular, such methods typically require explicit task decomposition, hand-crafted utility/reward design, and structured symbolic state abstractions, whereas our setting emphasizes long-horizon heterogeneous collaboration from semantic task descriptions, textual feedback, and dynamically changing task conditions. Constructing a fair classical baseline under our setting would therefore require substantial manual engineering, which would confound the comparison. All methods are carefully optimized to maximize their performance across different experimental settings, ensuring fairness in comparison. During model inference, we set the temperature parameter to 0.5 to strike a balance between stability and diversity in action sequence generation. This value ensures the rationality of the planning outcomes while preserving a degree of explorativeness, thereby enabling adaptation to complex and dynamic task scenarios. It is worth noting that all of our experiments were designed to ensure feasibility under the corresponding settings. In other words, we avoided scenarios where the task itself would be inherently impossible to complete. For example, tasks requiring aerial object manipulation were not assigned to robot configurations lacking a drone. This design choice guarantees the fairness and validity of the experimental comparisons. In all experiments, the memory queue length is set to $M_l = 10$ by default. We further investigate the effect of a larger memory size ($M_l = 15$) on the experimental results in App\ref{lab:diff_memory_length}.
\begin{itemize}
    \item \textit{MCTS}: We used Monte Carlo tree search~\cite{browne2012survey} with human prior knowledge, operating on global observations of all agents and exploring their joint action space. Since navigation actions dominate the search space and reduce efficiency through repetition, we apply a heuristic: prioritize non-navigation actions when the previous action was navigational and non-navigation options exist. We further improve efficiency by pruning infeasible actions (e.g., grasping out-of-reach objects) and restricting actions to those relevant to target objects.
    \item \textit{MCTS-NoINS}: We implemented a variant of the MCTS algorithm that removes the human instruction to further validate the performance of the basic MCTS.  Except for this modification, all other experimental settings, including the pruning strategy and action space constraints, remained identical to those of the MCTS baseline.
    \item \textit{COHERENT}: The COHERENT~\cite{liu2025coherent} employs a centralized, high-level LLM-based planning architecture. The system derives task and semantic scene representations from global observations; the LLM orchestrates high-level planning, producing the complete set of action commands for all robots at every environment timestep, while the robots solely execute the issued actions. The LLM has access to global robot observations and the complete global interaction history.
    \item \textit{CoELA}: The CoELA~\cite{zhang2023building} employs a decentralized, cognitive-inspired framework where LLMs perform high-level planning and communication, while dedicated modules manage low-level perception and execution. This design leverages the LLM's reasoning capabilities for complex multi-agent tasks in decentralized settings.
    \item \sysnameo: Based on our proposed heterogeneous multi-robot collaboration framework, task understanding, reasoning, and planning are performed using the closed-source model \textit{GPT-4o}.
    \item \sysnameq: We employ the open-source \textit{Qwen3-4B-Instruct} model for the same processes within our framework. This model is fully fine-tuned on the dataset we constructed to enhance its performance on domain-specific tasks. Furthermore, the relatively moderate scale of 4B parameters endows the model with promising feasibility for deployment on resource-constrained robotic edge devices. The detailed training configurations of \textit{Qwen3-4B} are provided in Tab.~\ref{tab:hyper}. The sensitivity analysis of training hyperparameters is provided in App\ref{lab:hyper_sensitivity}.
\end{itemize}

\subsection{Task Variations and Constraints}
Building upon the three primary task categories, we additionally introduce four types of variations during task execution to evaluate the robustness of the proposed framework in adapting to dynamic task conditions:
\begin{itemize}
    \item \textit{Changes in the overall task objective (CTO)}: Changes in the number and types of target objects may necessitate additional adaptations, particularly when constraints exist among the objects, in order to achieve the task objectives.
    \item \textit{Introduction of temporary restricted zones (IRZ)}: If a team member is temporarily restricted from accessing a specific area, the framework must reassign the corresponding exploration responsibilities to other collaborators to ensure successful task execution.
    \item \textit{Addition of new collaborators (ANC)}: The addition of new collaborators requires a reallocation of task responsibilities. The self-description of a newly joined collaborator is broadcast to all existing team members.
    \item \textit{Removal of existing collaborators (REC)}: The removal of collaborators likewise demands a redistribution of tasks to maintain overall performance. To simulate unexpected failures, robots are deliberately disconnected from the team without any prior indication.
\end{itemize}
 
In robotic tasks, fixed constraints play a crucial role alongside dynamic task variations. A notable example is the capability limitations induced by heterogeneous robot configurations. Beyond this, we introduce an implicit constraint into the task: only manipulation robots are permitted to execute the final step, such as placing sandwich ingredients onto the cutting board, while other robots are restricted from placing any objects there. This constraint is not explicitly defined in advance; instead, robots receive feedback only upon violating the constraint through their actions. We regard this implicit rule as a form of experiential learning, showcasing the adaptive learning capabilities of the collaborative framework.

\input{table/goal_change}

\subsection{Evaluation Metrics}
We introduce five metrics to evaluate the performance of different methods:

\begin{itemize}
    \item \textit{Success rate (Succ)}: The rate of episodes in which the robot completes the full task, meaning all objects are placed in the correct positions.
    \item \textit{Partial success rate (PS)}: The average ratio of correctly placed objects to the total number of objects per episode. This metric reflects task difficulty: as the number of objects and sequence length increase, the task becomes more challenging, but individual failures have less impact. In contrast, tasks with fewer objects are more sensitive to single failures.
    \item \textit{Average temporal steps (TS)}: the average number of temporal steps in all episodes.
    \item \textit{Average action steps (AS)}: the average number of action steps in all episodes, excluding \textit{wait} action. The robot's capability to execute \textit{wait} action at optimal moments can reduce energy consumption, demonstrating that the policy has a comprehensive understanding of the entire task.
    \item \textit{Average communication cost (CC)}: the average number of communication events per episode, defined as $\mathrm{CC}=\frac{1}{|n\mathcal{E}|}\sum_{e\in\mathcal{E}}\sum_{t=1}^{\mathcal{H}}\sum_{i=1}^{n} c_{i,e}^{t}$, where $c_{i,e}^{t}=1$ if robot $i$ executes \texttt{communicate} at step $t$ in episode $e$, and $c_{i,e}^{t}=0$ otherwise.
\end{itemize}

\input{table/area_change}
\input{table/addition_change}
\input{table/removal_change}

\subsection{Experiments on Simulation Scenarios}
\subsubsection{Static Comparative experiments}
To thoroughly validate the effectiveness of our framework, we conduct a series of combinatorial experiments across different settings, including varying scenarios, task types, difficulty levels, and robot configuration combinations. The experiments are performed under both traditional baselines and decentralized approaches driven by LLMs. Tab.~\ref{tab:constant_exp} presents the detailed experimental results, where $T=0$ indicates that no dynamic changes were introduced during the task execution.
The experiments involve combinations of $2–4$ robot types, including manipulation (Ma), mobile (Mo), mobile manipulation (MoMa), and drone (UAV) robots. Specifically, the configurations are ``Ma-MoMa'' and ``Ma-UAV'' for two types, ``Ma-MoMa-Mo'', ``Ma-MoMa-UAV'', and ``Ma-Mo-UAV'' for three types, and ``Ma-MoMa-Mo-UAV'' for four types.
% The combinations of robot types and quantities used in the experiments are as follows:
% \begin{itemize}
%     \item \textbf{2 types of robot}: [a manipulation robot and a mobile manipulation robot (Ma-MoMa)], [a manipulation robot and a drone (Ma-UAV)]
%     \item \textbf{3 types of robot}: [a manipulation robot, a mobile manipulation robot, and a mobile robot (Ma-MoMa-Mo)], [a manipulation robot, a mobile manipulation robot, and a drone (Ma-MoMa-UAV)], [a manipulation robot, a mobile robot, and a drone (Ma-Mo-UAV)]
%     \item \textbf{4 types of robot}: [a manipulation robot, a mobile manipulation robot, a mobile robot, and a drone (Ma-MoMa-Mo-UAV)]
% \end{itemize}

Tab.~\ref{tab:constant_exp}, Fig.~\ref{fig:robodiff}, and Fig.~\ref{fig:robocomb} present the performance of different methods under static task conditions, across variations in task type, task difficulty, and robot configuration combinations.
% For each task setting, we randomly sample three sets of target objects and conduct experiments on each set. To mitigate the hallucination problem of LLMs and ensure the stability and reproducibility of results, each experiment is repeated three times, and the average performance is reported. Therefore, the data presented in the ``2 types of robot'' section of the Tab.~\ref{tab:constant_exp} were obtained from 18 experimental trials, while the ``3 types of robot'' and ``4 types of robot'' settings required 27 and 9 trials, respectively. Such a large number of experiments ensures the statistical significance and reliability of our results.
For each task setting, we randomly sample three target object sets and repeat each experiment three times to reduce LLM hallucination and improve reproducibility, and the average performance is reported. Consequently, the ``2 types of robot'', ``3 types of robot'' and ``4 types of robot'' settings involve 18, 27, and 9 trials, respectively, ensuring statistical reliability and robustness of the results.
Overall, \sysnameq demonstrates the best comprehensive performance. Compared with other baseline methods, it achieves higher task and partial success rates while requiring fewer steps to complete the tasks, particularly in terms of action and communication steps, and demonstrates greater overall stability in the experimental results. This reduction in steps is crucial for robotic tasks, as fewer steps directly correspond to lower energy consumption. Compared with \textit{CoELA}, which also leverages \textit{GPT-4o}, the \sysnameo algorithm achieves higher overall and partial success rates; however, it does not demonstrate a significant advantage in terms of task-step efficiency across different scenarios and tasks. These results highlight the necessity of collecting domain-specific task data and training corresponding models. 

The baseline algorithm \textit{MCTS-NoINS} performs poorly across all metrics and achieves only limited success on simple tasks. This weakness stems from the absence of action guidance to prune search tree branches, leaving the search space excessively large and making it difficult to find optimal solutions within limited time. In contrast, \textit{MCTS} shows relatively better performance, achieving success rates comparable to \sysnameq on simpler tasks with fewer robots. However, the performance of all MCTS-based methods is fundamentally constrained by the size of the search space. Experimental results reveal a steep decline in performance as task difficulty increases and the number and heterogeneity of robots grow—sometimes dropping to zero. This trend is clearly reflected in the SUCC and PS subfigure of Fig.~\ref{fig:robocomb}, where the orange and blue bars decrease steadily as the number of robots increases. 

In contrast, LLM-based methods maintain more stable performance across the evaluated robot configurations, suggesting better scalability within the small-to-medium heterogeneous team settings studied in this work. Unlike \textit{MCTS}, which suffers from exponential growth in search space due to centralized joint modeling, the decentralized role-playing framework assigns each robot an independent LLM agent. This allows reasoning and decision-making to proceed in parallel from local perspectives, while lightweight communication enables effective collaboration. Such a design naturally supports heterogeneous robots without explicit joint modeling, yielding improved scalability and robustness trends in both quantity and diversity dimensions. We further demonstrate the generalization capability of \sysnameq to unseen target objects under static tasks of medium difficulty in App\ref{lab:obj_generalization}.

Moreover, we observe that MCTS-based methods perform particularly poorly on the Sort Solids task, compared with the other two task categories. This degradation arises because the task introduces an additional requirement of matching six distinct colors, which dramatically enlarges the search space. In contrast, LLMs-based methods leverage their strong semantic understanding capabilities to handle such challenges effectively. To further enhance the capability of MCTS in handling semantic tasks, the comparative analysis experiments with the LLM-MCTS baseline are provided in App\ref{lab:llm_mcts_baseline}.

\begin{figure*}[htbp]
    \centering
    \includegraphics[width=\textwidth, height=0.4\textwidth]{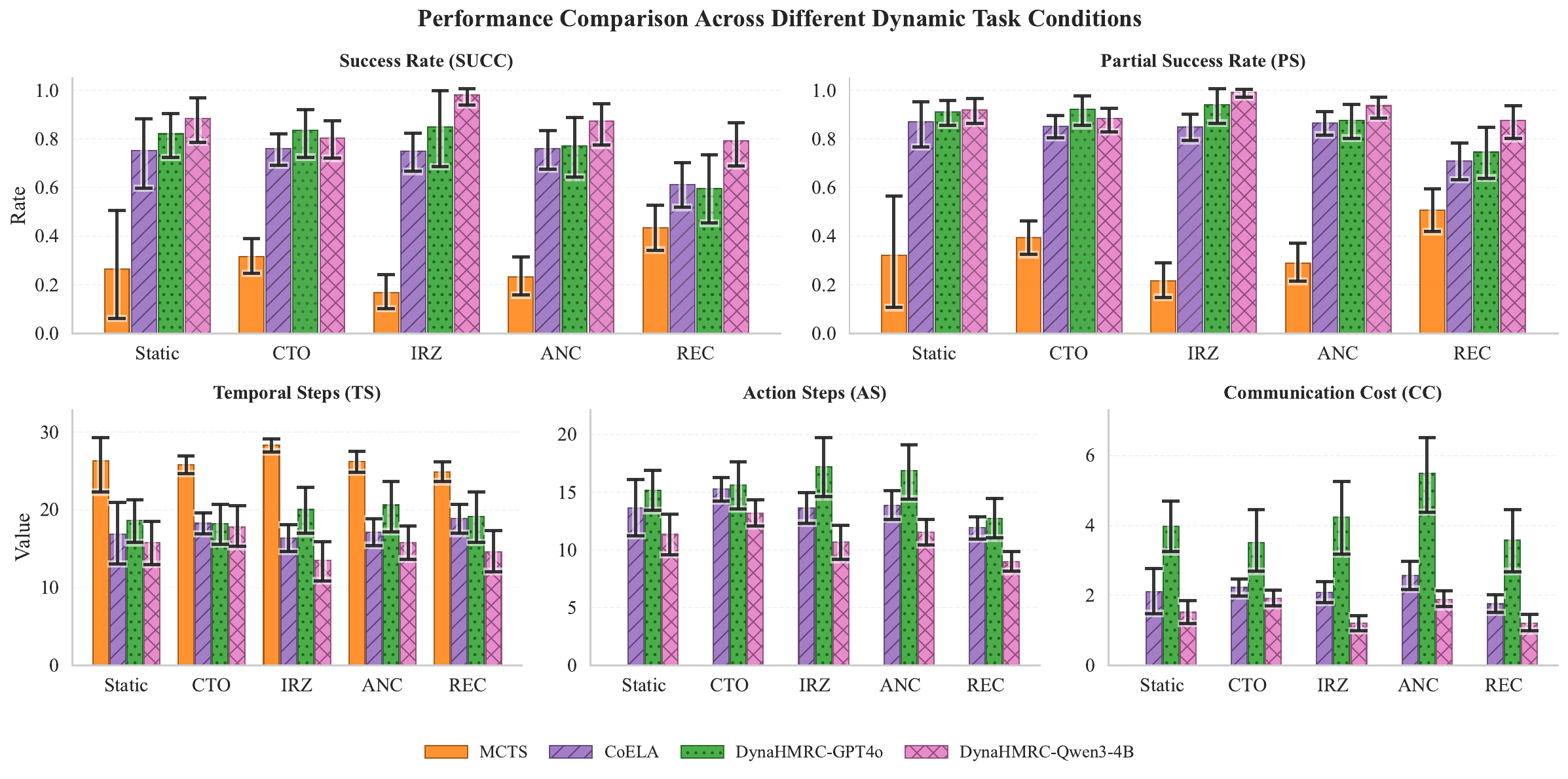}
    \caption{Performance of different methods under different dynamic task conditions. The five subplots, arranged from top to bottom and left to right, present the overall task success rate (SUCC), partial success rate (PS), average temporal steps (TS), average action steps (AS), and average communication cost (CC), respectively. The error bars represent 95\% confidence intervals computed via bootstrap resampling with 10,000 iterations. All experiments are conducted under the medium-difficulty setting, where the maximum number of steps is capped at 30.}
    \label{fig:robodyta}
\end{figure*}

\input{table/central_1}

\input{table/central_2}

\input{table/ablation}

\subsubsection{Dynamic Comparative experiments}

To further assess adaptability, the impact of dynamic task conditions on algorithmic performance is investigated under the medium-difficulty setting. Tab.~\ref{tab:goal_exp} -~\ref{tab:removal_exp} and Fig.~\ref{fig:robodyta} illustrate the performance of different algorithms under various dynamic task conditions. 

When task objectives change during execution, the success rate of \textit{MCTS} increases slightly from $26.5\%$ to $30.4\%$, indicating that its search process can partially adapt to goal shifts. Nevertheless, its overall performance remains unsatisfactory. In contrast, LLM-based methods demonstrate substantially stronger performance, with \sysnameo achieving the highest success rate of $83.5\%$. Interestingly, the results show that \sysnameq underperforms \sysnameo in the CTO task, despite consistently outperforming it in the other three tasks. To account for this discrepancy, we empirically identify an efficiency–robustness trade-off emerging in heterogeneous multi-robot collaboration: efficient strategies can complete tasks quickly under static goals, but their strong structural commitments to initial objectives make them less adaptable when goals change dynamically. In such cases, additional backtracking is required to correct mismatches, limiting adaptability. In contrast, less efficient strategies progress more slowly but benefit from delayed commitment, enabling smoother transitions to new goals when changes occur, thereby exhibiting stronger robustness. For example, in the sandwich-making task, once the base ingredient changes, the more layers stacked before the change, the more steps are required to remove the top ingredients, amplifying the adaptation cost.

Under restricted-area conditions, the performance gap among methods becomes particularly pronounced in the IRZ task. The success rate of \textit{MCTS} drops sharply to $12.4\%$, with a partial success rate of only $16.9\%$. By contrast, \textit{CoELA} maintains $77.2\%$, \sysnameo achieves $84.9\%$, and \sysnameq further improves to $98.2\%$. This large discrepancy arises because temporary regional constraints are introduced only through textual feedback, where navigation actions into restricted areas are passed into the simulator but do not modify the environment state. As a result, search-based algorithms such as \textit{MCTS}, when lacking additional pruning strategies, repeatedly revisit the same nodes, thereby degrading performance. Conversely, LLM-based methods, by leveraging their semantic understanding and reasoning capabilities, can exploit these constraints to effectively reduce the exploration space and improve task performance.

When new robots are introduced during collaboration, \textit{MCTS} achieves only a $23.2\%$ success rate, reflecting its limited dynamic scalability. \textit{CoELA} remains stable at $75.9\%$, \sysnameo improves to $77.2\%$, and \sysnameq reaches $87.4\%$. These findings highlight that the decentralized collaboration framework offers clear advantages in scaling to heterogeneous teams, while fine-tuned models demonstrate strong specialization for domain-specific tasks.

In scenarios where collaborators are removed, the success rate of \textit{MCTS} rises to $43.5\%$, as reducing the number of robots decreases the search space and accelerates convergence. However, LLM-based methods continue to show superiority, with \sysnameo maintaining $59.6\%$ and \sysnameq reaching $79.2\%$. In contrast, \textit{CoELA} experiences a significant performance decline in this setting, underscoring its limited adaptability to sudden reductions in team size. We provide a statistical significance comparison between \sysnameo and \sysnameq on both static and dynamic tasks in App\ref{lab:statistical_significance}, demonstrating the statistically robust and significant advantages of \sysnameq. In addition, to further evaluate the robustness of different methods under varying levels of dynamic changes, we conduct comparative experiments in App\ref{lab:sensitivity_dyna} under different dynamic trigger timings and different numbers of restricted zones.

\subsubsection{Centralized vs. Decentralized LLM-Based Comparative Experiments}
To better demonstrate the advantages of our LLM-based decentralized framework over the centralized approach in handling long-context and multi-objective tasks, Tab.~\ref{tab:central_1} and Tab.~\ref{tab:central_2} present a comparative evaluation among \textit{COHERENT}, \sysnameo, and \sysnameq across various robot configurations and task types.

Overall, our decentralized method \sysname exhibits consistently more competitive performance compared to the centralized baseline \textit{COHERENT}. As the number and configuration of robots increase, \sysname maintains robust adaptability and stable performance across different settings, whereas \textit{COHERENT} suffers from performance degradation with the growth of robot count and context length.

An interesting observation is that although \textit{COHERENT} sometimes achieves the best results on the AS metric, its overall task success rate remains low. Our analysis suggests that \textit{COHERENT}’s single-pass reasoning must simultaneously attend to numerous task-related factors, which reduces both the efficiency and accuracy of its outputs. This leads to repetitive wait actions, further lowering the AS metric.

In addition, we emphasize that the advantage of \sysname is not merely due to shorter prompts. Although centralized planners can be improved through decomposition, summarization, and staged calls, centralized coordination still relies on a compressed global task state. In long-horizon heterogeneous tasks, robot-specific information such as reachability constraints, local occlusions, failed grasps, blocked navigation targets, newly observed objects, or embodiment-specific limitations may be compressed away, misinterpreted, or become stale before the next global decision. This can lead to repeated infeasible actions, delayed reaction to changed task goals, redundant assignments, or embodiment-mismatched action allocation.

In contrast, \sysname preserves first-person planning at the robot level. Each robot reasons over its own observation, memory, execution feedback, and embodiment description, while periodic discussion provides lightweight global synchronization. Thus, the benefit of \sysname comes from combining local embodiment-grounded planning with low-frequency team-level coordination, rather than only reducing context length. A representative \textit{COHERENT} failure case is provided in App\ref{lab:coherent_fail_case}, where an incorrect centralized progress estimate causes global idling and delayed recovery for a missing target object, illustrating how a centralized belief can become a single point of failure, with one error affecting the behavior of all robots.

In addition, we conducted a simple token usage analysis using tiktoken. The average number of tokens per inference for \sysnameq is approximately 1,099, while \textit{COHERENT} requires around 2,227 tokens—more than twice as many. This result indicates that decentralized planning can reduce the single-call context burden, while the total episode-level token usage is further analyzed in App\ref{lab:llm_cost} and App\ref{lab:comm_cost_composition}.

\subsubsection{Ablation Experiments}
To comprehensively evaluate our framework and models, we design two types of ablation studies:
(1) Inference model ablation. We replace \textit{GPT-4o} with alternative closed-source and open-source models to serve as the reasoning core of the overall framework.
(2) Framework component ablation. Building on the fine-tuned \textit{Qwen3-4B-Instruct} in our work, we conduct comparative experiments by ablating individual modules, including \textit{history}, \textit{feedback}, \textit{leader}, and \textit{reflection}. These two sets of studies allow us to disentangle the contributions of the central reasoning model and the auxiliary components, thereby clarifying their respective roles in enhancing efficiency, adaptability, and robustness of heterogeneous multi-robot collaboration. 

Tab.~\ref{tab:ablation} presents the experimental results for these ablation studies. Across five task regimes, \sysnameq achieves the best overall performance, with an average success rate of $86.1\%$. Compared with \sysnameo, it improves SUCC by $8.7\%$ while reducing TS, AS, and CC by $20.1\%$, $28.2\%$, and $62.9\%$, respectively; relative to Claude, SUCC increases by $11.3\%$ with $25.8\%$, $28.8\%$, and $54.8\%$ reductions in the same metrics. The advantages are most pronounced under IRZ, indicating strong adaptability to textual constraints and re-planning. In the CTO, although \sysnameo achieves slightly higher SUCC, \sysnameq compensates with fewer robotic steps, reflecting the fine-tuned model’s domain expertise. For ANC and REC, \sysnameq consistently achieves the highest success rates and the fewest robotic steps, highlighting its robustness to team expansion and reduction.

Ablation studies confirm the necessity of each component. Removing feedback causes the largest drop in task success rate and number of task steps, as robots lose the ability to assess execution outcomes, breaking the closed-loop control. Removing history or reflection degrades performance particularly under CTO and IRZ, underscoring their role in preserving contextual continuity and mid-horizon correction. Introducing a lightweight leader election slightly increases communication but improves decision alignment and reduces redundant actions, validating its effectiveness in decentralized collaboration. These results collectively validate the effectiveness of \sysnameq in achieving efficient, robust, and adaptive collaboration across dynamic heterogeneous multi-robot scenarios.

\subsubsection{Empirical Study}~\label{sec:empirical_study}
Although the proposed system framework enables efficient and robust collaboration among heterogeneous multi-robot teams, residual issues still emerge during execution due to the intrinsic hallucination tendencies of LLMs and the limited domain expertise of proprietary models in specialized environments. These residual problems can compromise task success and efficiency, ultimately degrading overall system performance. The main failure modes are summarized as follows:

\begin{enumerate}[label=\roman*.]  
\item \textbf{State–Task Misalignment.}  
Robots exhibit insufficient awareness of the current task state and may continue performing actions or communications toward subgoals that have already been completed. For example, even when the \textit{bowl} has already been placed inside the \textit{tray}, the \textit{drone} continues to communicate with the \textit{manipulation} robot, reminding it to execute the ``place bowl'' action.

\item \textbf{Progress–Goal–Experience Fusion Gap.}  
Robots fail to integrate prior observations and their own execution experiences into consistent reasoning about the current task progress and goal set, leading to off-topic behaviors and redundant queries. For instance, the \textit{mobile} robot continues to navigate aimlessly after identifying and reporting the target object’s location; the \textit{mobile manipulation} robot repeatedly queries teammates about task progress despite having completed its assigned transportation; and manipulators occasionally grasp objects not included in the designated target list.

\item \textbf{Myopic Redundancy.}  
During task execution, robots demonstrate repetitive and short-sighted behaviors. Some repeatedly send identical messages to collaborators; mobile robots oscillate between a limited set of locations, indicating restricted exploration ability; and manipulators execute consecutive pick-and-place actions on the same object multiple times.

\item \textbf{Capability Grounding Error.}  
Robots show incomplete understanding of their teammates’ action feasibility and capability boundaries, resulting in infeasible or inefficient delegation requests. For example, a \textit{drone} may request a \textit{manipulation} robot to open a refrigerator door even though the tabletop-fixed manipulator cannot reach the target or perform such an action.

\item \textbf{Over-Constrained Division of Labor.}  
Robots tend to rigidly adhere to their initial task assignments and fail to act autonomously even when capable of contributing to other subtasks, leading to systemic idle waiting and reduced overall efficiency. For example, a \textit{mobile manipulation} robot strictly follows its designated role of object transportation and passively waits for the \textit{mobile} robot to complete exploration, instead of proactively assisting in the collaborative search process.

\item \textbf{Spatial/Affordance Misbinding.}  
% Robots exhibit limited understanding of spatial positioning, reachability, and container–support relationships, resulting in inefficient interactions and delayed responses even under graspable conditions. For instance, a \textit{mobile manipulation} robot may need multiple base adjustments before a successful grasp or attempt an opening action before reaching interactable furniture; likewise, a \textit{manipulation} robot may delay grasping even when a target object lies within its reachable workspace.
Robots show limited understanding of spatial positioning, reachability, and container–support relations, leading to inefficient interactions and delayed responses even under graspable conditions. For example, a \textit{mobile manipulation} robot may require several base adjustments before grasping, or attempt to open furniture prematurely, while a \textit{manipulation} robot may hesitate despite the target being within reach.

\item \textbf{Object-Level Contention.}  
Multiple robots may simultaneously initiate actions on the same object, causing conflicts or behavioral reversals. For example, after a \textit{mobile manipulation} robot places an object within the \textit{manipulation} robot’s reachable range, it redundantly performs another pick action, directly conflicting with the manipulator’s grasping operation.

\item \textbf{Communication Inefficiency.}  
Although prompts are designed to encourage LLMs to generate \textit{communicate} actions only when necessary, inter-robot communication remains noisy. Robots frequently send redundant or repetitive messages with low information value, leading to inefficiencies in collaborative reasoning.  
\end{enumerate}

Through systematic observation of these experimental processes and detailed analysis of the identified failure modes, we aim to inspire future research toward more adaptive, grounded, and communication-efficient LLM-based heterogeneous multi-robot collaboration systems. Guided by these observations, we collect targeted datasets and, building on the collected data, further develop data-driven methods to address or mitigate the aforementioned issues.

\subsubsection{Case Study}
\begin{figure*}[htbp]
    \centering
    \includegraphics[width=\textwidth]{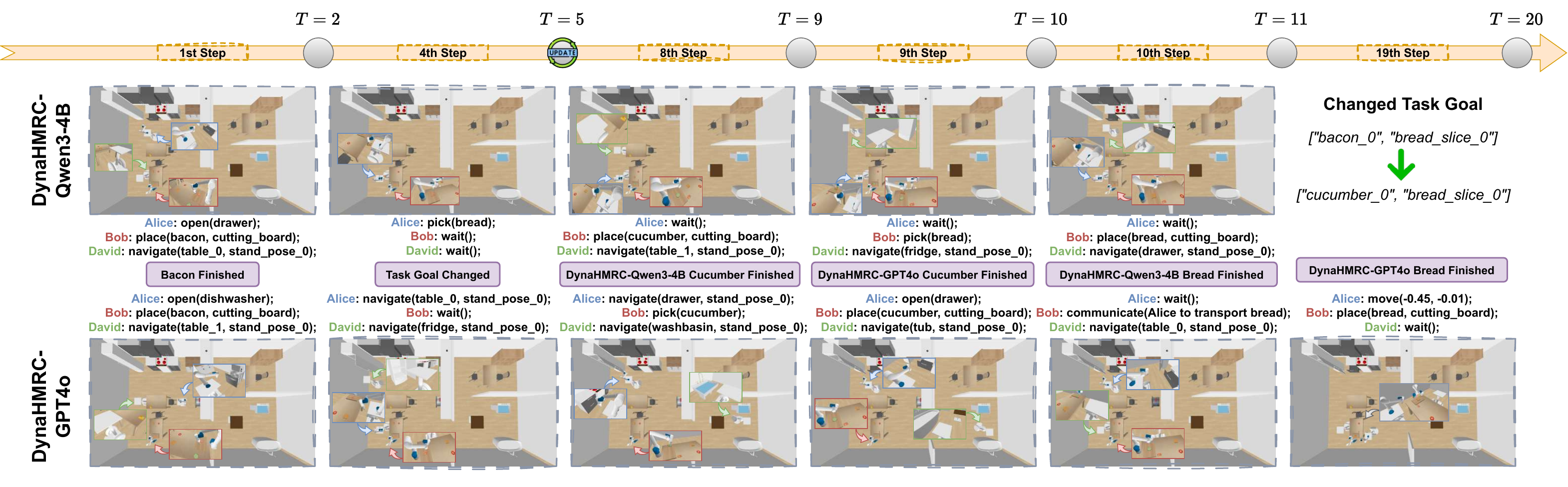}
    \caption{A comparative illustration of a successful experimental case using the Ma–MoMa–Mo configuration combination as an example. Key moments during the task execution are selected to showcase the system’s overall state at each stage. The regions highlighted with purple boxes provide supplementary explanations that clarify the significance of these specific moments.}
    \label{fig:case}
\end{figure*}

To better illustrate the task execution process, Fig.~\ref{fig:case} compares the performance of \sysnameo and \sysnameq on a CTO task. During this task, the goal changes at $T=5$ from [``bacon\_0", ``bread\_slice\_0"] to [``cucumber\_0", ``bread\_slice\_0"]. While both frameworks successfully complete the task, \sysnameq achieves it with higher efficiency. The supplementary video includes additional examples of task execution. In addition, we present a failure case study of \sysnameq in App\ref{lab:failure_case}.

\subsection{Experiments on Real Scenarios}
To demonstrate the efficiency and generalization capability of \sysname in real-world scenarios, we conduct comparative experiments using a ``Ma–MoMa–Mo'' robot configuration. The employed robots include the Galaxea R1 Lite (a mobile dual-arm manipulator), Agilex Tracer Mini (a differential-drive mobile base), and Agilex Piper (a six-degree-of-freedom tabletop manipulator). The sensor devices we used include the RealSense D435i camera, the RealSense D455 camera, and the Mid-360 LiDAR. We design distinct atomic action sets for each robot, encapsulating their specific physical capabilities. For the scene graph, we construct the furniture-level hierarchy following the ConceptGraphs~\cite{gu2024conceptgraphs, ye2026c, yip2026sope} pipeline and generate navigation anchors using SLAM-based pose estimates in the map coordinate frame, supplemented by manual corrections for errors and missing information.

In the real-robot system, for manipulation tasks, we employ AnyGrasp to predict grasping points and use Grounded SAM to filter candidate points according to semantic targets. The resulting grasp poses are then executed through inverse kinematics planning. For navigation tasks, we utilize a SLAM pipeline for mapping and localization, with A*~\cite{Kurzer1057261} as the global path planner and DWA~\cite{fox1997dynamic} as the local planner.

We further integrate the \textit{GPT-4o} multimodal model to obtain scene understanding and action feedback in real-world environments. Specifically, after each action is completed, the wrist camera on the manipulator captures the current scene to assess both action success and task progress. For example, after the mobile robot navigates to a new furniture location, \textit{GPT-4o} identifies the objects on it to facilitate exploration of unknown environments. Similarly, after each picking or placing action, \textit{GPT-4o} provides a judgmental response to verify whether the operation was executed successfully. In addition, feedback such as pose estimation and inverse kinematics results can also be incorporated into the evaluation process.

\begin{figure*}[htbp]
    \centering
    \includegraphics[width=\textwidth]{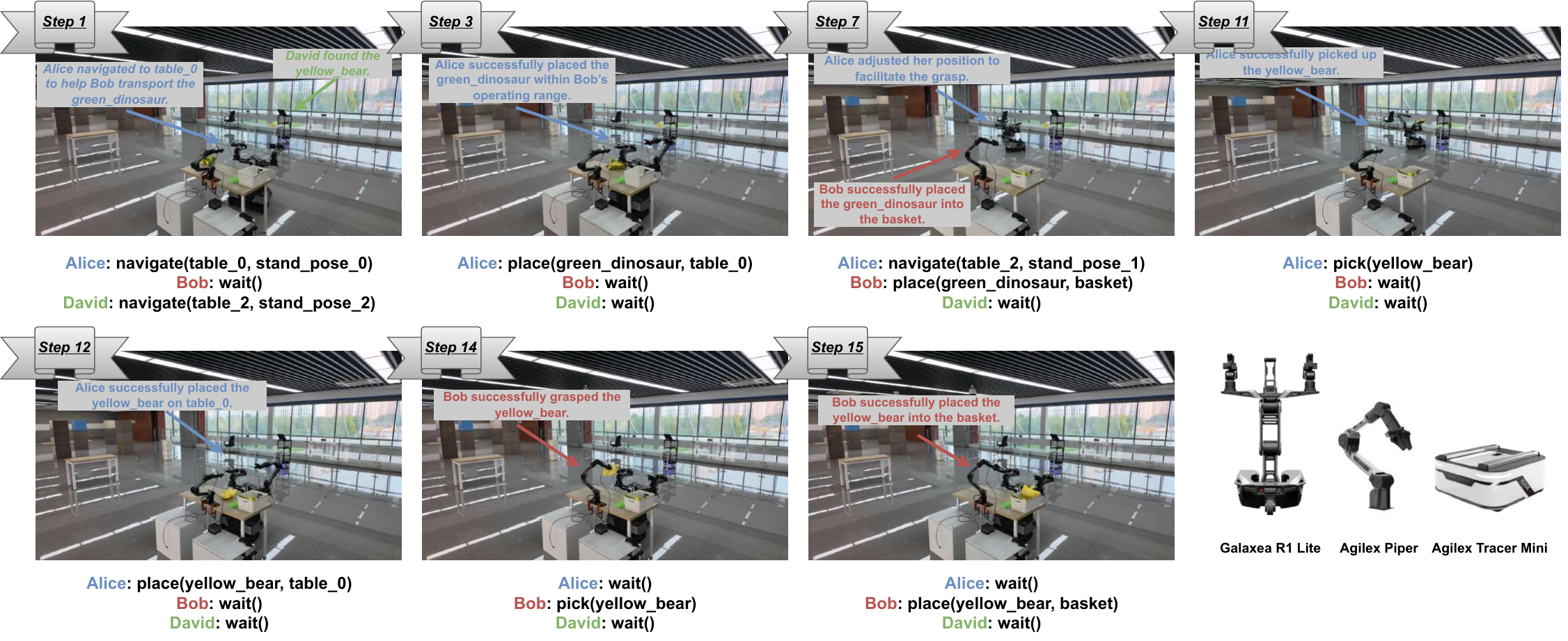}
    \caption{An example of a real-world experimental scenario is shown, where three robots collaborate in exploration and transportation tasks to place the green\_dinosaur and yellow\_bear into the basket. Several key moments of this process are illustrated in the figure. In App\ref{lab:fail_mode_breakdown}, we summarize the failure mode breakdown in real-world scenarios.}
    \label{fig:real}
\end{figure*}

We compare the performance of \textit{CoELA}, \sysnameo, and \sysnameq in the indoor scene shown in Fig.~\ref{fig:real}, with quantitative results summarized in Tab.~\ref{tab:real}. Each method is evaluated across eight trials under varying target object types, positions, and quantities, and the reported results represent the average performance over all experiments.
\sysnameq outperforms the baseline algorithms across all evaluation metrics. For additional demonstrations of real-world experimental performance, please refer to our supplementary video. We further report the local inference latency and API call latency of the real-world deployment in App\ref{lab:latency}.

\input{table/real}

%% file: table/constant.tex
\definecolor{easyblue}{RGB}{235,244,255}
\definecolor{mediumyellow}{RGB}{255,248,225}
\definecolor{hardgreen}{RGB}{238,248,238}
\begin{table}[!htbp] 

\centering

\caption{Evaluation Results with Varying Robot Types and Task Difficulties ($T=0$)} 

\label{tab:constant_exp} 

\renewcommand{\arraystretch}{1.0} 

\scriptsize
\setlength{\tabcolsep}{2pt}
\fitintolinewidth{ 

% \begin{tabular}{>{\centering\arraybackslash}p{2.8cm} 
%                 *{15}{>{\centering\arraybackslash}p{0.75cm}}} 

\begin{tabular}{
>{\centering\arraybackslash}p{3.0cm}
*{5}{>{\columncolor{easyblue}\centering\arraybackslash}p{0.75cm}}
*{5}{>{\columncolor{mediumyellow}\centering\arraybackslash}p{0.75cm}}
*{5}{>{\columncolor{hardgreen}\centering\arraybackslash}p{0.75cm}}
}

\toprule

% \multicolumn{16}{c}{\textbf{T=0}} \\ 
% \midrule

\multirow{2}{*}{\textbf{Method}} 
& \multicolumn{5}{>{\columncolor{easyblue}}c}{\textbf{Easy}} 
& \multicolumn{5}{>{\columncolor{mediumyellow}}c}{\textbf{Medium}} 
& \multicolumn{5}{>{\columncolor{hardgreen}}c}{\textbf{Hard}} \\ 
\cmidrule(lr){2-6} \cmidrule(lr){7-11} \cmidrule(lr){12-16} 
& SUCC$\uparrow$ & PS$\uparrow$ & TS$\downarrow$ & AS$\downarrow$ & CC$\downarrow$ 
& SUCC$\uparrow$ & PS$\uparrow$ & TS$\downarrow$ & AS$\downarrow$ & CC$\downarrow$ 
& SUCC$\uparrow$ & PS$\uparrow$ & TS$\downarrow$ & AS$\downarrow$ & CC$\downarrow$ \\ 
\midrule

% --- 2 types of robots --- 
\rowcolor[gray]{.7} \multicolumn{16}{c}{\textbf{2 Types of Robot}} \\ 

\rowcolor[gray]{.9} \multicolumn{16}{c}{Pack Objects} \\ 
MCTS-NoINS & 0.6111 & 0.6111 & 14.1667 & - & - & 0.0000 & 0.0185 & 30.0000 & - & - & 0.0000 & 0.0278 & 50.0000 & - & - \\ 
MCTS &\textbf{1.0000} &\textbf{1.0000} & 7.1667 & - & - & 0.6667 & 0.8333 & 22.6667 & - & - & \textbf{1.0000} & \textbf{1.0000} & \textbf{37.1667} & - & - \\ 
% LLM-MCTS & 1.0000 & 1.0000 & 4.6667 & - & - & 0.3333 & 0.7222 & 23.6667 & - & - & 0.8333 & 0.9583 & 34.3333 & - & - \\ 
CoELA & 0.8889 & 0.8889 & 6.7222 & 4.8889 & 0.6389 & 0.4444 & 0.7037 & 26.1111 & 17.4444 & 3.7222 & 0.0000 & 0.5556 & 50.0000 & 38.1667 & 7.9444 \\ 
\sysnameo &\textbf{1.0000} &\textbf{1.0000} & 5.7222 & 4.3333 & 1.0000 & \textbf{0.7222} & \textbf{0.8889} & 23.3888 & 16.4167 & 4.3333 & 0.2778 & 0.6833 & 45.2778 & 37.2500 & 10.4722 \\ 
\sysnameq &\textbf{1.0000} & \textbf{1.0000} & \textbf{4.6667} & \textbf{2.7666} & \textbf{0.2333} & \textbf{0.7222} & 0.8796 & \textbf{22.1111} & \textbf{13.3611} & \textbf{2.0000} & 0.5556 & 0.7917 & 43.0556 & \textbf{26.5556} & \textbf{3.8333} \\ 

\rowcolor[gray]{.9} \multicolumn{16}{c}{Make Sandwich} \\ 
MCTS-NoINS & 0.8889 & 0.8889 & 10.5000 & - & - & 0.0556 & 0.1667 & 29.3333 & - & - & 0.1667 & 0.2824 & 48.0000 & - & - \\ 
MCTS & \textbf{1.0000} & \textbf{1.0000} & 5.1667 & - & - & \textbf{1.0000} & \textbf{1.0000} & \textbf{12.6667} & - & - & \textbf{1.0000} & \textbf{1.0000} & \textbf{21.5000} & - & - \\ 
% LLM-MCTS & 1.0000 & 1.0000 & 6.5000 & - & - & 1.0000 & 1.0000 & 10.6667 & - & - & 1.0000 & 1.0000 & 19.0833 & - & - \\ 
CoELA & 0.6667 & 0.7222 & 8.0000 & 7.1111 & 0.5278 & 0.7778 & 0.8889 & 16.0000 & 12.5278 & \textbf{1.7222} & 0.6111 & 0.8056 & 29.8889 & \textbf{21.6111} & \textbf{2.6944} \\ 
\sysnameo & \textbf{1.0000} & \textbf{1.0000} & \textbf{2.7778} & \textbf{2.7778} & 0.1944 & 0.8333 & 0.8611 & 20.7222 & 15.8611 & 4.4444 & 0.6667 & 0.8056 & 31.7222 & 25.7778 & 7.8056 \\ 
\sysnameq & \textbf{1.0000} & \textbf{1.0000} & 3.0556 & 2.8611 & \textbf{0.1389} & 0.7222 & 0.7778 & 20.0000 & \textbf{11.7500} & 2.1667 & 0.4444 & 0.6296 & 37.0000 & 23.8611 & 3.8333 \\

\rowcolor[gray]{.9} \multicolumn{16}{c}{Sort Solids} \\ 
MCTS-NoINS & 0.3333 & 0.3333 & 15.9444 & - & - & 0.0556 & 0.2269 & 29.6111 & - & - & 0.0000 & 0.0556 & 50.0000 & - & - \\ 
MCTS & 0.5000 & 0.5000 & 13.0000 & - & - & 0.1667 & 0.2639 & 28.5000 & - & - & 0.5000 & 0.5556 & 31.8333 & - & - \\ 
% LLM-MCTS & 1.0000 & 1.0000 & 5.6667 & - & - & 1.0000 & 1.0000 & 10.1667 & - & - & 0.8333 & 0.9444 & 19.0000 & - & - \\ 
CoELA & \textbf{1.0000} & \textbf{1.0000} & \textbf{3.1667} & \textbf{3.1667} & 0.1111 & 0.8333 & 0.9583 & 10.6111 & 9.6667 & 0.6667 & 0.9444 & 0.9815 & 14.6667 & 12.1667 & 1.3056 \\ 
\sysnameo & \textbf{1.0000} & \textbf{1.0000} & 4.0556 & 3.8056 & 0.4167 & 0.8333 & 0.9352 & 13.1111 & 11.7222 & 2.5278 & 0.9444 & 0.9722 & 9.6111 & 8.4722 &  1.3056 \\ 
\sysnameq & \textbf{1.0000} & \textbf{1.0000} & 3.2222 & 3.1945 & \textbf{0.0555} & \textbf{1.0000} & \textbf{1.0000} & \textbf{6.6111} & \textbf{6.6111} & \textbf{0.0556} & \textbf{0.9745} & \textbf{0.9859} & \textbf{7.6667} & \textbf{6.4825} &  \textbf{0.8321} \\ 

% --- 3 types of robots --- 
\rowcolor[gray]{.7} \multicolumn{16}{c}{\textbf{3 Types of Robot}} \\ 

\rowcolor[gray]{.9} \multicolumn{16}{c}{Pack Objects} \\ 
MCTS-NoINS & 0.2963 & 0.2963 & 18.4815 & - & - & 0.0000 & 0.0000 & 30.0000 & - & - & 0.0000 & 0.0093 & 50.0000 & - & - \\ 
MCTS & 0.7778 & 0.7778 & 10.1111 & - & - & 0.2222 & 0.2593 & 28.5556 & - & - & 0.1111 & 0.3333 & 49.3333 & - & - \\ 
% LLM-MCTS & 1.0000 & 1.0000 & 4.4444 & - & - & 0.3333 & 0.6296 & 25.7778 & - & - & 0.2222 & 0.4815 & 45.6667 & - & - \\ 
CoELA & 0.8889 & 0.8889 & 6.3333 & 4.9383 & 0.7901 & 0.3704 & 0.5679 & 26.2593 & 20.1605 & 3.2716 & 0.2593 & 0.5586 & 46.2222 & 34.6049 & 5.7407 \\ 
\sysnameo & 0.8889 & 0.9630 & 7.1481 & 5.2469 & 0.9136 & 0.5556 & 0.7346 & 24.2222 & 19.1481 & 6.0000 & 0.2963 & 0.7611 & 43.5556 & 36.7901 & 11.9136 \\ 
\sysnameq & \textbf{0.9259} & \textbf{0.9259} & \textbf{6.2222} & \textbf{4.6049} & \textbf{0.5802} & \textbf{0.6071} & \textbf{0.8086} & \textbf{23.2592} & \textbf{15.1543} & \textbf{2.4444} & \textbf{0.5357} & \textbf{0.8148} & \textbf{38.3333} & \textbf{26.7006} & \textbf{3.7160} \\ 

\rowcolor[gray]{.9} \multicolumn{16}{c}{Make Sandwich} \\ 
MCTS-NoINS & 0.4444 & 0.4444 & 15.4815 & - & - & 0.0370 & 0.1049 & 29.5556 & - & - & 0.1111 & 0.1642 & 48.6296 & - & - \\ 
MCTS & 0.7778 & 0.8889 & 10.5556 & - & - & 0.2222 & 0.3148 & 26.0000 & - & - & 0.3333 & 0.4519 & 39.5556 & - & - \\ 
% LLM-MCTS & 1.0000 & 1.0000 & 6.3889 & - & - & 1.0000 & 1.0000 & 12.9444 & - & - & 0.6667 & 0.7111 & 31.5000 & - & - \\ 
CoELA & 0.6667 & 0.7037 & 8.0000 & 6.9877 & 0.8395 & \textbf{0.8889} & 0.9383 & \textbf{17.2593} & \textbf{13.7531} & 2.6667 & \textbf{0.6296} & \textbf{0.7728} & 30.6296 & 26.7160 & \textbf{4.3580} \\ 
\sysnameo & \textbf{1.0000} & \textbf{1.0000} & 2.9630 & 2.9012 & 0.3086 & \textbf{0.8889} & \textbf{0.9506} & 19.5556 & 15.7531 & 4.3580 & \textbf{0.6296} & 0.7494 & \textbf{30.5556} & \textbf{25.3210} & 7.1728 \\ 
\sysnameq & \textbf{1.0000} & \textbf{1.0000} & \textbf{2.6667} & \textbf{2.6420} & \textbf{0.0370} & 0.7778 & 0.8580 & 19.5926 & 13.7778 & \textbf{2.1605} & 0.5556 & 0.7228 & 38.0741 & 28.3210 & 4.6173 \\ 

\rowcolor[gray]{.9} \multicolumn{16}{c}{Sort Solids} \\ 
MCTS-NoINS & 0.3333 & 0.3704 & 16.9259 & - & - & 0.0000 & 0.1019 & 30.0000 & - & - & 0.0370 & 0.1543 & 48.8148 & - & - \\ 
MCTS & 0.6667 & 0.6667 & 13.2222 & - & - & 0.1111 & 0.2130 & 28.5556 & - & - & 0.3333 & 0.3981 & 44.2222 & - & - \\ 
% LLM-MCTS & 1.0000 & 1.0000 & 5.1111 & - & - & 0.4444 & 0.6389 & 21.5556 & - & - & 0.5556 & 0.8148 & 40.1111 & - & - \\ 
CoELA & \textbf{1.0000} & \textbf{1.0000} & \textbf{3.3333} & \textbf{3.2593} & 0.2716 & \textbf{0.9630} & 0.9877 & 9.2593 & 8.6790 & 1.0123 & 0.8889 & 0.9630 & 16.8519 & 15.1235 & 2.0864 \\ 
\sysnameo & \textbf{1.0000} & \textbf{1.0000} & 5.0000 & 4.5556 & 0.8271 & \textbf{0.9630} & \textbf{0.9907} & 11.9630 & 11.0123 & 2.3704 & 0.9200 & 0.9815 & 16.4074 & 14.1852 & 4.0247 \\ 
\sysnameq & \textbf{1.0000} & \textbf{1.0000} & 4.5556 & 4.1481 & \textbf{0.1950} & 0.9615 & 0.9877 & \textbf{8.5185} & \textbf{8.0062} & \textbf{0.2778} & \textbf{0.9630} & \textbf{0.9907} & \textbf{14.9629} & \textbf{11.8642} & \textbf{1.4444} \\ 

% --- 4 types of robots --- 
\rowcolor[gray]{.7} \multicolumn{16}{c}{\textbf{4 Types of Robot}} \\ 

\rowcolor[gray]{.9} \multicolumn{16}{c}{Pack Objects} \\ 
MCTS-NoINS & 0.0000 & 0.0000 & 20.0000 & - & - & 0.0000 & 0.0000 & 30.0000 & - & - & 0.0000 & 0.0000 & 50.0000 & - & - \\ 
MCTS & 0.6667 & 0.6667 & 12.3333 & - & - & 0.0000 & 0.0000 & 30.0000 & - & - & 0.0000 & 0.0000 & 50.0000 & - & - \\ 
% LLM-MCTS & 1.0000 & 1.0000 & 3.6667 & - & - & 0.3333 & 0.5000 & 25.0000 & - & - & 0.0000 & 0.3889 & 50.0000 & - & - \\ 
CoELA & 0.7778 & 0.7778 & 7.5556 & 6.6667 & 0.9167 & 0.6667 & 0.8704 & 20.2222 & 16.7500 & 2.4444 & 0.4444 & 0.6667 & 45.5556 & 36.5833 & 4.5000 \\ 
\sysnameo & \textbf{1.0000} & \textbf{1.0000} & 7.0000 & 5.6111 & 0.6389 & 0.6667 & 0.8704 & 22.0000 & 18.7222 & 4.9722 & 0.5556 & 0.8333 & 39.0000 & 34.6111 & 8.9167 \\ 
\sysnameq & \textbf{1.0000} & \textbf{1.0000} & \textbf{4.5556} & \textbf{3.6667} & \textbf{0.5556} & \textbf{0.8889} & \textbf{0.9630} & \textbf{17.2222} & \textbf{12.6389} & \textbf{1.9722} & \textbf{0.9167} & \textbf{0.9722} & \textbf{23.1667} & \textbf{16.6528} & \textbf{2.0278} \\ 

\rowcolor[gray]{.9} \multicolumn{16}{c}{Make Sandwich} \\ 
MCTS-NoINS & 0.5556 & 0.5556 & 14.6667 & - & - & 0.0000 & 0.1111 & 30.0000 & - & - & 0.0000 & 0.0000 & 50.0000 & - & - \\ 
MCTS & 0.3333 & 0.3333 & 15.3333 & - & - & 0.0000 & 0.0000 & 30.0000 & - & - & 0.0000 & 0.1111 & 50.0000 & - & - \\ 
% LLM-MCTS & 1.0000 & 1.0000 & 6.5000 & - & - & 1.0000 & 1.0000 & 14.0000 & - & - & 0.8333 & 0.8333 & 29.0000 & - & - \\ 
CoELA & 0.6667 & 0.7222 & 8.0000 & 7.1667 & 0.7222 & 0.8889 & 0.9444 & 14.8889 & 12.9444 & 2.1667 & 0.6667 & 0.7222 & 28.3333 & 26.8333 & 3.3333 \\ 
\sysnameo & \textbf{1.0000} & \textbf{1.0000} & \textbf{2.7776} & \textbf{2.7500} & 0.1944 & 0.9259 & 0.9630 & 16.2296 & 13.5991 & 3.4264 & 0.6250 & 0.7667 & 34.4000 & 28.3250 & 7.5500 \\ 
\sysnameq & \textbf{1.0000} & \textbf{1.0000} & 3.8889 & 3.5000 & \textbf{0.1111} & \textbf{1.0000} & \textbf{1.0000} & \textbf{12.7778} & \textbf{10.2500} & \textbf{1.1389} & \textbf{0.7778} & \textbf{0.8333} & \textbf{26.4000} & \textbf{22.3250} & \textbf{2.1500} \\ 

\rowcolor[gray]{.9} \multicolumn{16}{c}{Sort Solids} \\ 
MCTS-NoINS & 0.0000 & 0.0556 & 20.0000 & - & - & 0.0000 & 0.0370 & 30.0000 & - & - & 0.0000 & 0.0815 & 50.0000 & - & - \\ 
MCTS & 0.0000 & 0.0000 & 20.0000 & - & - & 0.0000 & 0.0000 & 30.0000 & - & - & 0.0000 & 0.0667 & 50.0000 & - & - \\ 
% LLM-MCTS & 0.6667 & 0.6667 & 13.0000 & - & - & 0.3333 & 0.6111 & 23.6667 & - & - & 0.0000 & 0.3722 & 50.0000 & - & - \\ 
CoELA & \textbf{1.0000} & \textbf{1.0000} & 4.5556 & 4.4167 & 0.5278 & 0.8889 & 0.9630 & \textbf{11.6667} & 10.9444 & 1.4444 & 0.7778 & 0.9407 & 21.3333 & 18.9722 & \textbf{2.6111} \\ 
\sysnameo & \textbf{1.0000} & \textbf{1.0000} & 4.8889 & 4.7778 & 0.5833 & \textbf{1.0000} & \textbf{1.0000} & 16.2222 & 14.2500 & 3.2778 & \textbf{1.0000} & \textbf{1.0000} & \textbf{20.2222} & \textbf{18.3333} & 4.4722 \\ 
\sysnameq & \textbf{1.0000} & \textbf{1.0000} & \textbf{4.0634} & \textbf{3.8626} & \textbf{0.2584} & \textbf{1.0000} & \textbf{1.0000} & 12.1111 & \textbf{10.4444} & \textbf{1.3889} & 0.9444 & 0.9630 & 28.3333 & 23.6667 & 3.4167 \\ 

\bottomrule

\end{tabular}} 
\end{table}

%% file: table/hyperparameters.tex
\begin{table}[htbp]
\centering
\caption{Hyperparameters}
\label{tab:hyper}
\renewcommand{\arraystretch}{1.08}
\small
\begin{tabularx}{0.68\textwidth}{>{\centering\arraybackslash}X 
                >{\centering\arraybackslash}p{0.18\textwidth} 
                }
\toprule
\textbf{Hyperparameter} & \textbf{Value} \\
\midrule
\rowcolor[gray]{.9} \multicolumn{2}{c}{Hyperparameters for Qwen3-4B Full Fine-tuning} \\
Training epochs & $6$ \\
Learning rate & $10^{-5}$ \\
LR scheduler type & Cosine \\
Warmup ratio & $0.1$ \\
Batch size & $8$ \\
Precision & BF16 \\
Cutoff length & $2048$ \\
Gradient accumulation steps & $2$ \\
DeepSpeed & $ds\_z3$ \\

\bottomrule
\end{tabularx}
\end{table}

%% file: table/goal_change.tex
\begingroup
\small
\setlength{\tabcolsep}{3pt}
\renewcommand{\arraystretch}{1.0}
\begin{longtable}{>{\centering\arraybackslash}p{4.0cm}
                *{5}{>{\centering\arraybackslash}p{1.5cm}}}
\caption{Experimental Results of Changing the Task Objectives ($T=5$)}
\label{tab:goal_exp}\\
\toprule
\multirow{2}{*}{\textbf{Method}}
& \multicolumn{5}{c}{\textbf{CTO}} \\
\cmidrule(lr){2-6}
& SUCC$\uparrow$ & PS$\uparrow$ & TS$\downarrow$ & AS$\downarrow$ & CC$\downarrow$ \\
\midrule
\endfirsthead

\caption[]{Experimental Results of Changing the Task Objectives ($T=5$), continued}\\
\toprule
\multirow{2}{*}{\textbf{Method}}
& \multicolumn{5}{c}{\textbf{CTO}} \\
\cmidrule(lr){2-6}
& SUCC$\uparrow$ & PS$\uparrow$ & TS$\downarrow$ & AS$\downarrow$ & CC$\downarrow$ \\
\midrule
\endhead

\midrule
\multicolumn{6}{r}{\small Continued on next page}\\
\endfoot

\bottomrule
\endlastfoot

% --- 2 types of robots --- 

\rowcolor[gray]{.8} \multicolumn{6}{c}{\textbf{2 Types of Robot}} \\*
\rowcolor[gray]{.9} \multicolumn{6}{c}{Pack Objects} \\*
MCTS &0.6111 &0.7870 &24.8333 &- &- \\ 
% CONHERENT &0.5000 &0.7871 &21.6112 &14.0000 &- \\

% LLM-MCTS &0.7222 &0.8519 &21.89 &- &- \\ 

COELA &0.5000 &0.7130 &25.1100 &17.6700 &3.2200 \\ 

\sysnameo &0.6316 &\textbf{0.9087} &\textbf{21.1984} &17.5040 &5.9702 \\ 

\sysnameq &\textbf{0.7222} &0.9074 &25.6667 &\textbf{15.0556} &\textbf{3.0000} \\ 

\rowcolor[gray]{.9} \multicolumn{6}{c}{Make Sandwich} \\*
MCTS &\textbf{1.0000} &\textbf{1.0000} &\textbf{12.3333} &- &- \\ 
% CONHERENT &0.4445 &0.5463 &20.3334 &6.5000 &- \\

% LLM-MCTS &1.0000 &1.0000 &12.56 &- &- \\ 

COELA &0.9444 &0.9444 &14.8300 &12.7800 &1.6100 \\ 

\sysnameo &0.9630 &0.9815 &14.2963 &12.5556 &\textbf{1.0833} \\ 

\sysnameq &0.6111 &0.6852 &17.3889 &\textbf{12.5000} &2.1944 \\ 

\rowcolor[gray]{.9} \multicolumn{6}{c}{Sort Solids} \\*
MCTS &0.0556 &0.2454 &29.5556 &- &- \\ 
% CONHERENT &0.8889 &0.9630 &13.1112 &10.4445 &- \\

% LLM-MCTS &0.8889 &0.9676 &15.17 &- &- \\ 

COELA &0.9444 &0.9815 &\textbf{9.4400} &\textbf{8.6100} &0.6100 \\ 

\sysnameo &\textbf{1.0000} &\textbf{1.0000} &12.7222 &11.4722 &2.4722 \\ 

\sysnameq &0.9444 &0.9815 &10.0556 &9.3889 &\textbf{0.4444} \\ 

% --- 3 types of robots --- 

\rowcolor[gray]{.8} \multicolumn{6}{c}{\textbf{3 Types of Robot}} \\*
\rowcolor[gray]{.9} \multicolumn{6}{c}{Pack Objects} \\*
MCTS &0.2593 &0.3642 &28.3333 &- &- \\ 
% CONHERENT &0.3333 &0.5432 &24.7778 &16.6914 &- \\

% LLM-MCTS &0.3704 &0.6173 &26.67 &- &- \\ 

COELA &0.4074 &0.6481 &26.3300 &20.8600 &\textbf{3.4000} \\ 

\sysnameo &\textbf{0.4815} &0.7407 &\textbf{24.5556} &19.1975 &6.0000 \\ 

\sysnameq &\textbf{0.4815} &\textbf{0.7778} &25.6296 &\textbf{17.2716} &3.4691 \\ 

\rowcolor[gray]{.9} \multicolumn{6}{c}{Make Sandwich} \\*
MCTS &0.3704 &0.4012 &24.4444 &- &- \\ 
% CONHERENT &0.2593 &0.3148 &25.7037 &6.9012 &- \\

% LLM-MCTS &1.0000 &1.0000 &13.85 &- &- \\ 

COELA &0.7778 &0.8333 &\textbf{18.6700} &16.4100 &2.4900 \\ 

\sysnameo &\textbf{0.8059} &\textbf{0.9076} &18.5701 &15.6532 &4.1887 \\ 

\sysnameq &0.6667 &0.6914 &18.7037 &\textbf{13.3827} &\textbf{2.2346} \\ 

\rowcolor[gray]{.9} \multicolumn{6}{c}{Sort Solids} \\*
MCTS &0.0000 &0.0463 &30.0000 &- &- \\ 
% CONHERENT &0.4815 &0.8426 &22.5185 &14.8519 &- \\

% LLM-MCTS &0.4815 &0.8302 &24.37 &- &- \\ 

COELA &0.9259 &\textbf{0.9815} &15.9600 &14.1600 &1.7700 \\ 

\sysnameo &\textbf{0.9444} &0.9630 &15.9444 &14.2037 &3.4259 \\ 

\sysnameq &0.9259 &\textbf{0.9784} &\textbf{13.9259} &\textbf{12.0494} &\textbf{1.0493} \\

% --- 4 types of robots --- 

\rowcolor[gray]{.8} \multicolumn{6}{c}{\textbf{4 Types of Robot}} \\*
\rowcolor[gray]{.9} \multicolumn{6}{c}{Pack Objects} \\*
MCTS &0.0000 &0.0000 &30.0000 &- &- \\ 
% CONHERENT &0.4444 &0.6667 &26.2222 &11.0000 &- \\

% LLM-MCTS &0.4444 &0.6667 &25.11 &- &- \\ 

COELA &0.6667 &0.7222 &20.4400 &17.5000 &3.1100 \\ 

\sysnameo &0.7778 &0.8333 &20.0000 &16.9167 &3.7222 \\ 

\sysnameq &\textbf{1.0000} &\textbf{1.0000} &\textbf{16.3889} &\textbf{12.2917} &\textbf{1.7292} \\ 

\rowcolor[gray]{.9} \multicolumn{6}{c}{Make Sandwich} \\*
MCTS &0.4444 &0.5000 &23.1111 &- &- \\ 
% CONHERENT &0.5556 &0.6111 &19.8889 &6.8611 &- \\

% LLM-MCTS &1.0000 &1.0000 &13.56 &- &- \\ 

COELA &0.8889 &0.9444 &\textbf{14.3300} &\textbf{13.0800} &1.7200 \\ 

\sysnameo &\textbf{0.9074} &\textbf{0.9630} &19.3333 &18.2500 &\textbf{1.1667} \\ 

\sysnameq &0.8889 &0.9444 &18.4444 &14.5556 &1.9722 \\ 

\rowcolor[gray]{.9} \multicolumn{6}{c}{Sort Solids} \\*
MCTS &0.0000 &0.0722 &30.0000 &- &- \\ 
% CONHERENT &0.6667 &0.9167 &19.2222 &11.4722 &- \\

% LLM-MCTS &0.3333 &0.6389 &24.78 &- &- \\ 

COELA &\textbf{1.0000} &\textbf{1.0000} &\textbf{12.2200} &\textbf{11.5600} &1.3300 \\ 

\sysnameo &\textbf{1.0000} &\textbf{1.0000} &17.0000 &14.6250 &3.5417 \\ 

\sysnameq &\textbf{1.0000} &\textbf{1.0000} &13.8889 &12.0833 &\textbf{1.0833} \\ 

\end{longtable}
\endgroup

%% file: table/area_change.tex
\begingroup
\small
\setlength{\tabcolsep}{3pt}
\renewcommand{\arraystretch}{1.0}
\begin{longtable}{>{\centering\arraybackslash}p{4.0cm}
                *{5}{>{\centering\arraybackslash}p{1.5cm}}}
\caption{Experimental Results of Introducing Restricted Areas ($T=5$)}
\label{tab:area_exp}\\
\toprule
\multirow{2}{*}{\textbf{Method}}
& \multicolumn{5}{c}{\textbf{IRZ}} \\
\cmidrule(lr){2-6}
& SUCC$\uparrow$ & PS$\uparrow$ & TS$\downarrow$ & AS$\downarrow$ & CC$\downarrow$ \\
\midrule
\endfirsthead

\caption[]{Experimental Results of Introducing Restricted Areas ($T=5$), continued}\\
\toprule
\multirow{2}{*}{\textbf{Method}}
& \multicolumn{5}{c}{\textbf{IRZ}} \\
\cmidrule(lr){2-6}
& SUCC$\uparrow$ & PS$\uparrow$ & TS$\downarrow$ & AS$\downarrow$ & CC$\downarrow$ \\
\midrule
\endhead

\midrule
\multicolumn{6}{r}{\small Continued on next page}\\
\endfoot

\bottomrule
\endlastfoot

% --- 3 types of robots --- 
\rowcolor[gray]{.8} \multicolumn{6}{c}{\textbf{3 Types of Robot}} \\*
\rowcolor[gray]{.9} \multicolumn{6}{c}{Pack Objects} \\*
MCTS & 0.1852 & 0.2284 & 28.4815 & - & - \\ 
% CONHERENT & 0.3704 & 0.6543 & 25.2963 & 16.3457 & - \\ 
% LLM-MCTS &0.3333 &0.6173 &25.8148 & - & - \\ 
COELA &0.3333 &0.5864 &24.7037 &19.7901 &3.2222 \\ 
\sysnameo &0.7619 &0.8889 &21.0926 &16.8704 &4.9506 \\ 
\sysnameq &\textbf{0.9630} &\textbf{0.9877} &\textbf{17.7407} &\textbf{12.6543} &\textbf{1.6173} \\ 

\rowcolor[gray]{.9} \multicolumn{6}{c}{Make Sandwich} \\*
MCTS & 0.4074 & 0.4815 & 25.5185 & - & - \\ 
% CONHERENT & 0.5556 & 0.6605 & 19.5926 & 11.0494 & - \\ 
% LLM-MCTS &0.9630 &0.9630 &13.5185 & - & - \\ 
COELA &0.8889 &0.9506 &\textbf{16.0370} &13.2840 &2.2469 \\ 
\sysnameo &0.9167 &0.9630 &25.2500 &22.0833 &4.5556 \\ 
\sysnameq &\textbf{0.9259} &\textbf{0.9691} &17.1852 &\textbf{12.5679} &\textbf{2.0864} \\ 

\rowcolor[gray]{.9} \multicolumn{6}{c}{Sort Solids} \\*
MCTS & 0.0370 & 0.0802 & 29.4074 & - & - \\ 
% CONHERENT & 0.9259 & 0.9784 & 10.6667 & 7.2469 & - \\ 
% LLM-MCTS &0.7778 &0.9074 &16.5556 & - & - \\ 
COELA &0.9630 &0.9877 &8.5926 &7.8889 &0.9136 \\ 
\sysnameo &0.6667 &0.8750 &18.7778 &16.9815 &5.2778 \\ 
\sysnameq &\textbf{1.0000} &\textbf{1.0000} &\textbf{7.0370} &\textbf{6.9136} &\textbf{0.1605} \\ 

% --- 4 types of robots --- 
\rowcolor[gray]{.8} \multicolumn{6}{c}{\textbf{4 Types of Robot}} \\*
\rowcolor[gray]{.9} \multicolumn{6}{c}{Pack Objects} \\*
MCTS & 0.1111 & 0.1111 & 30.0000 & - & - \\ 
% CONHERENT & 0.6667 & 0.8148 & 18.0000 & 10.7778 & - \\ 
% LLM-MCTS &0.4444 &0.6111 &24.1111 & - & - \\ 
COELA &0.5556 &0.6667 &25.0000 &20.2222 &3.1944 \\ 
\sysnameo &\textbf{1.0000} &\textbf{1.0000} &17.2778 &13.8472 &2.9167 \\
\sysnameq &\textbf{1.0000} &\textbf{1.0000} &\textbf{15.2222} &\textbf{11.8056} &\textbf{1.2500} \\ 

\rowcolor[gray]{.9} \multicolumn{6}{c}{Make Sandwich} \\*
MCTS & 0.0000 & 0.1111 & 30.0000 & - & - \\ 
% CONHERENT & 0.8889 & 0.9444 & 12.3333 & 7.3056 & - \\ 
% LLM-MCTS &1.0000 &1.0000 &11.4444 & - & - \\ 
COELA &\textbf{1.0000} &\textbf{1.0000} &\textbf{12.6667} &\textbf{10.6389} &\textbf{1.3611} \\ 
\sysnameo &\textbf{1.0000} &\textbf{1.0000} &22.8333 &19.5926 &3.8148 \\ 
\sysnameq &\textbf{1.0000} &\textbf{1.0000} &16.1111 &12.8611 &1.8611 \\ 

\rowcolor[gray]{.9} \multicolumn{6}{c}{Sort Solids} \\*
MCTS & 0.0000 & 0.0000 & 30.0000 & - & - \\ 
% CONHERENT & 0.6667 & 0.8981 & 17.5556 & 9.4722 & - \\ 
% LLM-MCTS &1.0000 &1.0000 &12.0000 & - & - \\ 
COELA &0.8889 &0.9630 &10.3333 &9.7778 &1.3611 \\ 
\sysnameo &0.7500 &0.9167 &14.9167 &13.7500 &3.8542 \\ 
\sysnameq &\textbf{1.0000} &\textbf{1.0000} &\textbf{7.5556} &\textbf{7.2500} &\textbf{0.2778} \\ 

\end{longtable}
\endgroup

%% file: table/addition_change.tex
\begingroup
\small
\setlength{\tabcolsep}{3pt}
\renewcommand{\arraystretch}{0.86}
\begin{longtable}{>{\centering\arraybackslash}p{4.0cm}
                *{5}{>{\centering\arraybackslash}p{1.5cm}}}
\caption{Experimental results of adding new collaborators and removing existing collaborators ($T=5$)}
\label{tab:addition_exp}\\
\toprule
\multirow{2}{*}{\textbf{Method}}
& \multicolumn{5}{c}{\textbf{ANC}} \\
\cmidrule(lr){2-6}
& SUCC$\uparrow$ & PS$\uparrow$ & TS$\downarrow$ & AS$\downarrow$ & CC$\downarrow$ \\
\midrule
\endfirsthead

\caption[]{Experimental results of adding new collaborators and removing existing collaborators ($T=5$), continued}\\
\toprule
\multirow{2}{*}{\textbf{Method}}
& \multicolumn{5}{c}{\textbf{ANC}} \\
\cmidrule(lr){2-6}
& SUCC$\uparrow$ & PS$\uparrow$ & TS$\downarrow$ & AS$\downarrow$ & CC$\downarrow$ \\
\midrule
\endhead

\midrule
\multicolumn{6}{r}{\small Continued on next page}\\
\endfoot

\bottomrule
\endlastfoot
\rowcolor[gray]{.8} \multicolumn{6}{c}{\textbf{Ma-MoMa}\boldmath$\mathbf{\;\rightarrow\;}$\textbf{Ma-MoMa-UAV}} \\*
\rowcolor[gray]{.9} \multicolumn{6}{c}{Pack Objects} \\*
MCTS & 0.0000 & 0.1852 & 30.0000 & - & - \\
% CONHERENT & 0.5556 & 0.7963 & 26.3333 & 16.7778 & 0.0000 \\
% LLM-MCTS & & & & & & & & & & \\
CoELA & 0.5556 & 0.7778 & 23.5556 & 17.5000 & 3.2778 \\
\sysnameo &\textbf{1.0000} &\textbf{1.0000} &\textbf{20.0000} &15.1667 &3.5000 \\
\sysnameq &0.6667 &0.8333 &24.5556 &\textbf{14.9444} &\textbf{2.9444} \\

\rowcolor[gray]{.9} \multicolumn{6}{c}{Make Sandwich} \\*
MCTS & \textbf{1.0000} & \textbf{1.0000} & \textbf{12.4444} & - & - \\
% CONHERENT & 1.0000 & 1.0000 & 10.3333 & 6.7778 & 0.0000 \\
% LLM-MCTS & & & & & & & & & & \\
CoELA & \textbf{1.0000} & \textbf{1.0000} & 14.8889 & 11.5000 & 1.2778 \\
\sysnameo &0.6667 &0.7500 &23.1667 &18.7500 &6.5833 \\
\sysnameq &\textbf{1.0000} &\textbf{1.0000} &17.1111 &\textbf{11.3889} &\textbf{1.1000} \\

\rowcolor[gray]{.9} \multicolumn{6}{c}{Sort Solids} \\*
MCTS & 0.1111 & 0.1944 & 28.8889 & - & - \\
% CONHERENT & 0.8889 & 0.9722 & 11.0000 & 9.2778 & 0.0000 \\
% LLM-MCTS & & & & & & & & & & \\
CoELA & 0.8889 & 0.9630 & 10.6667 & 9.7778 & 0.8889 \\
\sysnameo &\textbf{1.0000} &\textbf{1.0000} &\textbf{9.3333} &9.5833 &1.1667 \\
\sysnameq &\textbf{1.0000} &\textbf{1.0000} &9.7778 &\textbf{8.1111} &\textbf{0.5000} \\

\rowcolor[gray]{.8} \multicolumn{6}{c}{\textbf{Ma-UAV}\boldmath$\mathbf{\;\rightarrow\;}$\textbf{Ma-MoMa-UAV}} \\*
\rowcolor[gray]{.9} \multicolumn{6}{c}{Pack Objects} \\*
MCTS & 0.3333 & 0.4259 & 25.7778 & - & - \\
% CONHERENT & 0.0000 & 0.3889 & 30.0000 & 17.2222 & 0.0000 \\
% LLM-MCTS & & & & & & & & & & \\
CoELA & 0.1111 & 0.4815 & 29.4444 & 22.2778 & 6.3333 \\
\sysnameo &\textbf{0.8333} &\textbf{0.9444} &\textbf{25.1667} &18.6667 &7.0833 \\
\sysnameq &0.3333 &0.6481 &26.0000 &\textbf{15.6111} &\textbf{4.5556}\\

\rowcolor[gray]{.9} \multicolumn{6}{c}{Make Sandwich} \\*
MCTS & 0.7778 & 0.7778 & \textbf{16.4444} & - & - \\
% CONHERENT & 0.7778 & 0.7778 & 14.7778 & 9.5000 & 0.0000 \\
% LLM-MCTS & & & & & & & & & & \\
CoELA & 0.8889 & 0.9444 & 19.2222 & 14.5000 & \textbf{2.9444} \\
\sysnameo &0.5000 &0.7500 &25.6667 &21.5000 &8.5833  \\
\sysnameq &\textbf{0.9259} &\textbf{0.9630} &18.2222 &\textbf{11.8889} &3.6111 \\

\rowcolor[gray]{.9} \multicolumn{6}{c}{Sort Solids} \\*
MCTS & 0.0000 & 0.1111 & 30.0000 & - & - \\
% CONHERENT & 1.0000 & 1.0000 & 9.1111 & 6.9444 & 0.0000 \\
% LLM-MCTS & & & & & & & & & & \\
CoELA & \textbf{1.0000} & \textbf{1.0000} & 7.6667 & 7.3333 & 0.7778 \\
\sysnameo &0.5556 &0.8148 &24.0000 &19.0556 &6.8889 \\
\sysnameq &\textbf{1.0000} &\textbf{1.0000} &\textbf{6.7778} &\textbf{6.5000} &\textbf{0.3889} \\

\rowcolor[gray]{.8} \multicolumn{6}{c}{\textbf{Ma-MoMa-Mo}\boldmath$\mathbf{\;\rightarrow\;}$\textbf{Ma-MoMa-Mo-UAV}} \\*
\rowcolor[gray]{.9} \multicolumn{6}{c}{Pack Objects} \\*
MCTS & 0.0000 & 0.0556 & 30.0000 & - & - \\
% CONHERENT & 0.4444 & 0.7593 & 26.2222 & 18.5185 & 0.0000 \\
% LLM-MCTS & & & & & & & & & & \\
CoELA & 0.7778 & 0.9074 & 21.4444 & 17.1481 & 3.0370 \\
\sysnameo &\textbf{0.8750} &\textbf{0.9583} &24.2500 &19.7083 &6.2083 \\
\sysnameq &0.7778 &0.8889 &\textbf{19.7778} &\textbf{14.9259} &\textbf{2.2593} \\

\rowcolor[gray]{.9} \multicolumn{6}{c}{Make Sandwich} \\*
MCTS & 0.4444 & 0.5556 & 22.0000 & - & - \\
% CONHERENT & 0.5556 & 0.6667 & 20.2222 & 13.8889 & 0.0000 \\
% LLM-MCTS & & & & & & & & & & \\
CoELA & 0.7778 & 0.8889 & 18.1111 & 15.1852 & 2.2593 \\
\sysnameo &\textbf{1.0000} &\textbf{1.0000} &20.0000 &15.2222 &4.6667 \\
\sysnameq &\textbf{1.0000} &\textbf{1.0000} &\textbf{15.6667} &\textbf{12.2593} &\textbf{1.9259} \\

\rowcolor[gray]{.9} \multicolumn{6}{c}{Sort Solids} \\*
MCTS & 0.1111 & 0.1481 & 29.0000 & - & - \\
% CONHERENT & 0.8889 & 0.9722 & 9.7778 & 7.2963 & 0.0000 \\
% LLM-MCTS & & & & & & & & & & \\
CoELA & \textbf{1.0000} & \textbf{1.0000} & 8.2222 & 7.8148 & 1.0741 \\
\sysnameo &0.8333 &0.9167 &11.0000 &10.4444 &2.0000 \\
\sysnameq &\textbf{1.0000} &\textbf{1.0000} &\textbf{7.1111} &\textbf{7.0000} &\textbf{0.2593} \\

\rowcolor[gray]{.8} \multicolumn{6}{c}{\textbf{Ma-Mo-UAV}\boldmath$\mathbf{\;\rightarrow\;}$\textbf{Ma-MoMa-Mo-UAV}} \\*
\rowcolor[gray]{.9} \multicolumn{6}{c}{Pack Objects} \\*
MCTS & 0.0000 & 0.0000 & 30.0000 & - & - \\
% CONHERENT & 0.0000 & 0.3148 & 30.0000 & 20.5926 & 0.0000 \\
% LLM-MCTS & & & & & & & & & & \\
CoELA & 0.1111 & 0.4259 & 29.6667 & 23.7037 & 5.0741 \\
\sysnameo &0.5000 &0.7500 &26.0000 &20.8750 &7.5000 \\
\sysnameq &\textbf{0.7778} &\textbf{0.9074} &\textbf{23.5556} &\textbf{17.6296} &\textbf{3.4074} \\

\rowcolor[gray]{.9} \multicolumn{6}{c}{Make Sandwich} \\*
MCTS & 0.0000 & 0.0000 & 30.0000 & - & - \\
% CONHERENT & 0.5556 & 0.6852 & 22.1111 & 12.9259 & 0.0000 \\
% LLM-MCTS & & & & & & & & & & \\
CoELA & \textbf{1.0000} & \textbf{1.0000} & 15.5556 & 12.9630 & 2.9259 \\
\sysnameo &0.8333 &0.8333 &18.5000 &15.8333 &4.8889 \\
\sysnameq &\textbf{1.0000} &\textbf{1.0000} &\textbf{14.8889} &\textbf{12.1111} &\textbf{1.1481} \\

\rowcolor[gray]{.9} \multicolumn{6}{c}{Sort Solids} \\*
MCTS & 0.0000 & 0.0278 & 30.0000 & - & - \\
% CONHERENT & 1.0000 & 1.0000 & 8.6667 & 5.8148 & 0.0000 \\
% LLM-MCTS & & & & & & & & & & \\
CoELA & \textbf{1.0000} & \textbf{1.0000} & 6.6667 & 6.5926 & 0.9259 \\
\sysnameo &0.6667 &0.8056 &20.3333 &17.6667 &6.7222 \\
\sysnameq &\textbf{1.0000} &\textbf{1.0000} &\textbf{6.2778} &\textbf{6.0185} &\textbf{0.4444} \\

\end{longtable}
\endgroup

%% file: table/removal_change.tex
\begingroup
\small
\setlength{\tabcolsep}{3pt}
\renewcommand{\arraystretch}{1.0}
\begin{longtable}{>{\centering\arraybackslash}p{4.0cm}
                *{5}{>{\centering\arraybackslash}p{1.5cm}}}
\caption{Experimental results of adding new collaborators and removing existing collaborators ($T=5$)}
\label{tab:removal_exp}\\
\toprule
\multirow{2}{*}{\textbf{Method}}
& \multicolumn{5}{c}{\textbf{REC}} \\
\cmidrule(lr){2-6}
& SUCC$\uparrow$ & PS$\uparrow$ & TS$\downarrow$ & AS$\downarrow$ & CC$\downarrow$ \\
\midrule
\endfirsthead

\caption[]{Experimental results of adding new collaborators and removing existing collaborators ($T=5$), continued}\\
\toprule
\multirow{2}{*}{\textbf{Method}}
& \multicolumn{5}{c}{\textbf{REC}} \\
\cmidrule(lr){2-6}
& SUCC$\uparrow$ & PS$\uparrow$ & TS$\downarrow$ & AS$\downarrow$ & CC$\downarrow$ \\
\midrule
\endhead

\midrule
\multicolumn{6}{r}{\small Continued on next page}\\
\endfoot

\bottomrule
\endlastfoot

\rowcolor[gray]{.8} \multicolumn{6}{c}{\textbf{Ma-MoMa-UAV}\boldmath$\mathbf{\;\rightarrow\;}$\textbf{Ma-MoMa}} \\*
\rowcolor[gray]{.9} \multicolumn{6}{c}{Pack Objects} \\*
MCTS & 0.6667 & 0.7778 & 23.5556 & - & - \\
% CONHERENT & 0.7778 & 0.9259 & 18.1111 & 10.0000 & - \\
% LLM-MCTS & & & & & & & & & & \\ 
CoELA & 0.4444 & 0.5741 & 22.0000 & 11.7778 & \textbf{1.2593} \\
\sysnameo &0.7143 &0.7143 &19.2857 &11.6667 &4.1429 \\
\sysnameq &\textbf{0.8889} &\textbf{0.9630} &\textbf{17.3333} &\textbf{9.4815} &2.0000 \\

\rowcolor[gray]{.9} \multicolumn{6}{c}{Make Sandwich} \\*
MCTS & \textbf{0.8889} & \textbf{0.8889} & 19.0000 & - & - \\
% CONHERENT & 0.6667 & 0.7222 & 17.1111 & 8.7778 & - \\
% LLM-MCTS & & & & & & & & & & \\ 
CoELA & 0.7778 & 0.8333 & 18.7778 & 11.9630 & 1.6296 \\
\sysnameo &0.4444 &0.5556 &26.0000 &14.2222 &3.9630 \\
\sysnameq &0.7778 &0.8333 &\textbf{17.8889} &\textbf{10.8519} &\textbf{1.2259} \\

\rowcolor[gray]{.9} \multicolumn{6}{c}{Sort Solids} \\*
MCTS & 0.2222 & 0.2222 & 26.7778 & - & - \\
% CONHERENT & 0.7778 & 0.9352 & 13.4444 & 9.2963 & - \\
% LLM-MCTS & & & & & & & & & & \\ 
CoELA & 0.6667 & 0.8889 & 14.8889 & 10.4815 & 0.8889 \\
\sysnameo &0.6667 &0.8750 &15.5000 &11.1111 &2.3333 \\
\sysnameq &\textbf{0.8889} &\textbf{0.9722} &\textbf{8.4444} &\textbf{7.4074} &\textbf{0.1481} \\

\rowcolor[gray]{.8} \multicolumn{6}{c}{\textbf{Ma-MoMa-UAV}\boldmath$\mathbf{\;\rightarrow\;}$\textbf{Ma-UAV}} \\*
\rowcolor[gray]{.9} \multicolumn{6}{c}{Pack Objects} \\*
MCTS & 0.3333 & 0.6111 & 24.5556 & - & - \\
% CONHERENT & 0.7778 & 0.8889 & 17.7778 & 10.0000 & - \\
% LLM-MCTS & & & & & & & & & & \\ 
CoELA & 0.2222 & 0.4074 & 25.1111 & 12.2963 & 2.2593 \\
\sysnameo &\textbf{0.6000} &\textbf{0.8889} &\textbf{19.2000} &12.5111 &3.6444\\
\sysnameq &0.3333 &0.5556 &23.0000 &\textbf{10.2222} &\textbf{2.1852}\\

\rowcolor[gray]{.9} \multicolumn{6}{c}{Make Sandwich} \\*
MCTS & \textbf{1.0000} & \textbf{1.0000} & \textbf{17.2222} & - & - \\
% CONHERENT & 1.0000 & 1.0000 & 12.6667 & 6.8148 & - \\
% LLM-MCTS & & & & & & & & & & \\ 
CoELA & \textbf{0.5556} & 0.6111 & 22.6667 & 12.8889 & \textbf{1.8519} \\
\sysnameo &\textbf{0.5556} &0.6667 &22.8889 &12.1111 &4.0370\\
\sysnameq &0.5556 &0.7222 &21.6667 &\textbf{10.6296} &2.7778\\

\rowcolor[gray]{.9} \multicolumn{6}{c}{Sort Solids} \\*
MCTS & 0.1111 & 0.2500 & 29.1111 & - & - \\
% CONHERENT & 1.0000 & 1.0000 & 8.0000 & 6.1852 & - \\
% LLM-MCTS & & & & & & & & & & \\ 
CoELA & \textbf{1.0000} & \textbf{1.0000} & 7.7778 & 6.7778 & 0.5556 \\
\sysnameo&\textbf{1.0000} &\textbf{1.0000} &9.1667 &8.1667 &0.5556 \\
\sysnameq &\textbf{1.0000} &\textbf{1.0000} &\textbf{6.2222} &\textbf{6.3704} &\textbf{0.0741} \\

\rowcolor[gray]{.8} \multicolumn{6}{c}{\textbf{Ma-MoMa-Mo-UAV}\boldmath$\mathbf{\;\rightarrow\;}$\textbf{Ma-MoMa-Mo}} \\*
\rowcolor[gray]{.9} \multicolumn{6}{c}{Pack Objects} \\*
MCTS & 0.5556 & 0.6481 & 24.4444 & - & - \\
% CONHERENT & 0.7778 & 0.9259 & 17.2222 & 10.8611 & - \\
% LLM-MCTS & & & & & & & & & & \\ 
CoELA & 0.1111 & 0.3704 & 30.0000 & 18.4722 & 3.5833 \\
\sysnameo &0.5556 &0.5926 &21.8889 &13.6944 &3.6944 \\
\sysnameq &\textbf{0.7778} &\textbf{0.8704} &\textbf{19.5556} &\textbf{11.7222} &\textbf{1.4722} \\

\rowcolor[gray]{.9} \multicolumn{6}{c}{Make Sandwich} \\*
MCTS & \textbf{1.0000} & \textbf{1.0000} & 17.3333 & - & - \\
% CONHERENT & 0.6667 & 0.7778 & 17.2222 & 10.8611 & - \\
% LLM-MCTS & & & & & & & & & & \\ 
CoELA & 0.7778 & 0.8889 & 20.5556 & 13.8333 & 2.0833 \\
\sysnameo&0.3333 &0.6111 &23.7778 &15.3889 &5.0833 \\
\sysnameq &\textbf{1.0000} &\textbf{1.0000} &\textbf{11.6667} &\textbf{8.4444} &\textbf{1.0000} \\

\rowcolor[gray]{.9} \multicolumn{6}{c}{Sort Solids} \\*
MCTS & 0.0000 & 0.0648 & 30.0000 & - & - \\
% CONHERENT & 0.8889 & 0.9722 & 10.5556 & 6.7500 & - \\
% LLM-MCTS & & & & & & & & & & \\ 
CoELA & \textbf{1.0000} & \textbf{1.0000} & 8.7778 & 7.5556 & 0.7500 \\
\sysnameo &0.3333 &0.7361 &26.1667 &18.7917 &5.7083 \\
\sysnameq &\textbf{1.0000} &\textbf{1.0000} &\textbf{6.8889} &\textbf{6.3611} &\textbf{0.2222} \\

\rowcolor[gray]{.8} \multicolumn{6}{c}{\textbf{Ma-MoMa-Mo-UAV}\boldmath$\mathbf{\;\rightarrow\;}$\textbf{Ma-Mo-UAV}} \\*
\rowcolor[gray]{.9} \multicolumn{6}{c}{Pack Objects} \\*
MCTS & 0.0000 & 0.0000 & 30.0000 & - & - \\
% CONHERENT & 0.7778 & 0.8889 & 17.3333 & 10.3611 & - \\
% LLM-MCTS & & & & & & & & & & \\ 
CoELA & 0.1111 & 0.2778 & 27.4444 & 17.2500 & 3.5556 \\
\sysnameo &0.4444 &0.6111 &20.8889 &13.8889 &4.5833 \\
\sysnameq &\textbf{0.5000} &\textbf{0.7778} &\textbf{18.8333} &\textbf{10.2083} &\textbf{1.5833} \\

\rowcolor[gray]{.9} \multicolumn{6}{c}{Make Sandwich} \\*
MCTS & 0.4444 & 0.5000 & 26.8889 & - & - \\
% CONHERENT & 0.8889 & 0.8889 & 19.3333 & 11.5000 & - \\
% LLM-MCTS & & & & & & & & & & \\ 
CoELA & 0.6667 & 0.6667 & 21.3333 & 13.3056 & 1.8889 \\
\sysnameo &0.6667 &0.7500 &\textbf{16.8333} &\textbf{10.1667} &2.5833 \\
\sysnameq &\textbf{0.7778} &\textbf{0.8333} &18.0000 &11.2778 &\textbf{1.4444} \\

\rowcolor[gray]{.9} \multicolumn{6}{c}{Sort Solids} \\*
MCTS & 0.0000 & 0.1111 & 30.0000 & - & - \\
% CONHERENT & 0.7778 & 0.8981 & 15.1111 & 9.0833 & - \\
% LLM-MCTS & & & & & & & & & & \\ 
CoELA & \textbf{1.0000} & \textbf{1.0000} & 6.8889 & 6.2778 & 0.7500 \\
\sysnameo &0.8333 &0.9444 &12.6667 &10.7500 &2.6250 \\
\sysnameq &\textbf{1.0000} &\textbf{1.0000} &\textbf{5.4444} &\textbf{4.5000} &\textbf{0.2222} \\

\end{longtable}
\endgroup

%% file: table/central_1.tex
\begingroup
\small
\setlength{\tabcolsep}{4pt}
\renewcommand{\arraystretch}{1.0}
\newcommand{\centralmethod}[1]{\multirow[c]{3}{3.0cm}{\centering\mbox{#1}}}
\begin{longtable}{
    >{\centering\arraybackslash}p{3.0cm}
    >{\centering\arraybackslash}p{1.5cm}
    *{5}{>{\centering\arraybackslash}p{1.5cm}}
}
\caption{Part 1: Centralized vs. Distributed LLM-Based Comparative Experiments}
\label{tab:central_1}\\
\toprule
\textbf{Method} & \textbf{Setting}
& SUCC$\uparrow$ & PS$\uparrow$ & TS$\downarrow$ & AS$\downarrow$ & CC$\downarrow$ \\
\midrule
\endfirsthead

\caption[]{Part 1: Centralized vs. Distributed LLM-Based Comparative Experiments, continued}\\
\toprule
\textbf{Method} & \textbf{Setting}
& SUCC$\uparrow$ & PS$\uparrow$ & TS$\downarrow$ & AS$\downarrow$ & CC$\downarrow$ \\
\midrule
\endhead

\midrule
\multicolumn{7}{r}{\small Continued on next page}\\
\endfoot

\bottomrule
\endlastfoot

\rowcolor[gray]{.8} \multicolumn{7}{c}{\textbf{2 Types of Robot}} \\*
\centralmethod{COHERENT}
& Static & 0.6482 & 0.7917 & 17.3333 & 11.0370 & - \\*
& CTO & 0.6111 & 0.7654 & 18.3519 & \textbf{10.3148} & - \\*
& IRZ & -- & -- & -- & -- & -- \\
\cmidrule(lr){2-7}

\centralmethod{\sysnameo}
& Static & 0.7963 & \textbf{0.8951} & 19.0740 & 14.6667 & 3.7685 \\*
& CTO & \textbf{0.8649} & \textbf{0.9634} & \textbf{16.0723} & 13.8440 & 3.1752 \\*
& IRZ & -- & -- & -- & -- & -- \\
\cmidrule(lr){2-7}

\centralmethod{\sysnameq}
& Static & \textbf{0.8148} & 0.8858 & \textbf{16.2407} & \textbf{10.5740} & \textbf{1.4074} \\*
& CTO & 0.7592 & 0.8580 & 17.7037 & 12.3148 & \textbf{1.8796} \\*
& IRZ & -- & -- & -- & -- & -- \\

\rowcolor[gray]{.8} \multicolumn{7}{c}{\textbf{3 Types of Robot}} \\*
\centralmethod{COHERENT}
& Static & 0.5309 & 0.7315 & 19.6790 & \textbf{11.8313} & - \\*
& CTO & 0.3580 & 0.5669 & 24.3333 & \textbf{12.8148} & - \\*
& IRZ & 0.6173 & 0.7644 & 18.5185 & 11.5473 & - \\
\cmidrule(lr){2-7}

\centralmethod{\sysnameo}
& Static & \textbf{0.8025} & \textbf{0.8920} & 18.5803 & 15.3045 & 4.2428 \\*
& CTO & \textbf{0.7439} & \textbf{0.8704} & 19.6900 & 16.3515 & 4.5382 \\*
& IRZ & 0.7818 & 0.9089 & 21.7068 & 18.6451 & 4.9280 \\
\cmidrule(lr){2-7}

\centralmethod{\sysnameq}
& Static & 0.7821 & 0.8848 & \textbf{17.1234} & 12.3128 & \textbf{1.6276} \\*
& CTO & 0.6914 & 0.8159 & \textbf{19.4197} & 14.2346 & \textbf{2.2510} \\*
& IRZ & \textbf{0.9630} & \textbf{0.9856} & \textbf{13.9876} & \textbf{10.7119} & \textbf{1.2881} \\

\rowcolor[gray]{.8} \multicolumn{7}{c}{\textbf{4 Types of Robot}} \\*
\centralmethod{COHERENT}
& Static & 0.5556 & 0.6358 & 18.4444 & \textbf{6.6296} & - \\*
& CTO & 0.5556 & 0.7315 & 21.7778 & \textbf{9.7778} & - \\*
& IRZ & 0.7408 & 0.8858 & 15.9630 & 9.1852 & - \\
\cmidrule(lr){2-7}

\centralmethod{\sysnameo}
& Static & 0.8642 & 0.9445 & 18.1506 & 15.5238 & 3.8921 \\*
& CTO & 0.8951 & 0.9321 & 18.7778 & 16.5972 & 2.8102 \\*
& IRZ & 0.9167 & 0.9722 & 18.3426 & 15.7299 & 3.5286 \\
\cmidrule(lr){2-7}

\centralmethod{\sysnameq}
& Static & \textbf{0.9630} & \textbf{0.9877} & \textbf{14.0370} & 11.1111 & \textbf{1.5000} \\*
& CTO & \textbf{0.9630} & \textbf{0.9815} & \textbf{16.2407} & 12.9770 & \textbf{1.5949} \\*
& IRZ & \textbf{1.0000} & \textbf{1.0000} & \textbf{12.9630} & \textbf{10.6389} & \textbf{1.1296} \\
\end{longtable}
\endgroup

%% file: table/central_2.tex
\begingroup
\small
\setlength{\tabcolsep}{4pt}
\renewcommand{\arraystretch}{1.0}
\newcommand{\centralchangemethod}[1]{\multirow[c]{2}{3.0cm}{\centering\mbox{#1}}}
\begin{longtable}{
    >{\centering\arraybackslash}p{3.0cm}
    >{\centering\arraybackslash}p{1.5cm}
    *{5}{>{\centering\arraybackslash}p{1.5cm}}
}
\caption{Part 2: Centralized vs. Distributed LLM-Based Comparative Experiments}
\label{tab:central_2}\\
\toprule
\textbf{Method} & \textbf{Setting}
& SUCC$\uparrow$ & PS$\uparrow$ & TS$\downarrow$ & AS$\downarrow$ & CC$\downarrow$ \\
\midrule
\endfirsthead

\caption[]{Part 2: Centralized vs. Distributed LLM-Based Comparative Experiments, continued}\\
\toprule
\textbf{Method} & \textbf{Setting}
& SUCC$\uparrow$ & PS$\uparrow$ & TS$\downarrow$ & AS$\downarrow$ & CC$\downarrow$ \\
\midrule
\endhead

\midrule
\multicolumn{7}{r}{\small Continued on next page}\\
\endfoot

\bottomrule
\endlastfoot

\rowcolor[gray]{.8} \multicolumn{7}{c}{\textbf{Ma-MoMa $\rightarrow$ Ma-MoMa-UAV / Ma-MoMa-UAV $\rightarrow$ Ma-MoMa}} \\*
\centralchangemethod{COHERENT}
& ANC & 0.8148 & 0.9228 & \textbf{15.8889} & \textbf{10.9444} & - \\*
& REC & 0.5185 & 0.6235 & 20.5556 & 9.3704 & - \\
\cmidrule(lr){2-7}

\centralchangemethod{\sysnameo}
& ANC & \textbf{0.8889} & 0.9167 & 17.5000 & 14.5000 & 3.7500 \\*
& REC & 0.6085 & 0.7149 & 20.2620 & 12.3330 & 3.4800 \\
\cmidrule(lr){2-7}

\centralchangemethod{\sysnameq}
& ANC & \textbf{0.8889} & \textbf{0.9444} & 17.1482 & 11.4815 & \textbf{1.5148} \\*
& REC & \textbf{0.8519} & \textbf{0.9228} & \textbf{14.5560} & \textbf{9.2470} & \textbf{1.1240} \\

\rowcolor[gray]{.8} \multicolumn{7}{c}{\textbf{Ma-UAV $\rightarrow$ Ma-MoMa-UAV / Ma-MoMa-UAV $\rightarrow$ Ma-UAV}} \\*
\centralchangemethod{COHERENT}
& ANC & 0.5926 & 0.7222 & 17.9630 & \textbf{11.2222} & - \\*
& REC & 0.5926 & 0.7192 & 17.8519 & \textbf{8.0494} & - \\
\cmidrule(lr){2-7}

\centralchangemethod{\sysnameo}
& ANC & 0.6296 & 0.8364 & 24.9445 & 19.7408 & 7.5185 \\*
& REC & \textbf{0.7185} & \textbf{0.8520} & 17.0850 & 10.9290 & 2.7457 \\
\cmidrule(lr){2-7}

\centralchangemethod{\sysnameq}
& ANC & \textbf{0.7531} & \textbf{0.8704} & \textbf{17.0000} & 11.3333 & \textbf{2.8519} \\*
& REC & 0.6296 & 0.7593 & \textbf{16.9630} & 9.0740 & \textbf{1.6790} \\

\rowcolor[gray]{.8} \multicolumn{7}{c}{\textbf{Ma-MoMa-Mo $\rightarrow$ Ma-MoMa-Mo-UAV / Ma-MoMa-Mo-UAV $\rightarrow$ Ma-MoMa-Mo}} \\*
\centralchangemethod{COHERENT}
& ANC & 0.6296 & 0.7994 & 18.7407 & 13.2346 & - \\*
& REC & 0.5185 & 0.7407 & 20.7037 & 10.7037 & - \\
\cmidrule(lr){2-7}

\centralchangemethod{\sysnameo}
& ANC & 0.9028 & 0.9583 & 18.4167 & 15.1250 & 4.2917 \\*
& REC & 0.4074 & 0.6466 & 23.9440 & 15.9580 & 4.8290 \\
\cmidrule(lr){2-7}

\centralchangemethod{\sysnameq}
& ANC & \textbf{0.9259} & \textbf{0.9630} & \textbf{14.1852} & \textbf{11.3951} & \textbf{1.4815} \\*
& REC & \textbf{0.9259} & \textbf{0.9568} & \textbf{12.7040} & \textbf{8.8420} & \textbf{0.8980} \\

\rowcolor[gray]{.8} \multicolumn{7}{c}{\textbf{Ma-Mo-UAV $\rightarrow$ Ma-MoMa-Mo-UAV / Ma-MoMa-Mo-UAV $\rightarrow$ Ma-Mo-UAV}} \\*
\centralchangemethod{COHERENT}
& ANC & 0.5185 & 0.6667 & 20.2593 & 13.1111 & - \\*
& REC & 0.5926 & 0.7438 & 17.5926 & 9.3148 & - \\
\cmidrule(lr){2-7}

\centralchangemethod{\sysnameo}
& ANC & 0.6667 & 0.7963 & 21.6111 & 18.1250 & 6.3704 \\*
& REC & 0.6481 & 0.7685 & 16.7960 & 11.6020 & 3.2640 \\
\cmidrule(lr){2-7}

\centralchangemethod{\sysnameq}
& ANC & \textbf{0.9259} & \textbf{0.9691} & \textbf{14.9074} & \textbf{11.9197} & \textbf{1.6666} \\*
& REC & \textbf{0.7593} & \textbf{0.8704} & \textbf{14.0930} & \textbf{8.6620} & \textbf{1.0830} \\
\end{longtable}
\endgroup

%% file: table/ablation.tex
\begingroup
\small
\setlength{\tabcolsep}{3.5pt}
\renewcommand{\arraystretch}{1.0}
\newcommand{\ablationmethod}[1]{\multirow[c]{5}{4.0cm}{\centering #1}}
\begin{longtable}{@{}
    >{\centering\arraybackslash}p{4.0cm}
    >{\centering\arraybackslash}p{1.15cm}
    *{5}{>{\centering\arraybackslash}p{1.22cm}}
@{}}
\caption{Ablation Experiments}
\label{tab:ablation}\\
\toprule
\textbf{Method} & \textbf{Setting}
& SUCC$\uparrow$ & PS$\uparrow$ & TS$\downarrow$ & AS$\downarrow$ & CC$\downarrow$ \\
\midrule
\endfirsthead

\caption[]{Ablation Experiments, continued}\\
\toprule
\textbf{Method} & \textbf{Setting}
& SUCC$\uparrow$ & PS$\uparrow$ & TS$\downarrow$ & AS$\downarrow$ & CC$\downarrow$ \\
\midrule
\endhead

\midrule
\multicolumn{7}{r}{\footnotesize Continued on next page}\\
\endfoot

\bottomrule
\endlastfoot

\rowcolor[gray]{.85} \multicolumn{7}{c}{\textbf{Closed Source Model}} \\*
\ablationmethod{Gemini-2.5-Flash}
& Static & 0.3704 & 0.6199 & 25.3722 & 16.4588 & 3.6264 \\*
& CTO & 0.4963 & 0.6488 & 23.9815 & 13.7238 & 3.0588 \\*
& IRZ & 0.3333 & 0.5069 & 26.0556 & 14.8526 & 3.3725 \\*
& ANC & 0.5000 & 0.6898 & 21.0000 & 14.2381 & 3.0810 \\*
& REC & 0.1111 & 0.3889 & 28.8889 & 10.1997 & 2.5926 \\
\cmidrule(lr){2-7}

\ablationmethod{Claude-3.7-Sonnet}
& Static & 0.7593 & 0.8086 & 20.1586 & 17.8563 & 3.1846 \\*
& CTO & 0.7963 & 0.8457 & 20.9563 & 16.5783 & 3.2586 \\*
& IRZ & 0.8086 & 0.9074 & 22.5643 & 16.5561 & 3.8533 \\*
& ANC & 0.7593 & 0.8272 & 22.8423 & 13.5698 & 2.9630 \\*
& REC & 0.6173 & 0.7840 & 17.8333 & 13.6389 & 3.7700 \\
\cmidrule(lr){2-7}

\ablationmethod{\sysnameo}
& Static & 0.8210 & 0.9105 & 18.6016 & 15.1650 & 3.9678 \\*
& CTO & \textbf{0.8346} & \textbf{0.9220} & 18.1800 & 15.5975 & 3.5079 \\*
& IRZ & 0.8492 & 0.9406 & 20.0247 & 17.1875 & 4.2283 \\*
& ANC & 0.7720 & 0.8769 & 20.6181 & 16.8727 & 5.4826 \\*
& REC & 0.5956 & 0.7455 & 19.5220 & 12.7058 & 3.5795 \\

\rowcolor[gray]{.85} \multicolumn{7}{c}{\textbf{Open Source Model}} \\*
\ablationmethod{DeepSeek-V3}
& Static & 0.7222 & 0.7574 & 21.8278 & 16.2147 & 3.7364 \\*
& CTO & 0.6111 & 0.7315 & 20.8611 & 17.0080 & 4.2891 \\*
& IRZ & 0.3333 & 0.5602 & 25.3889 & 19.2603 & 5.2885 \\*
& ANC & 0.6667 & 0.8056 & 18.8333 & 14.3619 & 3.9048 \\*
& REC & 0.3333 & 0.5694 & 27.0000 & 14.5516 & 3.4206 \\
\cmidrule(lr){2-7}

\ablationmethod{\makecell[c]{Qwen3-235B-A22B\\Instruct}}
& Static & 0.3704 & 0.6329 & 24.3352 & 13.3042 & 2.4306 \\*
& CTO & 0.2778 & 0.5241 & 24.7889 & 16.4026 & 2.5542 \\*
& IRZ & 0.4167 & 0.6389 & 22.5556 & 15.5778 & 2.8962 \\*
& ANC & 0.4167 & 0.7176 & 22.9444 & 14.5143 & 3.0190 \\*
& REC & 0.1667 & 0.5278 & 26.8333 & 12.4067 & 1.4187 \\

\rowcolor[gray]{.85} \multicolumn{7}{c}{\textbf{Component Ablation}} \\*
\ablationmethod{\makecell[c]{\sysnameq\\$w.o.$ history}}
& Static & 0.8333 & 0.9012 & 16.1730 & 11.9103 & 1.5633 \\*
& CTO & 0.6852 & 0.8377 & 21.1150 & 17.1475 & 2.2122 \\*
& IRZ & 0.8056 & 0.8889 & 15.8442 & 13.4937 & 1.6583 \\*
& ANC & 0.8611 & 0.9259 & 16.8760 & 13.3588 & 1.9054 \\*
& REC & 0.7778 & 0.8420 & 17.0625 & 10.0403 & 1.4332 \\
\cmidrule(lr){2-7}

\ablationmethod{\makecell[c]{\sysnameq\\$w.o.$ feedback}}
& Static & 0.5224 & 0.7147 & 20.9709 & 16.1838 & 1.9947 \\*
& CTO & 0.3889 & 0.6345 & 22.3621 & 18.7633 & 2.4925 \\*
& IRZ & 0.5278 & 0.7108 & 20.0119 & 16.4538 & 1.9605 \\*
& ANC & 0.5000 & 0.7312 & 21.3869 & 16.6796 & 2.9991 \\*
& REC & 0.5667 & 0.6562 & 20.2569 & 12.4185 & 1.5578 \\
\cmidrule(lr){2-7}

\ablationmethod{\makecell[c]{\sysnameq\\$w.o.$ leader}}
& Static & 0.7222 & 0.8642 & 16.9470 & 12.2340 & \textbf{1.1905} \\*
& CTO & 0.6013 & 0.8310 & 18.6377 & 13.3396 & 2.1262 \\*
& IRZ & 0.8889 & 0.9484 & 16.4524 & 12.5648 & 1.5922 \\*
& ANC & 0.7500 & 0.8869 & 17.5546 & 12.1269 & \textbf{1.6921} \\*
& REC & 0.7500 & 0.8455 & 17.2917 & 10.5222 & 1.3095 \\
\cmidrule(lr){2-7}

\ablationmethod{\makecell[c]{\sysnameq\\$w.o.$ reflection}}
& Static & 0.7407 & 0.8642 & 17.6667 & 16.0833 & 1.6111 \\*
& CTO & 0.6790 & 0.8148 & 18.7778 & 16.8611 & 2.0833 \\*
& IRZ & 0.8704 & 0.9321 & 14.5000 & 13.1111 & 1.4722 \\*
& ANC & 0.8333 & 0.8827 & 17.2593 & 14.2854 & 2.0564 \\*
& REC & 0.7593 & 0.8086 & 19.0634 & 11.6548 & 1.4963 \\
\cmidrule(lr){2-7}

\ablationmethod{\sysnameq}
& Static & \textbf{0.8533} & \textbf{0.9194} & \textbf{15.8004} & \textbf{11.3326} & 1.5117 \\*
& CTO & 0.8045 & 0.8851 & \textbf{17.7881} & \textbf{13.1754} & \textbf{1.9085} \\*
& IRZ & \textbf{0.9815} & \textbf{0.9928} & \textbf{13.4753} & \textbf{10.6754} & \textbf{1.2089} \\*
& ANC & \textbf{0.8735} & \textbf{0.9367} & \textbf{15.8102} & \textbf{11.5324} & 1.8787 \\*
& REC & \textbf{0.7917} & \textbf{0.8773} & \textbf{14.5787} & \textbf{8.9564} & \textbf{1.1963} \\
\end{longtable}
\endgroup

%% file: table/real.tex
\begin{table}[!htbp] 

\centering

\caption{Comparison results of real-world experiments} 
\label{tab:real} 
\renewcommand{\arraystretch}{1.0} 
\small
\setlength{\tabcolsep}{5pt}
\begin{tabular}{>{\centering\arraybackslash}p{3.5cm} 
                >{\centering\arraybackslash}p{1.1cm}
                >{\centering\arraybackslash}p{1.1cm}
                >{\centering\arraybackslash}p{1.1cm}
                >{\centering\arraybackslash}p{1.1cm}
                >{\centering\arraybackslash}p{1.1cm}}
\toprule
\textbf{Method} & SUCC$\uparrow$ & PS$\uparrow$ & TS$\downarrow$ & AS$\downarrow$ & CC$\downarrow$ \\ 
\midrule

COELA &4/8 & 6/10 &11.75 &8.43 &1.81 \\ 
\sysnameo &5/8 & 7/10 &10.56 &8.00 &2.06 \\ 
\sysnameq &\textbf{8/8} & \textbf{10/10} &\textbf{8.13} &\textbf{6.88} &\textbf{0.94} \\ 

\bottomrule
\end{tabular}
\end{table}

%% file: chapters/5.conclusion.tex
\section{conclusion and future work}
In this paper, we presented \sysname, a decentralized framework that leverages LLMs to enable robust and adaptive collaboration among heterogeneous multi-robot systems under dynamic task conditions. By formulating the problem within the DEC-POMDP paradigm and integrating reasoning-driven communication and feedback-based replanning, our approach achieves continuous collaboration and resilience in the face of evolving task goals, fluctuating team compositions, and environmental constraints. Through extensive simulations and real-world experiments, \sysname demonstrates superior performance in success rate, efficiency, and generalization compared with baselines, validating the effectiveness of LLM-based reasoning for decentralized robotic collaboration. Although the current benchmark evaluates teams with 2--4 robot types, scaling \sysname to substantially larger teams, such as tens of robots, remains an important future direction. Such settings may require hierarchical coordination, message filtering, summary compression, and more efficient communication protocols to prevent discussion overhead from growing with team size.

Looking forward, we plan to extend this framework toward broader real-world deployments and more diverse modalities. Future work will focus on leveraging multimodal models to better perceive and interpret success and failure during task execution, thereby generating more diverse and informative feedback. In addition, we plan to adopt reinforcement learning paradigms to fine-tune LLMs, enhancing their long-horizon reasoning and self-correction capabilities, enabling robots to autonomously refine their cooperative strategies from interaction data. We believe that such advances will bridge the gap between high-level language reasoning and low-level robotic control, paving the way for scalable embodied intelligence in complex, causally unstructured environments.

%% file: chapters/6.appendix.tex
\subsection{Pseudocode}
\input{table/code}
\input{table/data_collection_code}
\subsection{Detailed prompt templates}
\subsubsection{Self-Description Prompts}~\label{lab:selfDP}
\input{prompts/self_description}
\subsubsection{Task Allocation and Leadership Bidding Prompts}~\label{lab:TALBP}
\input{prompts/task_allocation_leadership_bidding}
\subsubsection{Leader Election Prompts}~\label{lab:LEP}
\input{prompts/leader_election}
\subsubsection{Heterogeneous Multi-Robot Collaboration Prompts}~\label{lab:HMRCP}
\input{prompts/heterogeneous_multirobot_collaboration}
\subsubsection{Reflection and Group Discussion Prompts}~\label{lab:RGDP}
\input{prompts/reflection_group_discussion}

\input{table/weight_sensitivity}~\label{lab:sensitivity}

\subsection{Detailed Module Architecture Diagram}~\label{lab:detailed_arch}

\begin{figure*}[htbp]
    \centering
    \includegraphics[width=\textwidth]{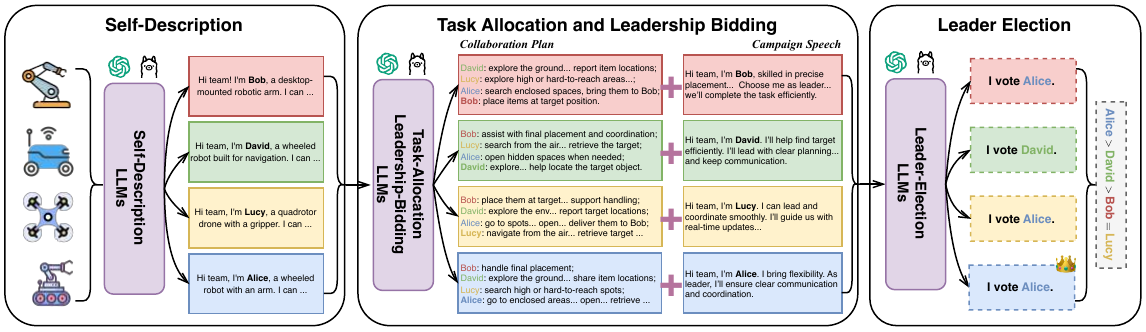}
    \caption{Detailed task initialization process. The collaborating robot team first generates individual profiles through self-description. Each robot then proposes a Collaboration Plan and Campaign Speech based on the task context and the information shared by other robots. These outputs are further used to elect the leader. The voting stage allows up to three attempts; if no agreement is reached, an additional LLM is introduced for final arbitration. This process serves as a warm-up stage for the closed-loop collaboration and occurs only once during a complete task. The leader role remains fixed throughout subsequent task execution.}
    \label{fig:init}
\end{figure*}

\begin{figure*}[htbp]
    \centering
    \includegraphics[width=\textwidth]{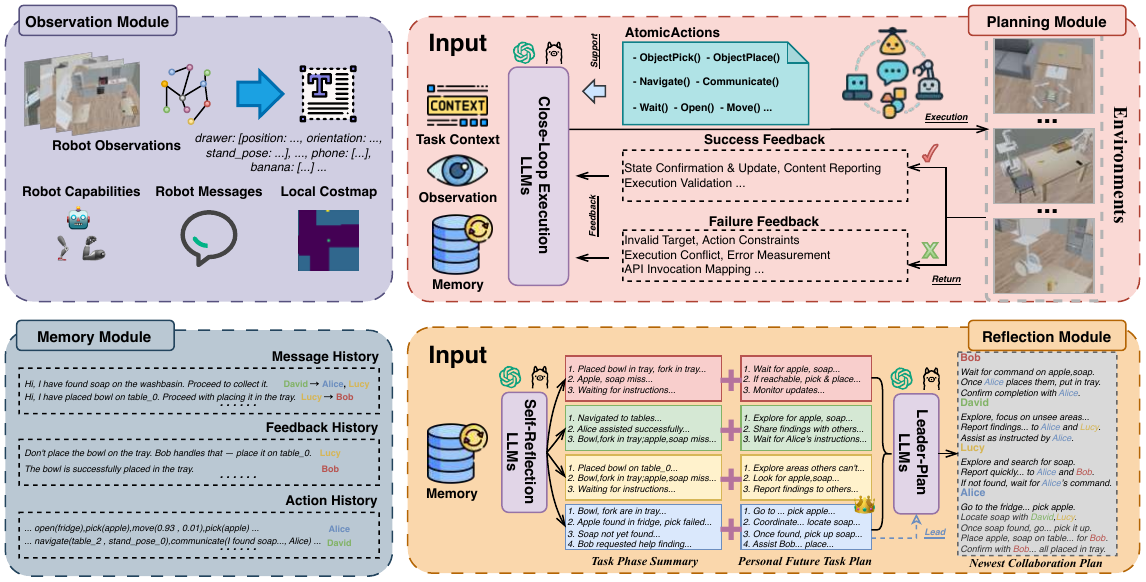}
    \caption{Detailed structural design of the four functional modules. The Observation Module acquires multimodal observation information; the Memory Module stores the task history, including messages, feedback, and actions; the Planning Module selects atomic actions based on the task context, observations, and memory, and interacts with the environment to obtain new information and continuously drive closed-loop task execution; the Reflection Module summarizes and reflects on the past history, after which the leader updates the latest global task plan.}
    \label{fig:module}
\end{figure*}

Fig.~\ref{fig:init} provides a detailed illustration of the task initialization process, and Fig.~\ref{fig:module} shows the detailed structural design of the four functional modules.

\subsection{Box Plot Comparison on Pack Objects Task}~\label{lab:box_plots}
\begin{figure}[h]
    \centering
    \includegraphics[width=\columnwidth]{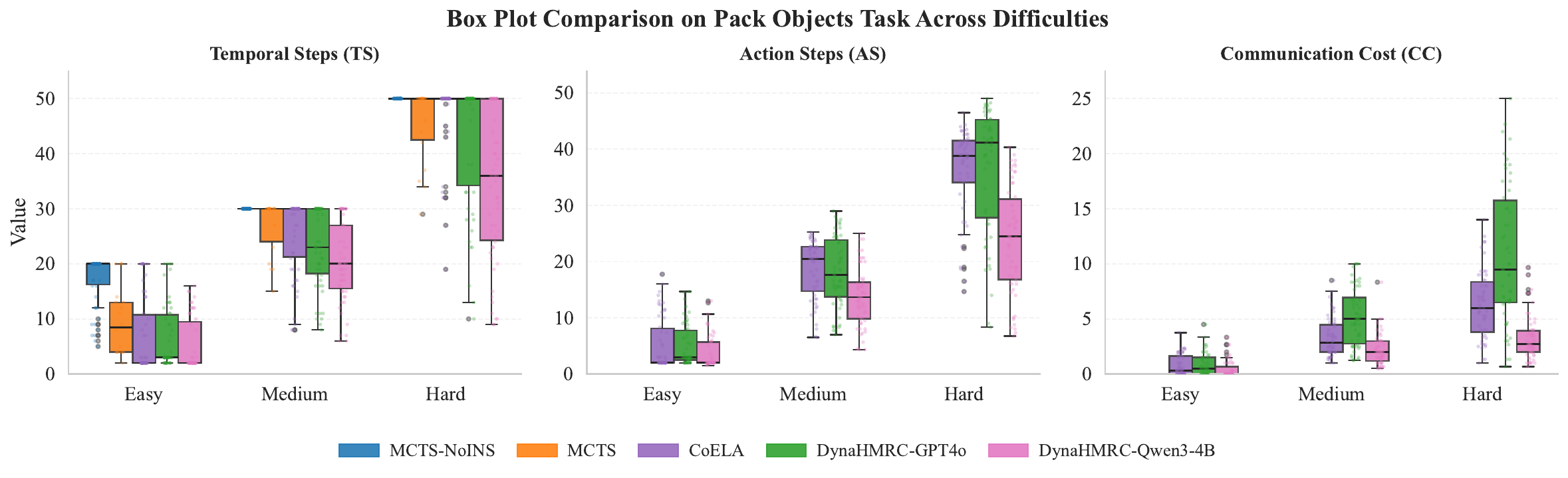}
    \caption{Performance of different methods under different task difficulties on the Pack Objects Task.}
    \label{fig:box}
\end{figure}

Fig.~\ref{fig:box} further shows the box-plot distributions of TS, AS, and CC for different methods on the Pack Objects Task under Easy, Medium, and Hard settings. As task difficulty increases, all methods incur higher temporal, action, and communication costs, but the degree of degradation differs substantially. MCTS-NoINS and MCTS both deteriorate rapidly, with their TS medians reaching the time budget limit in harder settings, indicating poor stability in complex tasks. CoELA and both DynaHMRC variants are generally more efficient, but CoELA still exhibits strong concentration near the upper budget bound in Hard tasks. Among all methods, \sysnameq shows the most favorable efficiency--robustness trade-off, with consistently lower medians in TS, AS, and CC than \sysnameo and CoELA, especially in Medium and Hard settings. This suggests that its advantage is not driven by a few lucky runs, but is reflected in a more stable trial-level distribution with lower action and communication overhead.

\subsection{Failure Case Study}~\label{lab:failure_case}
\begin{figure*}[htbp]
    \centering
    \includegraphics[width=\textwidth]{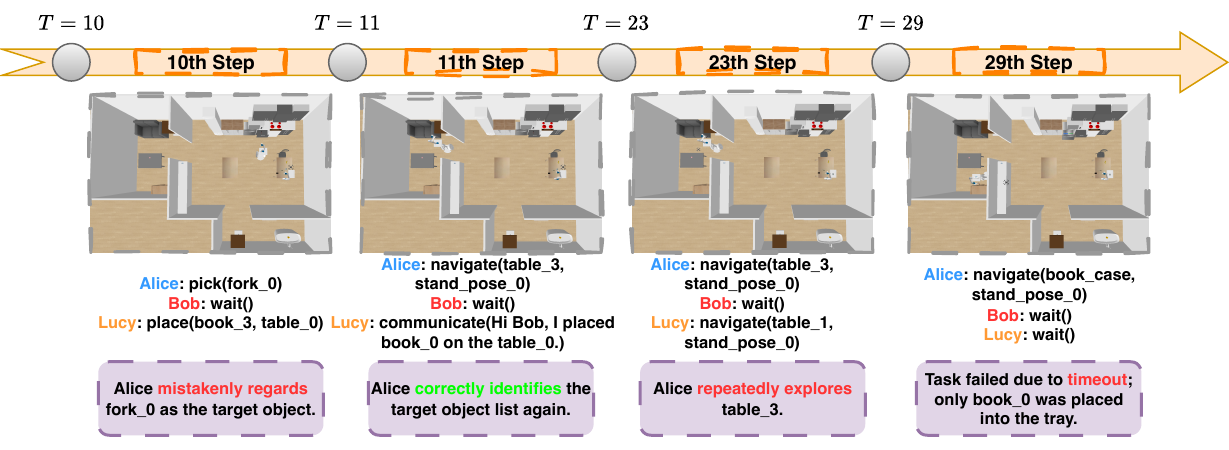}
    \caption{This figure illustrates a task failure process. The task goal is to place the book and bowl into the tray.}
    \label{fig:failure_case}
\end{figure*}

Fig.~\ref{fig:failure_case} and Sec.~\ref {lab:failure_case_logs} illustrate the reasoning and actions of the mobile manipulation robot Alice in a failed episode. At step 10, Alice mistakenly identifies \texttt{fork\_0} as the target object and executes an incorrect pick action. At step 11, Alice realizes that \texttt{fork\_0} is not the target and resumes searching for \texttt{bowl\_0}. At step 23, due to the bounded memory window, Alice redundantly explores \texttt{table\_3}. At step 29, since \texttt{bowl\_0} has still not been placed into the tray and the time limit is reached, the task ultimately fails.

\subsubsection{Key Step Logs of the Failure Case}~\label{lab:failure_case_logs}
\input{prompts/fail_log}

\subsection{Performance Comparison under Different Memory Queue Lengths}~\label{lab:diff_memory_length}
To analyze how large memory storage affects LLM reasoning, we conduct a memory-queue-length sensitivity study on the Pack Objects task under both Static and CTO settings. We compare the default memory length $M=10$ with a larger setting $M=15$. For \textit{COHERENT}, $M$ denotes the length of global interaction history retained in the centralized prompt; for \textit{CoELA} and \sysname, it denotes the per-agent memory queue length.

As shown in Tab.~\ref{tab:memory_length}, simply increasing the memory length does not consistently improve performance. In Static tasks, longer memory can help preserve useful task-progress information. For example, \sysnameo improves from 0.6482 to 0.8889 in SUCC when \(M\) increases from 10 to 15, and \sysnameq improves from 0.8000 to 0.8519 in SUCC and from 0.8788 to 0.9568 in PS. This suggests that, when past information remains relevant, a longer memory queue can provide the LLM with more complete execution history and improve task grounding.

However, in dynamic CTO tasks, longer memory may also retain outdated target information after the task objective changes, which can distract the LLM from the current goal. For \textit{COHERENT}, increasing $M$ from 10 to 15 decreases SUCC from 0.4074 to 0.2407 under CTO, and from 0.3148 to 0.0556 under Static. This indicates that directly extending the global history in a centralized prompt can substantially increase the context burden and dilute task-relevant information. Similarly, \sysnameo and \sysnameq also show performance drops under CTO when $M$ increases, suggesting that stale-context interference is a general challenge for LLM-based long-horizon reasoning.

Comparing across methods, \sysname remains more competitive than the baselines under both memory settings. Under the default $M=10$, \sysnameo and \sysnameq outperform \textit{COHERENT} and \textit{CoELA} in SUCC and PS under both Static and CTO settings. Even when $M=15$, \sysnameq still maintains higher CTO SUCC than the baselines. This advantage suggests that the performance gain of \sysname does not simply come from using decentralized memory, but from a more structured and efficient organization mechanism. Each robot maintains its own task-relevant memory and reasons from its local execution context, while periodic reflection and leader-plan updates help identify useful information for coordination instead of forcing all history into a single centralized prompt.

Overall, these results show that memory should be bounded and task-aware rather than simply enlarged. Longer memory is beneficial when historical information remains relevant, but can harm dynamic adaptation when obsolete goals or outdated plans remain in the prompt. Designing more structured, hierarchical, and dynamically updated memory mechanisms is therefore an important direction for future work.

\input{table/memory}

\subsection{Inference Latency Evaluation on Real Robot Platform}~\label{lab:latency}

To evaluate deployment feasibility, we measure inference latency and communication overhead on the real-robot platform. As shown in Tab.~\ref{tab:latency-summary}, we evaluate three inference settings used in our deployment pipeline: local Qwen3-4B inference on an RTX 5060 Ti host PC, local Qwen3-4B inference on a Jetson AGX Orin edge device, and API-based GPT-4o inference. The local Qwen3-4B settings are used for robot-side or host-side task planning, while GPT-4o API calls are used in the real-world system for visual feedback and execution-state assessment.

For local inference, the RTX 5060 Ti achieves an average latency of 4181.70 ms and a generation speed of 76.95 tokens/s, while the Jetson AGX Orin requires 12401.69 ms on average with a generation speed of 27.07 tokens/s. The GPT-4o API setting shows an average end-to-end latency of 6067.07 ms. We do not report token-level throughput for the API setting because device-side generation speed is not directly observable.

We further measure ROS2 communication latency over 100 samples. The average communication latency is only 1.13 ms, indicating that robot-to-robot communication is negligible compared with model inference in our settings. Therefore, the dominant deployment latency comes from high-level model inference and visual feedback inference rather than ROS2 communication. Since \sysname performs task-level planning and feedback-based coordination while leaving low-level navigation and manipulation control to dedicated modules, second-level inference latency is acceptable for the long-horizon tasks considered in this work. However, the current system is not intended for high-frequency servo-level real-time control.

The detailed inference hardware configuration is as follows:
\begin{itemize}
    \item \textbf{NVIDIA Jetson AGX Orin 32GB}: Local inference on the NVIDIA Jetson AGX Orin 32GB was used for the Galaxea R1 Lite and Agilex Piper platforms. This device provides up to 200 TOPS AI performance, an NVIDIA Ampere GPU with 56 Tensor Cores and 1792 CUDA cores, an 8-core Arm Cortex-A78AE CPU, and 32 GB 256-bit LPDDR5 memory. The Qwen3-4B model was deployed on this device for local inference.

    \item \textbf{Host PC with NVIDIA GeForce RTX 5060 Ti}: Local inference on the host PC was used for the Agilex Tracer Mini platform. The host PC was equipped with an Intel Core i7-12700KF CPU, 32 GB DDR4 memory, and an NVIDIA GeForce RTX 5060 Ti GPU with 16 GB memory. The Qwen3-4B model was deployed on this device for local inference.
\end{itemize}

\input{table/latency}

\subsection{Statistical Significance Comparison}~\label{lab:statistical_significance}

To assess whether the performance gap between \sysnameq and \sysnameo is statistically reliable, we conduct paired significance tests under static and dynamic aggregate settings. The mean difference is defined as \sysnameq minus \sysnameo. As shown in Tab.~\ref{tab:statistical_comparison}, \sysnameq achieves significant reductions in TS, AS, and CC under both static and dynamic settings. In dynamic tasks, \sysnameq also significantly improves SUCC and PS. These results suggest that the advantage of \sysnameq is mainly reflected in a stronger efficiency--performance trade-off, especially under dynamic task variations.

\input{table/statistical_comparison}

\subsection{Leader-Election Stability and Degraded Voting}~\label{lab:leader_vote}

Leader election in \sysname is performed only once during task initialization. The elected leader remains fixed throughout the episode, and later discussion stages only update the high-level division of labor without re-electing a new leader. To analyze the stability of this one-time election process, we report the number of voting rounds required to obtain a unique leader and the final leader distribution across team compositions.

\begin{figure}[h]
    \centering
    \includegraphics[width=\columnwidth]{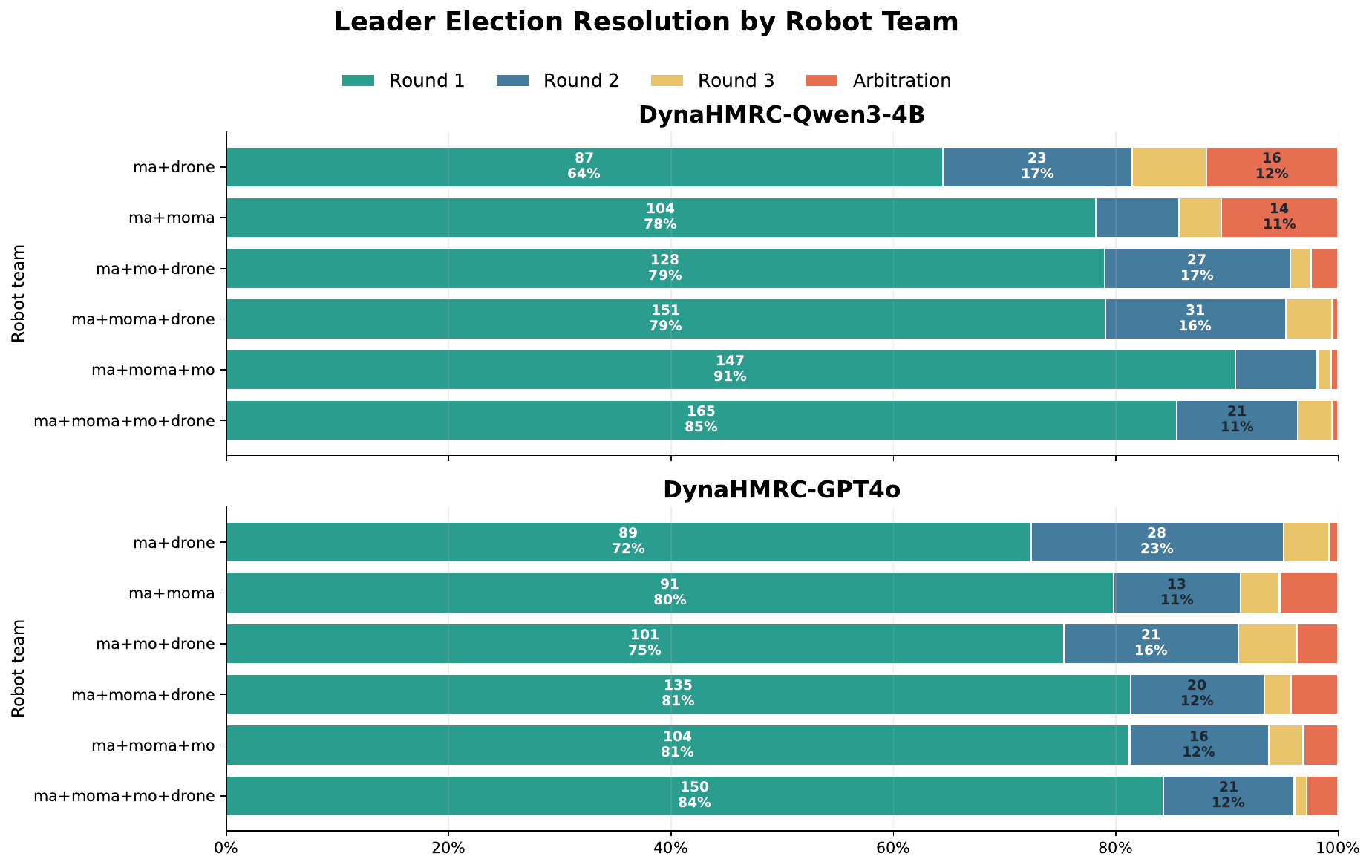}
    \caption{Leader Election Resolution}
    \label{fig:leader_rounds}
\end{figure}

As shown in Fig.~\ref{fig:leader_rounds}, most elections converge quickly. For \sysnameq and \sysnameo, 80.1\% and 79.5\% of cases obtain a unique leader in the first round, respectively, and 92.8\% and 93.6\% are resolved within two rounds. Only 3.8\% and 3.3\% require referee arbitration after three rounds. This indicates that leader election is generally stable and rarely depends on the fallback arbitration mechanism.

We further test robustness under degraded voting by randomly replacing 50\% of the votes during the leader election. As shown in Tab.~\ref{tab:random_votes}, \sysnameq remains robust under this perturbation. In Static tasks, the success rate changes only slightly for 2- and 3-type teams and remains unchanged for 4-type teams. In CTO tasks, the success rate remains unchanged across all team sizes. These results suggest that the downstream closed-loop execution and periodic coordination can tolerate partial voting noise, although high-quality votes remain important for selecting an appropriate leader.

\begin{figure}[h]
    \centering
    \includegraphics[width=\columnwidth]{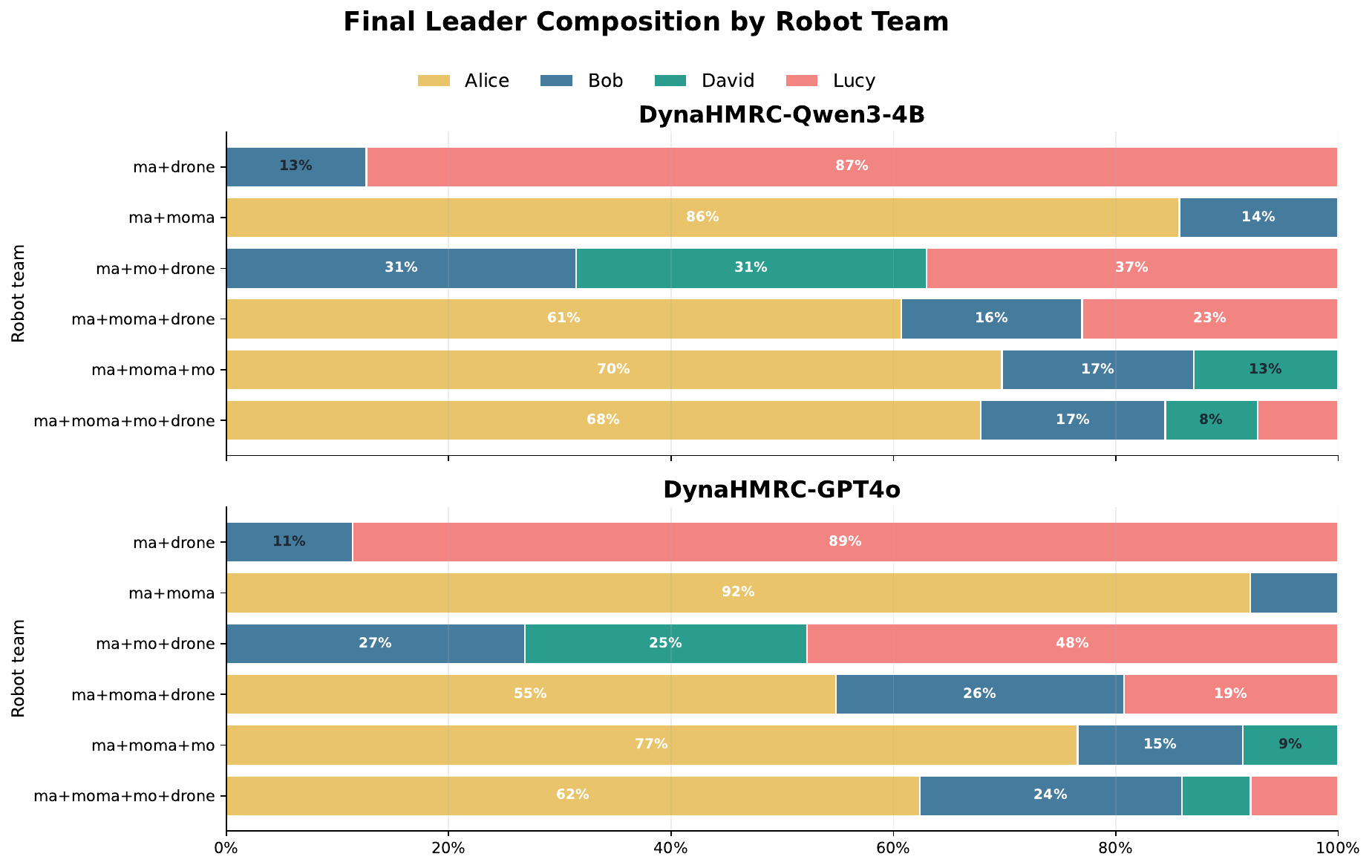}
    \caption{Final Leader Composition}
    \label{fig:final_leader}
\end{figure}

Fig.~\ref{fig:final_leader} further shows that final leader selection follows interpretable capability-aware patterns. When no mobile manipulator is available, the drone Lucy is frequently selected as the leader. When a mobile manipulator is included, Alice becomes the dominant leader in most team compositions, reflecting its combined mobility and manipulation capabilities. More ambiguous team compositions exhibit more distributed leadership, indicating that multiple robots can plausibly serve as coordination centers.

\input{table/select_leader}

\subsection{Fine-tuning Ablations}~\label{lab:fine_tuning}

To disentangle the contributions of different supervision signals, we train three Qwen3-4B variants using the same backbone, data split, and SFT hyperparameters. The only difference is the supervised target: action-only supervision, reasoning-only supervision, and joint reasoning-and-action supervision. As shown in Tab.~\ref{tab:supervision}, reasoning-only fails to produce reliable executable behavior, while action-only can generate actions but still performs poorly in long-horizon collaboration. In particular, all action-only runs reach the 30-step budget, with frequent off-target picks, irrelevant communication, and repeated stale actions. This indicates that action supervision alone mainly learns locally plausible next-action imitation, but does not sufficiently preserve task goals, teammate roles, or recovery strategies.

By contrast, joint reasoning-and-action supervision achieves substantially higher SUCC and PS while reducing TS and AS across different robot configurations. Reasoning supervision provides intermediate signals for subgoal tracking, role allocation, coordination, and failure correction, while action supervision grounds these decisions into executable atomic commands. These results show that both supervision signals are complementary and that the performance gain of \sysnameq comes from aligning high-level reasoning with low-level action generation.

\input{table/supervision}

\subsection{Dataset-size Learning Curves}~\label{lab:dataset_size_curves}
\begin{wrapfigure}{r}{0.5\textwidth}
    \centering
    \includegraphics[width=\linewidth]{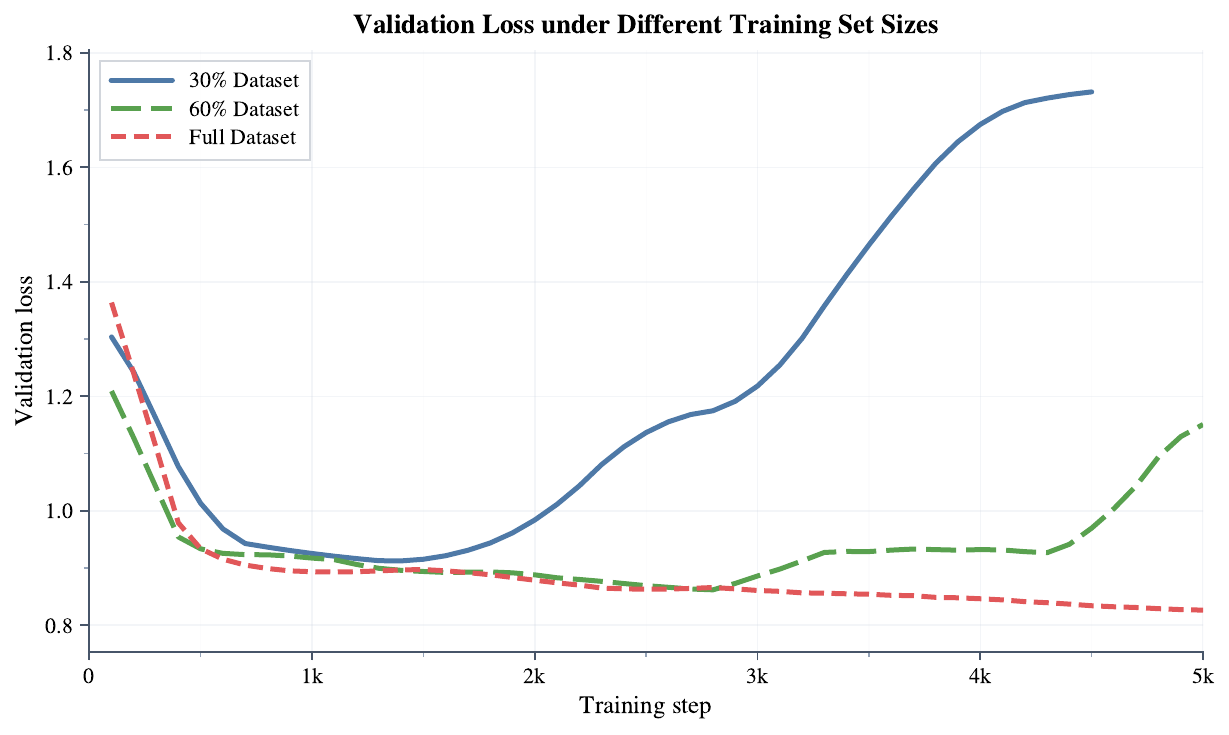}
    \caption{Comparison of validation loss curves under different training dataset sizes.}
    \label{fig:eval_curves}
    \vspace{-1.2cm}
\end{wrapfigure}
We further analyze the effect of dataset size by training Qwen3-4B with 30\%, 60\%, and 100\% of the curated dataset. All variants use the same validation set and training hyperparameters. The validation loss curves in Fig.~\ref{fig:eval_curves} show that using more training data improves generalization stability and delays overfitting, indicating that the model benefits from larger amounts of domain-specific reasoning-action supervision.

\subsection{Unseen Target Objects Generalization}~\label{lab:obj_generalization}

Tab.~\ref{tab:unseen_obj} reports the performance of \sysnameq on unseen target objects. The SUCC decreases from 0.8000 to 0.7407, while PS remains comparable to the original setting, indicating that the model can still achieve substantial partial progress despite increased object-level generalization difficulty. The higher TS and AS further suggest that unseen objects require longer execution and more corrective actions. Overall, benefiting from the strong semantic generalization ability of LLMs, \sysnameq demonstrates reasonable robustness under UTO, although unseen target objects still make complete task completion more challenging.

\input{table/generalization}

\subsection{Failure Mode Breakdown}~\label{lab:fail_mode_breakdown}

A complete real-world episode typically takes more than ten minutes, mainly due to the conservative execution protocol used for safe real-robot deployment. 
After each high-level decision, the system allows up to 120~s for navigation, manipulation, AnyGrasp-based grasp prediction, visual feedback, and recovery motions before proceeding to the next decision step. Thus, the reported completion time should be regarded as a conservative upper bound under our deployment setting, rather than the minimum achievable runtime of the framework.
Tab.~\ref{tab:real_failure_breakdown} presents the failure mode breakdown observed in real-world experiments. We categorize the failures into three types: perception errors, planning errors, and execution errors.

\input{table/real_failure}

\subsection{Sensitivity to Dynamic Variation Timing and Severity}~\label{lab:sensitivity_dyna}

To further evaluate robustness under dynamic task changes, we conduct additional experiments on the Pack Objects task. The default setting reported in the main experiments is $T=5$, and we additionally evaluate earlier and later trigger timings.

For CTO, the relevant results are reported in Tab.~\ref{tab:dT_CTO}, showing that earlier target changes are more disruptive to the baseline methods.
At $T=3$, \textit{COHERENT} and \textit{CoELA} achieve only 0.1481 and 0.2222 SUCC, respectively, whereas \sysnameq maintains 0.7593 SUCC and 0.9105 PS. Across $T=3,5,10$, \sysnameq consistently maintains strong performance, indicating better robustness to changing task objectives.

For IRZ, the relevant results are reported in Tab.~\ref{tab:dN_IRZ}, showing that performance does not strictly degrade as the number of restricted zones increases.
This is because the settings are constructed to remain feasible, and the actual difficulty depends on whether restricted zones block critical paths. In our experiments, restricted zones mainly invalidate certain \texttt{navigate} actions and return textual feedback, which \sysnameq can use to adjust its plan. Therefore, \sysnameq remains robust across 1, 2, and 3 restricted zones.

For ANC and REC, we vary the trigger timing to simulate early, middle, and late disruptions. The corresponding results are reported in Tab.~\ref{tab:dT_ANC_REC}. \sysnameq consistently achieves the best or near-best SUCC and PS, showing stronger adaptability to action-constraint changes and robot-team reconfiguration. These results demonstrate that \sysname is more robust than the baselines under different dynamic variation timings and severities.

\input{table/different_T_CTO}

\input{table/different_T_IRZ}

\input{table/different_T_ANC_REC}

\subsection{Hyperparameter Sensitivity}~\label{lab:hyper_sensitivity}

To justify the training configuration used for fine-tuning, we conduct a brief hyperparameter sensitivity analysis for Qwen3-4B. We mainly examine three factors: learning rate, training epochs, and batch size. Specifically, we compare learning rates of \(2\times10^{-5}\), \(1\times10^{-5}\), and \(5\times10^{-6}\), as well as batch sizes of 2, 4, and 8. For each setting, we train the model with the same dataset split, prompt template, and other hyperparameters, and report the validation loss curves over training epochs.

\begin{figure}[t]
    \centering
    \begin{minipage}[t]{0.48\textwidth}
        \centering
        \includegraphics[width=\linewidth]{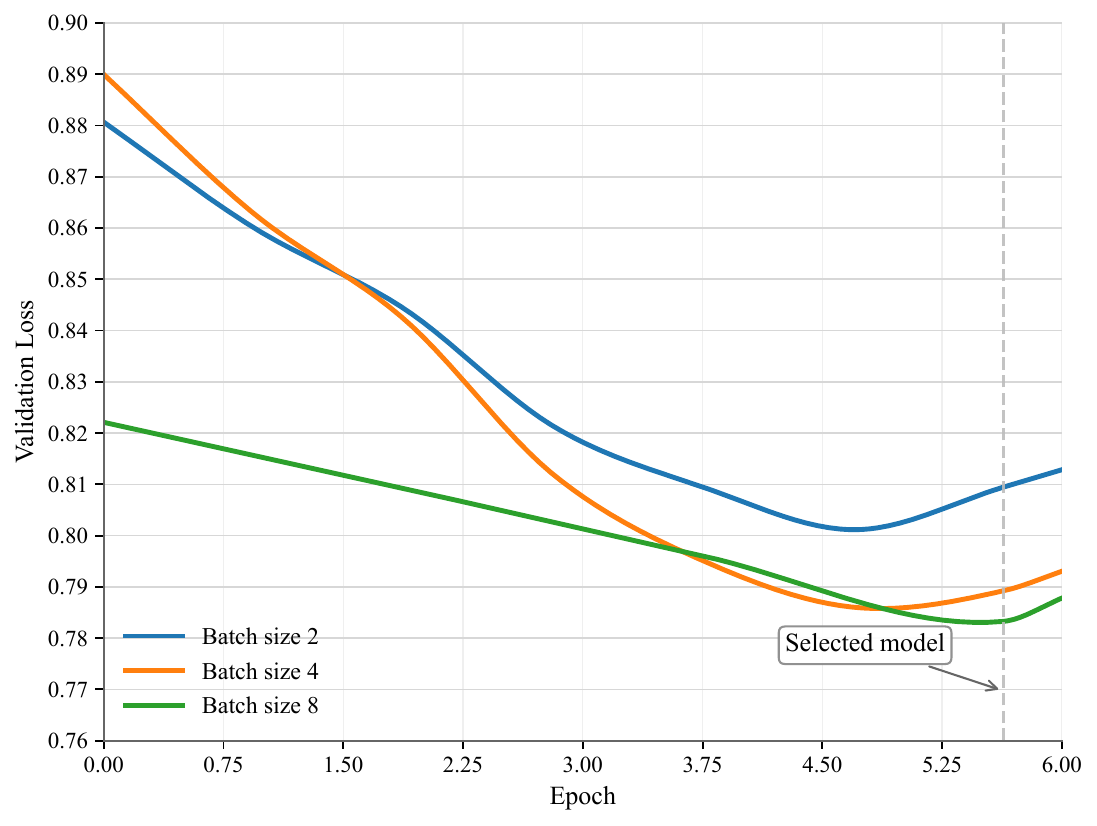}
        \caption{Comparison of validation loss curves under different training batch sizes.}
        \label{fig:batch}
    \end{minipage}
    \hfill
    \begin{minipage}[t]{0.48\textwidth}
        \centering
        \includegraphics[width=\linewidth]{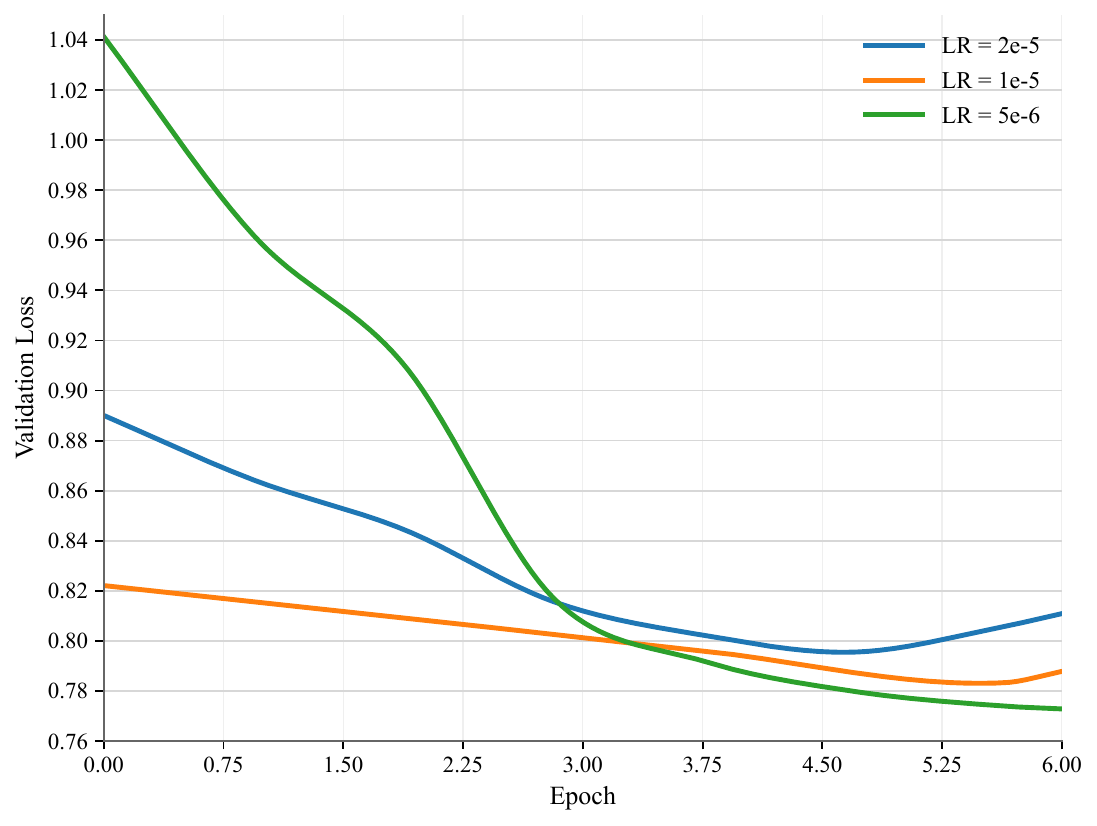}
        \caption{Comparison of validation loss curves under different learning rate.}
        \label{fig:lr}
    \end{minipage}
\end{figure}

As shown in Fig.~\ref{fig:batch}, the validation loss generally decreases in the early stage of training and then gradually plateaus or increases, indicating mild overfitting after approximately 5.8 epochs. Therefore, we use 6 epochs as the maximum training budget and select the checkpoint with the best validation loss rather than simply using the final checkpoint. This strategy avoids overfitting caused by excessive training while keeping the training procedure consistent across different settings. We observe that a batch size of 8 achieves the best validation performance among the tested settings. Smaller batch sizes lead to less stable optimization and slightly worse validation loss. Therefore, we adopt batch size 8 in the final configuration. 

As shown in Fig.~\ref{fig:lr}, \(2\times10^{-5}\) shows less stable convergence, while \(5\times10^{-6}\) achieves a slightly lower best validation loss but converges more slowly. In comparison, \(1\times10^{-5}\) provides a better trade-off between convergence speed and final performance. Based on these observations, we use batch size 8 and learning rate \(1\times10^{-5}\) in the main experiments.

\subsection{Comparison with LLM-MCTS Baseline}~\label{lab:llm_mcts_baseline}

We further introduce LLM-MCTS as an additional baseline in Tab.~\ref{tab:task3_sort_solids} to evaluate tasks requiring stronger semantic matching, such as Sort Solids. 
LLM-MCTS incorporates LLM-based semantic judgment and pruning into the search process, which substantially improves the performance of the original MCTS-based planner. 
Nevertheless, it still falls behind \sysname on more complex and challenging tasks, suggesting that semantic search alone cannot fully address long-horizon heterogeneous multi-robot collaboration.

\textit{LLM-MCTS}: The \textit{LLM-MCTS}~\cite{zhao2023large} integrates large language models into the Monte Carlo tree search procedure as both a commonsense world model and a heuristic policy. Specifically, the LLM-induced world model provides a prior belief over states for \textit{MCTS}, while the LLM-induced policy serves as \(\pi(a \mid h)\) in PUCT to guide action selection during simulation. This design leverages the commonsense knowledge encoded in LLMs to improve search efficiency and support more effective planning in large-scale, partially observable task environments.

\input{table/llm-mcts}

\subsection{COHERENT Failure Case Study}~\label{lab:coherent_fail_case}

To better understand the limitations of centralized coordination, we analyze a representative failed run from the centralized baseline. The task requires a heterogeneous team consisting of a fixed manipulator, a mobile manipulator, and a drone to place three target objects, \texttt{book\_0}, \texttt{fork\_0}, and \texttt{soap}, into the tray. The run terminates at the maximum step budget with \texttt{FinishedStep=30} and \texttt{PartSuccess=0.333}. The final task status shows that \texttt{book\_0}, \texttt{soap}, and the non-target object \texttt{toothpaste\_0} are in the tray, while the required object \texttt{fork\_0} is missing.

Tab.~\ref{tab:centralized_failure_case} summarizes the failure process. The key issue is that the centralized progress state incorrectly accepts \texttt{toothpaste\_0} as task progress, although it is not in the target object list. After \texttt{soap} is placed, the central coordinator repeatedly treats the task as nearly or fully completed and issues \texttt{wait} commands to the team. This prevents timely recovery for the missing \texttt{fork\_0}. When the coordinator later attempts to recover, it asks the fixed manipulator to pick \texttt{fork\_0}; however, the robot reports that \texttt{fork\_0} is not available in its local scene graph. Meanwhile, the drone repeatedly encounters obstacle-related failures, and the mobile manipulator performs additional exploration without locating the missing object.

This example reveals that the failure is not caused by a single low-level action error. Instead, it is caused by an incorrect centralized belief that propagates to the entire team. Since the central coordinator is responsible for maintaining the global task state and assigning actions to all robots, an error in this state can lead to coordinated but incorrect behavior, including unnecessary waiting, delayed search, and ineffective reallocation. This is a typical limitation of centralized coordination in long-horizon heterogeneous tasks: the centralized state acts as a single point of failure, and local robot feedback may not be incorporated quickly enough to correct the global decision.

By contrast, the proposed decentralized coordination strategy reduces the dependency on a single global progress estimate. Each robot maintains local observations and can share compact task-relevant information with teammates. For example, if one robot observes that \texttt{fork\_0} is absent from its local scene graph or that a non-target object has been placed, this information can directly influence local decisions and trigger reallocation, rather than waiting for a central coordinator to reconstruct the correct global state. This qualitative case therefore complements the aggregate results and helps explain why decentralized coordination is more robust in scenarios involving partial observability, object search, and heterogeneous robot capabilities.

\input{table/conhernet_case}

\subsection{LLM Cost Analysis}~\label{lab:llm_cost}

To disentangle whether the performance gain of \sysname comes from its decentralized structure or simply from a larger LLM compute budget, we further analyze the total LLM usage of \sysnameo and \textit{COHERENT}. Both methods use GPT-4o as the inference backbone in this comparison, so their token consumption and API call counts are directly comparable.

For each episode, we count the actual number of LLM invocations and the total token consumption. Each API request is counted as one call, and parallel calls from different robots are counted separately because they correspond to separate LLM invocations. For each call, token consumption includes both input tokens and output tokens. Therefore, the episode-level total token consumption is computed by summing input and output tokens over all LLM calls in the episode.

As shown in Tab.~\ref{tab:token_comparison}, \sysnameo uses more LLM calls than \textit{COHERENT} because it decomposes collaboration into multiple structured stages, including self-description (SD), task allocation and leadership bidding (TALB), leader election (LE), decentralized execution (HMRC), group discussion (GD), and leader-plan updating (LP). In contrast, \textit{COHERENT} relies on a centralized reasoning process and therefore uses fewer calls. However, each centralized call needs to aggregate information from multiple robots into a single context, leading to a longer per-call prompt. For example, \textit{COHERENT} requires around 6.7k--7.6k tokens per centralized reasoning call, while the main execution calls in \sysnameo use around 3.5k--3.9k tokens per call.

At the episode level, \sysnameo consumes approximately $1.8\times$ more total tokens than \textit{COHERENT}. This indicates that the improvement of \sysname is not obtained under a smaller overall token budget. Instead, \sysname uses additional LLM computation to support a more structured collaboration process. The key difference is that the computation is distributed across robot-specific reasoning, role allocation, leader election, periodic discussion, and leader-plan updates, rather than being concentrated in a single centralized prompt. This design reduces the single-call context burden and helps each robot preserve its local execution context, capability constraints, and feedback history.

These results suggest that the benefit of \sysname should be interpreted as a structural advantage in how LLM computation is organized, rather than as a reduction in total token usage. The framework trades higher total LLM usage for better task performance and robustness, especially in dynamic settings where centralized reasoning is more vulnerable to context overload and stale information. Reducing redundant LLM calls and improving token efficiency while preserving the coordination benefits of decentralized planning remain important directions for future work.

\input{table/token_comparison}

\subsection{Different Metrics Scaling across Robot Compositions}~\label{lab:comm_cost_composition}

We further analyze how robot-to-robot communication, LLM cost, coordination quality, and inference latency scale across different robot compositions. Although decentralization reduces the single-prompt context burden of centralized planning, it also introduces additional distributed planning, discussion, and coordination calls. Therefore, we evaluate scalability from four aspects: communication events, token consumption, coordination/alignment failures, and inference latency.

\paragraph{Communication and token cost} As shown in Fig.~\ref{fig:robocomb}, the average communication cost (CC) per robot remains relatively stable across robot configurations. This suggests that increasing the number of heterogeneous robots does not lead to a proportional increase in per-robot communication events within the evaluated team sizes. Therefore, the total communication cost grows approximately linearly with the number of robots in the studied regime, rather than super-linearly. This is because communication is selectively triggered through the \texttt{communicate} action and periodic discussion, rather than being performed at every step. Tab.~\ref{tab:token_robot_composition} compares the token consumption of \sysnameo and \textit{COHERENT} across tasks and robot compositions using the same GPT-4o backbone. \sysnameo generally consumes more total tokens, approximately twice that of \textit{COHERENT} in the evaluated settings. This overhead comes from the structured decentralized workflow, including self-description, task allocation \& leadership bidding, leader election, decentralized execution, group discussion, and leader-plan updating. In contrast, \textit{COHERENT} uses fewer centralized calls, but each call must aggregate multi-robot information into a single global context. Thus, the performance gain of \sysname should not be interpreted as lower total LLM usage; rather, \sysname uses additional LLM computation in a more structured way to preserve local execution context and maintain global coordination.

\paragraph{Coordination and alignment failures} To examine whether more complex robot compositions introduce coordination problems, we add diagnostic statistics on redundant and conflicting actions in Tab.~\ref{tab:fail_scaling}. For consistent comparison, redundant actions are counted only when a robot navigates to a previously visited region, while conflicting actions exclude \texttt{wait} and \texttt{communicate}. Across all reported settings, \sysnameo consistently produces fewer redundant actions than \textit{COHERENT}. On average, redundant actions decrease from 7.67 for \textit{COHERENT} to 3.64 for \sysnameo, corresponding to an approximately 52.6\% reduction. Conflict counts are low for both methods, but \sysnameo also reduces conflicts in several dynamic settings, especially ANC. These statistics show that coordination failures tend to increase when robot compositions become larger or more heterogeneous, since more robots introduce more local observations, action histories, capability constraints, and object-level dependencies. However, this increase is more pronounced for \textit{COHERENT}, especially in Static, ANC, and REC settings. In contrast, \sysnameo generally keeps redundant and conflicting actions at a lower level, suggesting that local embodiment-grounded planning and periodic synchronization help control the coordination overhead introduced by larger heterogeneous teams.

\paragraph{Latency analysis} We further report an inference-latency breakdown on the NVIDIA Jetson AGX Orin 32GB under different team sizes in Tab.~\ref{tab:team-scalability-orin-breakdown}. All values are averaged over per-robot LLM calls. Preprocessing time mainly consists of prefill and I/O overhead and is affected by the number of input tokens, while generation time is mainly determined by model inference speed and output length. The results show that the average total inference time remains at the seconds level and does not increase proportionally with team size. The generation speed is stable at around 26 tokens/s, indicating that backend throughput is not strongly affected by robot-team size in the evaluated regime. We also distinguish token cost from wall-clock latency. Total token consumption measures overall LLM compute cost, but it does not directly determine real execution time because multiple robots can invoke LLMs in parallel during decentralized planning and discussion. In a parallel synchronization stage, the effective waiting time is dominated by the slowest parallel agent call rather than the sum of all calls. Therefore, although \sysname incurs higher total LLM usage, parallel execution prevents wall-clock inference latency from growing proportionally with total token count. 

Overall, these results support the effectiveness of \sysname in the small-to-medium heterogeneous team regime evaluated in this paper, i.e., teams with 2--4 robot types. Scaling to substantially larger teams may introduce higher communication, token, latency, and coordination overhead, and will require message filtering, summary compression, asynchronous inference, hierarchical coordination, and more structured discussion protocols.

\input{table/token_robot_composition}

\input{table/latency_scaling}

\input{table/fail_scaling}

%% file: table/code.tex
\begin{algorithm}[htbp]
\caption{Heterogeneous Multi-Robot Collaboration}
\label{algo:collaboration}
\begin{algorithmic}[1]
\Require planning horizon $\mathcal{H}$, reflection interval $\Delta t$
\State $\mathcal{O}$ $\leftarrow$ env.reset()
\For{robot $i = 1 \  \textbf{to} \  n$}
    \State $d_i = g_i(f_i(o_i^0), \theta_i, \tau_i)$
    \Comment{self-description stage}
    \State $(p_i, g_i) = q_i(d_i, \tau_i)$
    \Comment{task allocations and leadership bidding stage}
\EndFor
\State $E = \arg\max_{k \in \{1,\dots,n\}} \sum_{i=1}^n v_i(p_k, g_k)$
\Comment{leader selection stage}
\For{step $t = 0 \  \textbf{to} \  \mathcal{H}$}
    \For{robot $i = 1 \  \textbf{to} \  n$}
        \State $a_i^t = \pi_i(f_i(o_i^t), h_i^t, fb_i^t, P_E^t)$
        \Comment{atomic action}
        \State $fb_i^{t+1}$ = env.step($a_i^t$)
        \Comment{limited feedback from env}
    \EndFor

    \If{$t \bmod \Delta t = 0$}
        \Comment{reflection stage}
        \For{robot $i = 1 \ \textbf{to} \ n$}
            \State $(rs_i^t, rp_i^t) = u_i(f_i(o_i^t), \overline{h}_i^t)$
            \Comment{local reflection summary and planning}
        \EndFor
        \State $P_E = \mathrm{Update}_E(RS^t, RP^t)$
        \Comment{leader integrates team reflection}
    \EndIf
\EndFor
\end{algorithmic}
\end{algorithm}

%% file: table/data_collection_code.tex
\begin{algorithm}[htbp]
\caption{Data Collection Pipeline}
\label{algo:data_collection}
\begin{algorithmic}[1]
\Require task set $\mathcal{T}$, environments $\mathcal{E}$, LLM policy $\pi_\theta$, human scripts $\mathcal{H}_s$, feasibility function $\Phi$
\State $\mathcal{D} \leftarrow \emptyset$ 
\Comment{initialize dataset}
\For{task $\tau \in \mathcal{T}$}
    \For{env $e \in \mathcal{E}$}
        \State $\mathcal{A}_h \leftarrow \mathrm{Execute}(\mathcal{H}_s, \tau, e)$
        \Comment{collect human-designed trajectory}
        \If{$\Phi(\mathcal{A}_h, e) = 1$} \Comment{task success}
            \State $\mathcal{D} \leftarrow \mathcal{D} \cup \mathrm{Transform}(\mathcal{A}_h)$
            \Comment{convert trajectory into step-wise decision samples}
        \EndIf
        \State $\mathcal{A}_\ell \sim \pi_n(\cdot \mid \tau, e)$
        \Comment{sample LLM-generated trajectory}
        \If{$\Phi(\mathcal{A}_\ell, e) = 1$}
            \State $\mathcal{A}_\ell \leftarrow \mathrm{Filter}(\mathcal{A}_\ell)$
            \Comment{human-in-the-loop filtering of suboptimal actions}
            \State $\mathcal{D} \leftarrow \mathcal{D} \cup \mathrm{Transform}(\mathcal{A}_\ell)$
            \Comment{add filtered trajectory as dataset entries}
        \EndIf
    \EndFor
\EndFor
\State \Return $\mathcal{D}$
\end{algorithmic}
\end{algorithm}

%% file: prompts/self_description.tex
\promptblock{\small 
\commenttext{==== System Prompt ====} \\
\textbf{Contexts:}
\begin{enumerate}
\item You are an intelligent robot capable of human-like reasoning and decision-making.
\item You must collaborate with heterogeneous robots to accomplish complex tasks.
\end{enumerate}
\textbf{Phase:}
Initial stage, where each robot introduces itself. \\
\textbf{CoT:}
Let's think step by step! \\
\commenttext{==== User Prompt ====} \\
\commenttext{==== Common Components Shared by All Robots ====} \\
Each robot introduces itself according to its configuration, capabilities, and understanding of the shared task. \\
\textbf{Task Objective and Context:}
\begin{enumerate}
\item The overall collaborative goal is \$\{target\_task\}.
\item Objects are scattered in an unknown indoor environment, requiring exploration and organization.
\item You should introduce yourself to help teammates \$\{teammates\} understand your role and abilities.
\end{enumerate}
\textbf{Output Response Format:}
\begin{enumerate}
\item \textbf{Thoughts:} step-by-step reasoning;
\item \textbf{Contents:} concise self-introduction for teammates.
\end{enumerate}
\commenttext{==== Distinct Components Specific to Each Robot ====} \\
\textcolor{light-yellow}{Mobile Manipulation Robot (Alice)} \\
\textbf{Role:}
\begin{enumerate}
\item You are an intelligent robot named \$\{name\}, configured with a wheeled chassis and a single manipulator arm. 
\item You possess the ability to navigate across the ground and perform manipulation tasks, including transporting various objects and opening hinged objects.
\end{enumerate}
\commenttext{==== Skills Omitted, specifically refer to Tab.~\ref{tab:atomic_action} ====} \\
\textbf{Skills:} ... \\
% \begin{enumerate}
% \item navigate(obj, pose); open(obj); pick(obj); place(obj, loc);
% \item move(dx, dy); communicate(content, robot); wait().
% \end{enumerate}
\textcolor{purple}{Manipulation Robot (Bob)} \\
\textbf{Role:}
\begin{enumerate}
\item You are an intelligent robot named \$\{name\}, and your configuration is a single robotic arm fixed on a desktop. 
\item You are capable of manipulating within a limited range.
\end{enumerate}
\textbf{Skills:} ... \\
% \begin{enumerate}
% \item pick(obj); place(obj, loc);
% \item communicate(content, robot); wait().
% \end{enumerate}
\textcolor{light-blue}{Mobile Robot (David)} \\
\textbf{Role:}
\begin{enumerate}
\item You are an intelligent robot named \$\{name\}, and your configuration is a wheeled chassis. 
\item You can navigate and move on the ground, cannot manipulate any objects and cannot open any hinged objects.
\end{enumerate}
\textbf{Skills:} ... \\
% \begin{enumerate}
% \item navigate(obj, pose);
% \item communicate(content, robot); wait().
% \end{enumerate}
\textcolor{pink}{Drone Robot (Lucy)} \\
\textbf{Role:}
\begin{enumerate}
\item You are an intelligent robot named \$\{name\}, configured as a quadrotor drone with a fixed suction gripper. 
\item You are capable of aerial navigation and manipulation.
\end{enumerate}
\textbf{Skills:} ...
% \begin{enumerate}
% \item navigate(obj, pose); pick(obj); place(obj, loc);
% \item communicate(content, robot); wait().
% \end{enumerate}
}

%% file: prompts/task_allocation_leadership_bidding.tex
\promptblock{\small 
\commenttext{==== System Prompt ====} \\
\textbf{Contexts:}
\begin{enumerate}
\item You are an intelligent robot that can think and make decisions like a human.
\item You need to cooperate with other robots of various configurations to complete complex and long-term tasks.
\end{enumerate}
\textbf{Phase:}
Now it is the second step of collaboration. \\
\textbf{Tasks:}
\begin{enumerate}
\item You need to propose a follow-up division of labor plan.
\item You need to propose a campaign speech to run for leader.
\end{enumerate}
\textbf{CoT:}
Let's think step by step! \\
\commenttext{==== User Prompt ====} \\
\textbf{Identity and Information:}
\begin{enumerate}
\item You are an intelligent robot named \$\{name\}. 
\item Below are the self-introductions from yourself and your collaborators: \$\{self\_introductions\}.
\end{enumerate}
\textbf{Plan Proposal and Leadership Campaign:}
\begin{enumerate}
\item Please analyze them carefully and thoroughly to develop your collaboration plan.
\item Reflect on your strengths from multiple perspectives and write a campaign speech to run for the leader role.
\end{enumerate}
\textbf{Principles:}
% \begin{enumerate}
% \item The proposed plan should arrange for robots to complete tasks in parallel as much as possible to improve the overall task efficiency.
% \item It should not only make full use of the common capabilities of multiple heterogeneous robots. For example, robots with navigation capabilities jointly explore the environment.
% \item But also give full play to the unique capabilities of individual robots efficiently. For example, robots with flight capabilities explore high altitudes, and robots with open capabilities open hinged objects.
% \end{enumerate}
\begin{enumerate}
\item The plan enables robots to work in parallel to maximize efficiency.
\item Utilize shared capabilities among heterogeneous robots, e.g., navigation robots jointly explore the environment.
\item Leverage unique abilities efficiently, e.g., flying robots explore high areas, opening robots handle hinged objects.
\end{enumerate}
\textbf{Output Response Format:}
\begin{enumerate}
\item Thoughts: think step by step to analyze the problem;
\item Contents: Include two parts: your proposed collaboration plan (aimed at improving teamwork), and your campaign speech for becoming the leader.
\end{enumerate}
}

%% file: prompts/leader_election.tex
\promptblock{\small 
\commenttext{==== System Prompt ====} \\
\textbf{Contexts:}
\begin{enumerate}
\item You are an intelligent robot capable of human-like thinking and decision-making.
\item You need to collaborate with other robots of various configurations to accomplish complex, long-term tasks.
\end{enumerate}
\textbf{Phase:}
Now it's the third step of collaboration. \\
\textbf{Tasks:}
% \begin{enumerate}
% \item You must carefully analyze the cooperation plans and leader candidacy speeches proposed by yourself and your collaborators in the previous step.
% \item Elect one person to serve as your leader objectively and fairly (You can recommend yourself).
% \end{enumerate}
\begin{enumerate}
\item Carefully analyze the collaboration plans and leadership proposals from all participants.
\item Objectively elect a leader (self-nomination allowed).
\end{enumerate}
\textbf{CoT:}
Let's think step by step! \\
\commenttext{==== User Prompt ====} \\
\textbf{Identity and Information:}
\begin{enumerate}
\item You are an intelligent robot named \$\{name\}.
\item Below are the collaboration plans and campaign speeches from yourself and other collaborators: \$\{plan\}.
\end{enumerate}
\textbf{Leader Election:}
Please analyze and judge fairly, justly, and objectively to elect a qualified leader. \\
\textbf{Output Response Format:}
\begin{enumerate}
\item Thoughts: think step by step to analyze the problem;
\item Reasons: state the reason for the choice made;
\item Leader: directly give the name of the selected leader.
\end{enumerate}
}

%% file: prompts/heterogeneous_multirobot_collaboration.tex
\promptblock{\small
\commenttext{==== System Prompt ====} \\
\commenttext{==== Common Components Shared by All Robots ====} \\
\commenttext{==== Omitted, same as the self-description user prompts ====} \\
\textbf{Role:} ... \\
\textbf{Skills:} ... \\
\textbf{Task Objective and Context:}
% \begin{enumerate}
% \item The overall task for everyone is: \$\{target\_task\}.
% \item All ingredients are scattered throughout an unknown indoor environment, which requires exploration and searching. Using a scene graph, you can locate the main furniture pieces, but you cannot know what items are on or inside the furniture. Based on the task requirements, the corresponding objects need to be placed in appropriate locations.
% \item You are expected to collaborate with other team members \$\{teammates\}, who possess different capabilities from yours, to accomplish the overall task.
% \item After previous group discussions and elections, \$\{leader\} will serve as the leader and has proposed the following collaboration and task allocation plan: \$\{plan\}. Therefore, \$\{different\_require\}.
% \end{enumerate}
\begin{enumerate}
\item The overall team task is: \$\{target\_task\}.
\item Ingredients are scattered in an unknown indoor environment. The scene graph shows furniture locations but not their contents. Based on task goal, objects must be placed.
\item Collaborate with teammates \$\{teammates\}, who have different capabilities, to complete the task.
\item \$\{leader\} is the elected leader and proposed the collaboration plan: \$\{plan\}. Thus, \$\{different\_require\}.
\end{enumerate}
\textbf{Output Response Format:}
\begin{enumerate}
\item Thoughts: think step by step to analyze the problem;
\item Contents: choose and execute only one action from the action functions above.
\end{enumerate}
\textbf{CoT:}
Let's think step by step! \\
\textbf{Examples:}
The following examples are provided for reference in decision-making. The related content involved has nothing to do with the actual task: [...] \\
\commenttext{==== Distinct Components Specific to Each Robot ====} \\
\textcolor{light-yellow}{Mobile Manipulation Robot (Alice)} \\
\textbf{Principles:}
% \begin{enumerate}
% \item Explore the environment as fully as possible, and navigate through the locations in the scene graph efficiently and without repetition in sequence.
% \item When encountering items related to the task, transport them back on time.
% \item When encountering an area that cannot be navigated, notify an assistant with relevant capabilities to explore;
% \item Monitor the changes in the latest task progress status and adjust the target object of the task on time.
% \item Respond to the requests of collaborators on time.
% \item In case of failure to grasp an object on a piece of furniture, first check other stand poses in the scene graph of this furniture, navigate and re-grasp, and secondly, rely on the move command to adjust the relative position between the robot base and the object.
% \item Remember to complete the task promptly and refrain from doing anything unrelated to the task.
% \end{enumerate}
\begin{enumerate}
\item Efficiently explore and navigate all locations in the scene graph without repetition.
\item Transport task-related items promptly.
\item When facing inaccessible areas, notify capable assistants.
\item Track task progress and adjust targets timely.
\item Respond promptly to collaborators' requests.
\item If grasp fails, try other stand poses or adjust base position.
\item Focus on completing the task without unrelated actions.
\end{enumerate}
\textcolor{purple}{Manipulation Robot (Bob)} \\
\textbf{Principles:}
% \begin{enumerate}
% \item Analyze the tasks and scene diagrams and prioritize completing your own tasks.
% \item Request assistance from other robots promptly for objects that are too far away or missing.
% \item Notify other collaborators of the task progress on time.
% \item Monitor the changes in the latest task progress status and adjust the target object of the task on time.
% \item Respond to the requests of collaborators on time.
% \item Remember to complete the task promptly and refrain from doing anything unrelated to the task.
% \end{enumerate}
\begin{enumerate}
\item Analyze tasks and scene graphs, prioritizing your work.
\item Request help promptly for distant or missing objects.
\item Notify collaborators of task progress timely.
\item Track progress changes and adjust targets as needed.
\item Respond promptly to collaborators' requests.
\item Focus on task completion without unrelated actions.
\end{enumerate}
\textcolor{light-blue}{Mobile Robot (David)} \\
\textbf{Principles:}
% \begin{enumerate}
% \item Explore the environment as fully as possible, and navigate the locations in the scene graph in sequence efficiently and without repetition.
% \item When encountering items related to the task, promptly notify the collaborators and request teammates with mobile operation capabilities to assist in carrying.
% \item When encountering areas that cannot be navigated, notify the assistants with relevant capabilities to conduct exploration.
% \item When encountering objects that need to be opened, promptly request the collaborators with relevant capabilities to assist in opening for further exploration.
% \item Monitor the changes in the latest task progress status and adjust the target object of the task on time.
% \item Respond promptly to the assistant's message requests.
% \item Remember to complete the task promptly and refrain from doing anything unrelated to the task.
% \end{enumerate}
\begin{enumerate}
\item Efficiently explore and navigate all locations in the scene graph without repetition.
\item Notify collaborators of task items and request mobile teammates for transport.
\item Notify capable assistants to explore inaccessible areas.
\item Request collaborators to open objects for exploration.
\item Track task progress and adjust targets timely.
\item Respond promptly to assistants' messages.
\item Focus on completing the task without unrelated actions.
\end{enumerate}
\textcolor{pink}{Drone Robot (Lucy)} \\
\textbf{Principles:}
% \begin{enumerate}
% \item Explore the environment as fully as possible, and navigate through the locations in the scene graph in an efficient and non-repetitive manner.
% \item When encountering items related to the task, transport them back on time.
% \item When encountering objects that need to be opened, promptly request the collaborators with relevant capabilities to assist in opening for further exploration.
% \item Monitor the changes in the latest task progress status and adjust the target object of the task on time.
% \item Respond to the requests of collaborators on time.
% \item Remember to complete the task promptly and refrain from doing anything unrelated to the task.
% \end{enumerate}
\begin{enumerate}
\item Efficiently explore and navigate all locations in the scene graph without repetition.
\item Transport task-related items promptly.
\item Request collaborators to open objects for exploration.
\item Track task progress and adjust targets timely.
\item Respond promptly to collaborators' requests.
\item Focus on task completion without unrelated actions.
\end{enumerate}
\commenttext{==== User Prompt ====} \\
\commenttext{==== Common Components Shared by All Robots ====} \\
\textbf{Task Status:} \\
Latest Task Progress Status: \$\{object finish\} has been placed in the tray. / \$\{object finish\} is placed on the cutting board one by one in stacking order. / The \$\{object finish\} are correctly placed on the panels of the corresponding colors. \\
\commenttext{==== Distinct Components Specific to Each Robot ====} \\
\textcolor{light-yellow}{Mobile Manipulation Robot (Alice)} \\
\textbf{Scene Graph:} \\
drawer: (pos: [...], ori: [...], state: close, stand: [...]), \\
cabinet: (...), sofa: (...), ... \\
\textbf{Robot Status:} \\
Current robot states: pos: [...], ori: [...]. The gripper is empty. \\
\textbf{Feedback History:} \\
The historical feedbacks, from oldest to newest, are as follows: [... Robot successfully reached the target book case-stand pose 0. [`book 0', `book 1'] are found on the book case ...] \\
\textbf{Action History:} \\
The historical actions, from oldest to newest, are as follows: [... navigate(book case, stand pose 0) ...] \\
\textbf{Receive Message History:} \\
The historical receive messages, from oldest to newest, are as follows: [... (Hi Alice! I am Bob. The book 0 is not on table 0. Please continue searching for it.) from Bob ...] \\ 
\commenttext{==== Omitted, same structure with robot-specific content ====} \\
\textcolor{purple}{Manipulation Robot (Bob)} ... \\
% \textbf{Scene Graph:} \\
% toothpaste: (pos: [...], ori: [...], pick: Yes), \\
% phone: (...), banana: (...), ... \\
% \textbf{Robot Status:} \\
% The robot is located on table 0 and its pos: [...], ori: [...]. The gripper is empty. \\
% \textbf{Feedback History:} \\
% The historical feedbacks, from oldest to newest, are as follows: [... The message has been successfully sent to `David' ...] \\
% \textbf{Action History:} \\
% The historical actions, from oldest to newest, are as follows: [... communicate('Hi, this is Bob. Please update me if either of you finds book 0 or apple 0 so we can proceed with the task ...] \\
% \textbf{Receive Message History:} \\
% The historical receive messages, from oldest to newest, are as follows: [... `Hi Bob, I have placed toothpaste 0 on table 0 for you to handle.' from Alice ...] \\
\textcolor{light-blue}{Mobile Robot (David)} ... \\
% \textbf{Scene Graph:} \\
% dishwasher: (pos: [...], ori: [...], state: close, stand: [...]), \\
% microwave: (...), fridge: (...), ... \\
% \textbf{Robot Status:} \\
% Current robot states: pos: [...], ori: [...]. \\
% \textbf{Feedback History:} \\
% The historical feedbacks, from oldest to newest, are as follows: [... Robot successfully reached the target cabinet 1-stand pose 0. No object was found on the cabinet 1 ...] \\
% \textbf{Action History:} \\
% The historical actions, from oldest to newest, are as follows: [... navigate(cabinet 1, stand pose 0) ...] \\
% \textbf{Receive Message History:} \\
% The historical receive messages, from oldest to newest, are as follows: [...] \\
\textcolor{pink}{Drone Robot (Lucy)} ...
% \textbf{Scene Graph:} \\
% tub: (pos: [...], ori: [...], stand: [...]), \\
% washbasin: (...), wall top: (...), ... \\
% \textbf{Robot Status:} \\
% Current robot states: pos: [...], ori: [...]. The gripper is empty. \\
% \textbf{Feedback History:} \\
% The historical feedbacks, from oldest to newest, are as follows: [... Robot successfully grasps the object book 0 ...] \\
% \textbf{Action History:} \\
% The historical actions, from oldest to newest, are as follows: [... pick(book 0) ...] \\
% \textbf{Receive Message History:} \\
% The historical receive messages, from oldest to newest, are as follows: [... `Hi David and Lucy, this is Bob. I need your assistance in locating toothpaste, as it is not visible on table 0. Please update me once you find it.' from Bob ...]
}

%% file: prompts/reflection_group_discussion.tex
\promptblock{\small 
\commenttext{==== Participants System Prompt ====} \\
\textbf{Contexts:} Please summarize and analyze the historical cooperation experiences and present your future collaboration plans. \\
\textbf{Phase:} Now it is the group discussion session of the heterogeneous robot collaboration phase. \\
\textbf{Principles:}
\begin{enumerate}
\item Compare the differences between the current task status and the target task status. 
\item Analyze the current scene graph content, historical feedback, action and message sequences, and summarize the successful experiences and failure lessons.
\end{enumerate}
\textbf{CoT:}
Let's think step by step! \\
\commenttext{==== Participants User Prompt ====} \\
\textbf{Output Response Format:}
\begin{enumerate}
\item Thoughts: think step by step to analyze the problem.
\item Summaries: summary, analysis of past, current task status.
\item Plans: plan for your subsequent tasks.
\end{enumerate}
\commenttext{==== Leader System Prompt ====} \\
\textbf{Contexts:}
% \begin{enumerate}
% \item You are an intelligent robot that can think and make decisions like a human. You need to cooperate with other robots to complete complex and long-term tasks.
% \item During the previous discussion process, you and other collaborators summarized your respective experiences and tasks, and put forward further task plans. Please conscientiously play the role of leader, summarize this information, analyze it, and put forward the final task plan.
% \end{enumerate}
\begin{enumerate}
\item You are an intelligent robot capable of human-like reasoning, collaborating with others on complex tasks.
\item As the leader, summarize and analyze collaborators’ experiences and tasks, then propose the final task plan.
\end{enumerate}
\textbf{Phase:} It is the leadership summary stage of group discussion. \\
\textbf{Principles:}
\begin{enumerate}
\item Assign specific tasks to each one, including yourself.
\item Ensure plan reflects current environment and object states.
\end{enumerate}
\textbf{CoT:}
Let's think step by step! \\
\commenttext{==== Leader User Prompt ====}
\begin{enumerate}
\item You are a smart robot named \$\{name\}, you are the leader.
\item The historical summaries and future plans of each one in the entire team received are as follows: \$\{latest plan\}.
\end{enumerate}
\textbf{Output Response Format:}
\begin{enumerate}
\item Thoughts: think step by step to analyze the problem.
\item Contents: output the latest heterogeneous robots plan.
\end{enumerate}
}

%% file: table/weight_sensitivity.tex
\subsection{Sensitivity of the Task Difficulty Metric}

The coefficients in Eq.~\ref{equ:difficulty_metric} are intended as an interpretable heuristic rather than an optimized fit. To examine the sensitivity of this design choice, we re-evaluated the matched subset of 38 tasks (114 runs) by varying the coefficients in Eq.~\ref{equ:difficulty_metric}. To isolate the effect of the object-count term, we scanned $\beta$ while keeping $\alpha=\gamma=(1-\beta)/2$. We then measured the consistency between the resulting difficulty score $D$ and the empirical task difficulty, where the latter is summarized by both the original performance metrics and the composite empirical hardness term $H_{\text{emp}} = z(1-\mathrm{PS}) + z(\mathrm{TS})$. Here, $z(\cdot)$ denotes z-score normalization, which standardizes each metric to have zero mean and unit variance across the evaluated tasks. This normalization makes $1-\mathrm{PS}$ and $\mathrm{TS}$ comparable despite their different scales, allowing them to be combined into a composite empirical hardness measure.

The results are summarized in Tab.~\ref{tab:weight_sensitivity}. Across $\beta \in [0.0, 0.8]$, the score remains consistently aligned with empirical task hardness: $\rho(D, H_{\text{emp}})$ stays within $[0.785, 0.807]$, and $\rho(D, \mathrm{TS})$ stays within $[0.856, 0.921]$. In addition, object count is more strongly associated with temporal steps than with partial success in the same matched subset: $\rho(\text{object count}, \mathrm{TS}) = 0.847$ versus $\rho(\text{object count}, \mathrm{PS}) = -0.571$. This supports our design choice of assigning a smaller coefficient to the object-count term than to the location-difficulty and collaboration-intensity terms.

\begin{figure}[h]
    \centering
    \includegraphics[width=\columnwidth]{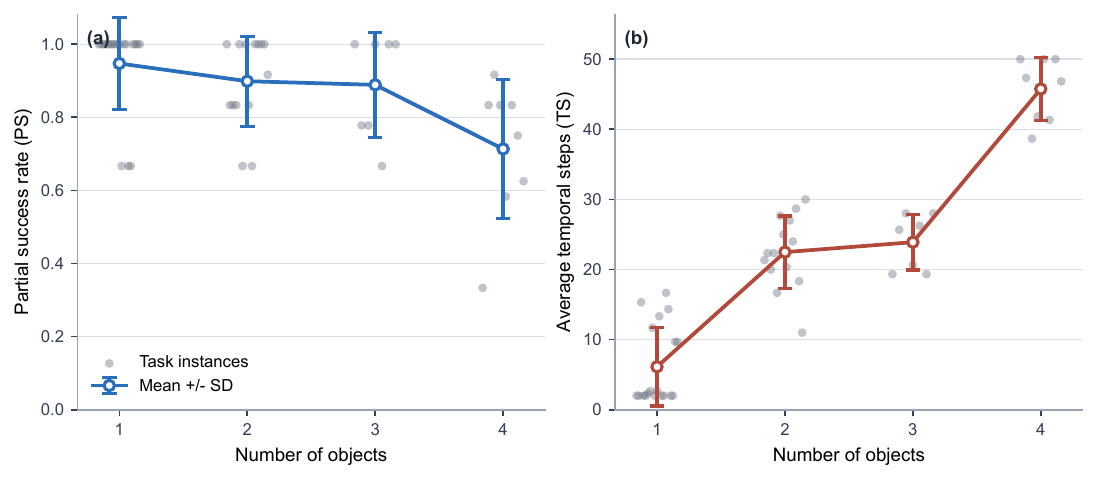}
    \caption{Correlation Plots between Object Quantity and the Task Metrics PS and TS}
    \label{fig:corr_plots}
\end{figure}

\begin{table}[htbp]
\centering
\caption{Sensitivity analysis of Eq.~(9) on the matched subset of pack objects task. The last column reports how many task labels change relative to the manuscript setting $(0.4, 0.2, 0.4)$.}
\label{tab:weight_sensitivity}
\renewcommand{\arraystretch}{1.0}
\resizebox{\columnwidth}{!}{
\begin{tabular}{c c c c c c c c}
\toprule
$\alpha$ & $\beta$ & $\gamma$ & $\rho(D,\mathrm{PS})$ & $\rho(D,\mathrm{TS})$ & $\rho(D,H_{\text{emp}})$ & pairwise acc. & label changes \\
\midrule
0.5 & 0.0 & 0.5 & -0.589 & 0.856 & 0.785 & 0.870 & 0/38 \\
0.4 & 0.2 & 0.4 & -0.602 & 0.913 & 0.801 & 0.851 & baseline \\
0.3 & 0.4 & 0.3 & -0.602 & 0.915 & 0.803 & 0.852 & 18/38 \\
0.2 & 0.6 & 0.2 & -0.604 & 0.921 & 0.807 & 0.856 & 20/38 \\
0.1 & 0.8 & 0.1 & -0.604 & 0.921 & 0.807 & 0.856 & 25/38 \\
0.0 & 1.0 & 0.0 & -0.571 & 0.847 & 0.714 & 0.885 & 25/38 \\
\bottomrule
\end{tabular}}
\end{table}

%% file: prompts/fail_log.tex
\promptblock{\small 
\commenttext{==== Reasoning Process\&Action Decision of Alice ====} \\
\commenttext{==== Step 10 ====} \\
\textbf{Thoughts:}
Alice, the robot, has successfully identified the location of `fork\_0' inside the open drawer. The task requires Alice to retrieve items like \textbf{\textcolor{pink}{`fork\_0'}} and `book\_0' and bring them closer to Bob's range for placement in the tray. \textbf{\textcolor{pink}{Since `fork\_0' is already located and accessible, Alice should prioritize picking it up to ensure efficient task progression.}} This action aligns with the collaboration plan, where Alice is responsible for retrieving items within her reach. Furthermore, \textbf{\textcolor{pink}{picking `fork\_0' is a direct contribution to the task}}, as it will eventually need to be placed in the tray. Alice's manipulator arm is capable of handling this task, and the gripper is currently empty, making it ready for object pickup. \textbf{\textcolor{pink}{Therefore, the decision to execute the `pick(fork\_0)' action is logical and appropriate to advance the task at hand.}}\\
\textbf{Contents:}
pick(fork\_0) \\
\commenttext{==== Step 11 ====} \\
\textbf{Thoughts:}
To decide the next action, I need to consider the current state, historical actions, and feedback. The task is to place `book\_0' and `bowl\_0' into the tray. From the historical actions and feedback, I've explored several locations but have not found `book\_0' or `bowl\_0'. However, \textbf{\textcolor{dark-green}{I have found `fork\_0' in the drawer, which is not part of the task.}} The pick attempt for `fork\_0' failed due to a significant position difference in the x-axis direction. This indicates that the robot's position needs adjustment to successfully pick the item. \textbf{\textcolor{dark-green}{Since `fork\_0' is not part of the task, I should focus on finding `book\_0' and `bowl\_0'.}} The next logical step is to continue exploring the environment efficiently without repetition. The scene graph shows various locations, and I have not yet explored table\_3. Navigating to table\_3 could potentially lead to discovering `book\_0' or `bowl\_0'. Therefore, I should navigate to table\_3, stand\_pose\_0 to continue the exploration. \\
\textbf{Contents:}
navigate(table\_3, stand\_pose\_0) \\
\commenttext{==== Step 23 ====} \\
\textbf{Thoughts:}
Alice needs to continue exploring the environment to locate `bowl\_0', as `book\_0' has already been placed in the tray. The scene graph provides various locations, and Alice has already explored several places like the drawer, cabinet\_1, fridge, and tables, without finding `bowl\_0'. Lucy's leadership plan emphasizes aerial exploration, but Alice's ground navigation capability complements this by exploring areas Lucy might not reach easily. Alice's current position is near table\_2, where `ham\_0' was found, but since `bowl\_0' hasn't been located yet, Alice should continue exploring other places. The historical actions show that Alice has been trying to pick fork\_0 unsuccessfully, indicating a need for precise navigation or positioning to find bowl\_0. \textbf{\textcolor{pink}{The decision to navigate to table\_3, stand\_pose\_0, is logical as it hasn't been explored yet and could potentially hold `bowl\_0'.}} This action aligns with the principles of efficient exploration and timely transportation of task-related items. \\
\textbf{Contents:}
navigate(table\_3, stand\_pose\_0) \\
\commenttext{==== Step 29 ====} \\
\textbf{Thoughts:}
The task is to find `bowl\_0' and `book\_0' and place them in the tray. `book\_0' has already been placed in the tray, so the focus is now on finding `bowl\_0'. Based on the scene graph and historical actions, I have explored several locations, including the fridge, dishwasher, and others. `apple\_0' was found in the fridge, but no relevant items were found in other locations. The historical feedback indicates that `bowl\_0' has not been located yet. Lucy is responsible for aerial exploration and has requested assistance in opening the fridge, which has already been done. Bob is handling items within his reach and has been informed about `fork\_0'. Since `bowl\_0' has not been found and there are still unexplored locations, I need to continue the exploration. The next logical step is to navigate to the next unexplored location, which is the book\_case. This is a logical place to search for `book\_0', although it has already been placed in the tray, ensuring no further relevant items are missed. \\
\textbf{Contents:}
navigate(book\_case, stand\_pose\_0)
}

%% file: table/memory.tex
\begin{table*}[htbp] 
\centering
\caption{Performance Comparison under Different Memory Queue Lengths on the Pack Objects Task} 
\label{tab:memory_length} 
\renewcommand{\arraystretch}{1.2} 

\resizebox{\textwidth}{!}{ 
\large
\begin{tabular}{
>{\centering\arraybackslash}p{6.5cm} 
*{10}{>{\centering\arraybackslash}p{1.2cm}}
} 

\toprule

\multirow{2}{*}{\textbf{Method}} 
& \multicolumn{5}{c}{\textbf{Static}}
& \multicolumn{5}{c}{\textbf{CTO}} \\ 
\cmidrule(lr){2-6} \cmidrule(lr){7-11}
& SUCC$\uparrow$ & PS$\uparrow$ & TS$\downarrow$ & AS$\downarrow$ & CC$\downarrow$ 
& SUCC$\uparrow$ & PS$\uparrow$ & TS$\downarrow$ & AS$\downarrow$ & CC$\downarrow$ \\ 
\midrule

COHERENT ($M_l=10$)
& 0.3148 & 0.5926 & 26.0185 & 14.4367 & --
& 0.4074 & 0.6451 & 23.9630 & \textbf{14.8457} & -- \\ 

COHERENT ($M_l=15$)
& 0.0556 & 0.4167 & 28.6667 & 18.4738 & --
& 0.2407 & 0.4506 & 26.4259 & 18.1250 & -- \\

CoELA ($M_l=10$)
& 0.4444 & 0.6636 & 25.2037 & 18.6867 & 3.2840
& 0.4815 & 0.6821 & 24.9444 & 19.2377 & 3.2901 \\

CoELA ($M_l=15$)
& 0.3889 & 0.6451 & 25.6481 & 18.2747 & 4.0031
& 0.5000 & 0.6451 & 25.1852 & 16.6343 & 3.0293 \\

\sysnameo ($M_l=10$)
& 0.6482 & 0.8179 & 23.1296 & 18.0926 & 5.1667
& \textbf{0.7091} & \textbf{0.8606} & \textbf{21.5273} & 18.0625 & 4.7967 \\

\sysnameo ($M_l=15$)
& \textbf{0.8889} & 0.9506 & 21.1111 & 14.7392 & 2.4614
& 0.5556 & 0.6296 & 26.5741 & 18.9691 & 5.6173 \\

\sysnameq ($M_l=10$)
& 0.8000 & 0.8788 & \textbf{20.1455} & 13.8796 & 2.1955
& 0.6909 & 0.8364 & 23.4909 & 15.6333 & 2.8711 \\ 

\sysnameq ($M_l=15$)
& 0.8519 & 0.9568 & 20.2963 & \textbf{13.5046} & \textbf{1.8935} 
& 0.5926 & 0.8272 & 23.9630 & 15.7114 & \textbf{2.8534} \\ 

\bottomrule

\end{tabular}} 
\end{table*}

%% file: table/latency.tex
\begin{table*}[htbp]
\centering
\small
\caption{Inference, Local Throughput, and ROS2 Communication Distributions on Real Robot Platform}
\label{tab:latency-summary}
\resizebox{\textwidth}{!}{
\begin{tabular}{
>{\centering\arraybackslash}p{3.0cm}
>{\centering\arraybackslash}p{4.0cm}
>{\centering\arraybackslash}p{1.5cm}
>{\centering\arraybackslash}p{1.4cm}
>{\centering\arraybackslash}p{1.4cm}
>{\centering\arraybackslash}p{1.4cm}
>{\centering\arraybackslash}p{1.4cm}
}
\toprule
\textbf{Platform / Channel} & \textbf{Metric} & \textbf{Unit} 
& \textbf{Min} & \textbf{Mean} & \textbf{Max} & \textbf{P95} \\
\midrule

RTX 5060 Ti & Input tokens & tokens 
& 32.00 & 2221.80 & 3261.00 & 3213.50 \\

RTX 5060 Ti & Output tokens & tokens 
& 170.00 & 321.95 & 441.00 & 420.10 \\

RTX 5060 Ti & Inference latency & ms 
& 2198.40 & 4181.70 & 5672.85 & 5387.38 \\

RTX 5060 Ti & Generation speed & tokens/s 
& 74.97 & 76.95 & 78.68 & 78.37 \\

\midrule

Jetson AGX Orin & Input tokens & tokens 
& 32.00 & 2277.80 & 3313.00 & 3270.25 \\

Jetson AGX Orin & Output tokens & tokens 
& 178.00 & 335.20 & 512.00 & 434.10 \\

Jetson AGX Orin & Inference latency & ms 
& 6522.98 & 12401.69 & 20069.66 & 15946.57 \\

Jetson AGX Orin & Generation speed & tokens/s 
& 25.51 & 27.07 & 27.63 & 27.62 \\

\midrule

API (gpt-4o) & Inference latency & ms 
& 2453.52 & 6067.07 & 14096.33 & 8578.06 \\

ROS2 & Communication latency & ms 
& 0.79 & 1.13 & 4.56 & 1.36 \\

\bottomrule
\end{tabular}}
\end{table*}

%% file: table/statistical_comparison.tex
\begin{table}[htbp]
\centering
\scriptsize
\setlength{\tabcolsep}{4pt}
\renewcommand{\arraystretch}{1.2}
\caption{Paired statistical comparison between \sysnameq and \sysnameo}
\label{tab:statistical_comparison}
\resizebox{\linewidth}{!}{
\begin{tabular}{>{\centering\arraybackslash}p{2.4cm} >{\centering\arraybackslash}p{1.2cm} >{\centering\arraybackslash}p{2.8cm} >{\centering\arraybackslash}p{3.0cm} >{\centering\arraybackslash}p{1.4cm}}
\toprule
\textbf{Setting} & \textbf{Metric} & \textbf{Mean Difference} & \textbf{95\% CI} & \textbf{\textit{p}-value} \\
\midrule
\multirow{5}{*}{Static Tasks}
& SUCC & $+0.0498$ & $[0.0017,\ 0.1000]$ & $0.0497$ \\
& PS   & $+0.0082$ & $[-0.0234,\ 0.0385]$ & $0.6087$ \\
& TS   & $-1.7147$ & $[-3.1424,\ -0.3214]$ & $0.0191$ \\
& AS   & $-2.9642$ & $[-4.3650,\ -1.6735]$ & $< 10^{-4}$ \\
& CC   & $-2.3648$ & $[-3.1331,\ -1.7126]$ & $< 10^{-4}$ \\
\midrule
\multirow{5}{*}{Dynamic Tasks}
& SUCC & $+0.0870$ & $[0.0191,\ 0.1558]$ & $0.0167$ \\
& PS   & $+0.0518$ & $[0.0091,\ 0.0928]$ & $0.0198$ \\
& TS   & $-3.6850$ & $[-5.3971,\ -2.0760]$ & $< 10^{-4}$ \\
& AS   & $-4.0456$ & $[-5.1974,\ -2.9488]$ & $< 10^{-4}$ \\
& CC   & $-2.5799$ & $[-3.0479,\ -2.1306]$ & $< 10^{-4}$ \\
\bottomrule
\end{tabular}}
\end{table}

%% file: table/select_leader.tex
\begin{table*}[htbp] 
\centering
\caption{Robustness Analysis under Degraded Leader Election with 50\% Random Votes on the Pack Objects Task.} 
\label{tab:random_votes} 
\renewcommand{\arraystretch}{1.1} 

\resizebox{\textwidth}{!}{ 
\large
\begin{tabular}{
>{\centering\arraybackslash}p{6.5cm} 
*{10}{>{\centering\arraybackslash}p{1.2cm}}
} 

\toprule

\multirow{2}{*}{\textbf{Method}} 
& \multicolumn{5}{c}{\textbf{Static}}
& \multicolumn{5}{c}{\textbf{CTO}} \\ 
\cmidrule(lr){2-6} \cmidrule(lr){7-11}
& SUCC$\uparrow$ & PS$\uparrow$ & TS$\downarrow$ & AS$\downarrow$ & CC$\downarrow$ 
& SUCC$\uparrow$ & PS$\uparrow$ & TS$\downarrow$ & AS$\downarrow$ & CC$\downarrow$ \\ 
\midrule

\rowcolor[gray]{.8}
\multicolumn{11}{c}{\textbf{2 Types of Robot}} \\ 

\sysnameq w/ 50\% RV
& 0.7778 & 0.8796 & 22.7222 & 13.7500 & 2.1944
& \textbf{0.7222} & 0.8241 & \textbf{23.3333} & \textbf{14.5556} & \textbf{2.4444} \\ 

\sysnameq 
& \textbf{0.8333} & \textbf{0.8889} & \textbf{20.2222} & \textbf{13.0556} & \textbf{2.0000} 
& \textbf{0.7222} & \textbf{0.8519} & 25.4444 & 15.0556 & 3.0000 \\ 

\midrule

\rowcolor[gray]{.8}
\multicolumn{11}{c}{\textbf{3 Types of Robot}} \\

\sysnameq w/ 50\% RV
& 0.7037 & \textbf{0.8889} & 21.4815 & 15.1410 & 2.48 
& \textbf{0.5556} & \textbf{0.8457} & \textbf{24.2222} & \textbf{16.9615} & \textbf{3.1538} \\ 

\sysnameq 
& \textbf{0.7143} & 0.8333 & \textbf{21.4286} & \textbf{15.0741} & \textbf{2.3929} 
& \textbf{0.5556} & 0.7654 & 24.8889 & 17.2716 & 3.2267 \\ 

\midrule

\rowcolor[gray]{.8}
\multicolumn{11}{c}{\textbf{4 Types of Robot}} \\

\sysnameq w/ 50\% RV
& \textbf{1.0000} & \textbf{1.0000} & \textbf{15.2222} & \textbf{11.8056} & \textbf{1.8056}
& \textbf{1.0000} & \textbf{1.0000} & 17.1111 & 12.9444 & 1.9444 \\ 

\sysnameq 
& \textbf{1.0000} & \textbf{1.0000} & 16.0000 & 11.9444 & 1.9722
& \textbf{1.0000} & \textbf{1.0000} & \textbf{16.2000} & \textbf{12.2500} & \textbf{1.7500} \\ 

\bottomrule

\end{tabular}} 
\end{table*}

%% file: table/supervision.tex
\begin{table}[H] 
\centering
\caption{Performance comparison of different supervision strategies on the static Pack Objects task under medium difficulty.} 
\label{tab:supervision} 
\renewcommand{\arraystretch}{1.0} 
\setlength{\tabcolsep}{4pt}

\resizebox{\columnwidth}{!}{ 
\begin{tabular}{
>{\centering\arraybackslash}p{6.5cm} 
*{5}{>{\centering\arraybackslash}p{1.5cm}}
} 
\toprule

\multirow{2}{*}{\textbf{Method}} 
& \multicolumn{5}{c}{\textbf{Static}} \\ 
\cmidrule(lr){2-6}
& SUCC$\uparrow$ & PS$\uparrow$ & TS$\downarrow$ & AS$\downarrow$ & CC$\downarrow$ \\ 
\midrule

\rowcolor[gray]{.85}
\multicolumn{6}{c}{\textbf{2 Types of Robot}} \\ 

\sysnameq\ w/ Reasoning-only
& 0.0000 & 0.0185 & 30.0000 & N/A & N/A \\

\sysnameq\ w/ Action-only
& 0.0000 & 0.0648 & 30.0000 & 20.3056 & 6.0556 \\

\sysnameq
& \textbf{0.8333} & \textbf{0.8889} & \textbf{20.2222} & \textbf{13.0556} & \textbf{2.0000} \\ 

\midrule

\rowcolor[gray]{.85}
\multicolumn{6}{c}{\textbf{3 Types of Robot}} \\

\sysnameq\ w/ Reasoning-only
& 0.0000 & 0.0000 & 30.0000 & N/A & N/A \\ 

\sysnameq\ w/ Action-only
& 0.0370 & 0.2160 & 30.0000 & 21.8148 & 1.8025 \\ 

\sysnameq
& \textbf{0.7143} & \textbf{0.8333} & \textbf{21.4286} & \textbf{15.0741} & 2.3929 \\ 

\midrule

\rowcolor[gray]{.85}
\multicolumn{6}{c}{\textbf{4 Types of Robot}} \\

\sysnameq\ w/ Reasoning-only
& 0.0000 & 0.0000 & 30.0000 & N/A & N/A \\ 

\sysnameq\ w/ Action-only
& 0.1111 & 0.2037 & 30.0000 & 22.0556 & 0.3056 \\ 

\sysnameq
& \textbf{1.0000} & \textbf{1.0000} & \textbf{16.0000} & \textbf{11.9444} & 1.9722 \\ 

\bottomrule
\end{tabular}} 

\begin{flushleft}
\scriptsize
N/A indicates that the reasoning-only variant is a diagnostic setting without direct supervision on executable atomic actions, making AS and CC not reliably comparable. Lower CC values for failed action-only runs do not necessarily indicate better communication efficiency, since little effective collaboration is achieved when tasks fail.
\end{flushleft}
\end{table}

%% file: table/generalization.tex
\begin{table}[H] 
\centering
\caption{Performance Comparison under Unseen Target Objects (UTO) on the Pack Objects Task} 
\label{tab:unseen_obj} 
\renewcommand{\arraystretch}{1.05} 

\resizebox{\columnwidth}{!}{ 
\begin{tabular}{
>{\centering\arraybackslash}p{6.0cm} 
*{5}{>{\centering\arraybackslash}p{1.2cm}}
} 

\toprule

\multirow{2}{*}{\textbf{Method}} 
& \multicolumn{5}{c}{\textbf{Static}} \\ 
\cmidrule(lr){2-6}
& SUCC$\uparrow$ & PS$\uparrow$ & TS$\downarrow$ & AS$\downarrow$ & CC$\downarrow$ \\ 
\midrule

\sysnameq w/ UTO
& 0.7407 & 0.8704 & 22.0185 & 15.0988 & 2.4614 \\ 

\sysnameq
& 0.8000 & 0.8788 & 20.1455 & 13.8796 & 2.1955 \\ 

\bottomrule

\end{tabular}} 
\end{table}

%% file: table/real_failure.tex
\begin{table*}[htbp]
\centering
\small
\setlength{\tabcolsep}{4pt}
\renewcommand{\arraystretch}{1.15}
\caption{Failure-mode breakdown in real-world deployment.}
\label{tab:real_failure_breakdown}
\resizebox{\linewidth}{!}{
\begin{tabular}{
>{\centering\arraybackslash}p{2.6cm}
>{\centering\arraybackslash}p{4.8cm}
>{\centering\arraybackslash}p{7.2cm}
>{\centering\arraybackslash}p{4.2cm}
}
\toprule
\textbf{Error Type} 
& \textbf{Source} 
& \textbf{Typical Failure} 
& \textbf{Effect} \\
\midrule

Perception Error 
& Object detection, segmentation, grasp-pose estimation, visual feedback 
& Missing or confusing target objects; incomplete mask; unstable AnyGrasp prediction; incorrect GPT-4o success judgment 
& Incorrect task-state update or infeasible manipulation \\

\midrule

Planning Error 
& LLM task planning, role allocation, recovery reasoning 
& Wrong target selection; redundant communication; repeated stale actions; poor recovery after feedback 
& Increased task steps or failure to complete long-horizon goals \\

\midrule

Execution Error 
& Navigation, grasp execution, placement, low-level control 
& Failed navigation, grasp slip, inaccurate placement, motion execution deviation 
& Physical failure despite a reasonably high-level plan \\

\bottomrule
\end{tabular}
}
\end{table*}

%% file: table/different_T_CTO.tex
\begin{table}[htbp] 
\centering
\caption{Performance Comparison under Different Trigger Timings on the Pack Objects Task (CTO)} 
\label{tab:dT_CTO} 
\renewcommand{\arraystretch}{1.0} 

\resizebox{\columnwidth}{!}{ 
\large
\begin{tabular}{
>{\centering\arraybackslash}p{4.5cm} 
*{5}{>{\centering\arraybackslash}p{1.4cm}}
} 

\toprule

\multirow{2}{*}{\textbf{Method}} 
& \multicolumn{5}{c}{\textbf{CTO}} \\ 
\cmidrule(lr){2-6}
& SUCC$\uparrow$ & PS$\uparrow$ & TS$\downarrow$ & AS$\downarrow$ & CC$\downarrow$ \\ 
\midrule

\rowcolor[gray]{.8}
\multicolumn{6}{c}{\textbf{T=3}} \\ 

COHERENT
& 0.1481 & 0.2685 & 27.3333 & \textbf{11.3534} & -- \\

CoELA (T=3)
& 0.2222 & 0.4198 & 27.3704 & 13.0278 & \textbf{2.3735} \\

\sysnameo
& 0.5000 & 0.7377 & 25.2407 & 18.3426 & 5.1497 \\

\sysnameq
& \textbf{0.7593} & \textbf{0.9105} & \textbf{22.0926} & 15.2994 & 2.4660 \\

\rowcolor[gray]{.8}
\multicolumn{6}{c}{\textbf{T=5}} \\ 

COHERENT
& 0.4074 & 0.6451 & 23.9630 & \textbf{14.8457} & -- \\

CoELA
& 0.4815 & 0.6821 & 24.9444 & 19.2377 & 3.2901 \\

\sysnameo
& \textbf{0.7091} & \textbf{0.8606} & \textbf{21.5273} & 18.0625 & 4.7967 \\

\sysnameq
& 0.6909 & 0.8364 & 23.4909 & 15.6333 & \textbf{2.8711} \\

\rowcolor[gray]{.8}
\multicolumn{6}{c}{\textbf{T=10}} \\ 

COHERENT
& 0.2222 & 0.4167 & 25.8333 & 16.4275 & -- \\

CoELA
& 0.3333 & 0.5370 & 26.2037 & 18.3380 & 3.5602 \\

\sysnameo
& 0.7037 & 0.8704 & 23.6296 & 17.8397 & 5.4967 \\

\sysnameq
& \textbf{0.7407} & \textbf{0.9352} & \textbf{23.1667} & \textbf{15.7485} & \textbf{2.8441} \\ 

\bottomrule

\end{tabular}} 
\end{table}

%% file: table/different_T_IRZ.tex
\begin{table}[t] 
\centering
\caption{Performance Comparison under Different Numbers of Restricted Zones on the Pack Objects Task (IRZ)} 
\label{tab:dN_IRZ} 
\renewcommand{\arraystretch}{1.0} 

\resizebox{\columnwidth}{!}{ 
\large
\begin{tabular}{
>{\centering\arraybackslash}p{4.5cm} 
*{5}{>{\centering\arraybackslash}p{1.5cm}}
} 

\toprule

\multirow{2}{*}{\textbf{Method}} 
& \multicolumn{5}{c}{\textbf{IRZ}} \\ 
\cmidrule(lr){2-6}
& SUCC$\uparrow$ & PS$\uparrow$ & TS$\downarrow$ & AS$\downarrow$ & CC$\downarrow$ \\ 
\midrule

\rowcolor[gray]{.8}
\multicolumn{6}{c}{\textbf{n=1}} \\ 

COHERENT
& 0.7500 & 0.8733 & 15.2292 & \textbf{10.0087} & -- \\

CoELA
& 0.7708 & 0.8785 & \textbf{13.8958} & 11.8559 & 1.7465 \\

\sysnameo
& 0.5714 & 0.8571 & 23.2407 & 18.3426 & 5.1497 \\

\sysnameq
& \textbf{1.0000} & \textbf{1.0000} & 20.2222 & 14.6296 & \textbf{1.6296} \\ 

\rowcolor[gray]{.8}
\multicolumn{6}{c}{\textbf{n=2}} \\ 

COHERENT
& 0.4762 & 0.6687 & 22.0714 & 12.8611 & -- \\

CoELA
& 0.7143 & 0.8254 & 18.0476 & 14.8790 & 2.3988 \\

\sysnameo
& 0.8750 & 0.9375 & 19.5000 & 15.4635 & 4.8021 \\

\sysnameq
& \textbf{0.9444} & \textbf{0.9815} & \textbf{16.4444} & \textbf{11.6667} & \textbf{1.6111} \\

\rowcolor[gray]{.8}
\multicolumn{6}{c}{\textbf{n=3}} \\ 

COHERENT
& 0.7778 & 0.8796 & 15.1667 & \textbf{9.0417} & -- \\

CoELA
& 0.7778 & 0.8333 & 18.8333 & 15.4306 & 2.2778 \\

\sysnameo
& \textbf{1.0000} & \textbf{1.0000} & \textbf{13.6667} & 10.3333 & 1.7500 \\

\sysnameq
& \textbf{1.0000} & \textbf{1.0000} & 15.2222 & 11.8056 & \textbf{1.2500} \\ 

\bottomrule

\end{tabular}} 
\end{table}

%% file: table/different_T_ANC_REC.tex
\begin{table*}[htbp] 
\centering
\caption{Performance Comparison under Different Trigger Timings on the Pack Objects Task (ANC \& REC)} 
\label{tab:dT_ANC_REC} 
\renewcommand{\arraystretch}{1.0} 

\resizebox{\textwidth}{!}{ 
\large
\begin{tabular}{
>{\centering\arraybackslash}p{5.0cm} 
*{10}{>{\centering\arraybackslash}p{1.2cm}}
} 

\toprule

\multirow{2}{*}{\textbf{Method}} 
& \multicolumn{5}{c}{\textbf{ANC}}
& \multicolumn{5}{c}{\textbf{REC}} \\ 
\cmidrule(lr){2-6} \cmidrule(lr){7-11}
& SUCC$\uparrow$ & PS$\uparrow$ & TS$\downarrow$ & AS$\downarrow$ & CC$\downarrow$ 
& SUCC$\uparrow$ & PS$\uparrow$ & TS$\downarrow$ & AS$\downarrow$ & CC$\downarrow$ \\ 
\midrule

\rowcolor[gray]{.8}
\multicolumn{11}{c}{\textbf{T=3}} \\ 

COHERENT
& 0.0278 & 0.3843 & 29.9167 & 19.9907 & --
& 0.1111 & 0.2269 & 28.6111 & 7.3796 & -- \\ 

CoELA
& 0.0833 & 0.5231 & 29.0278 & 21.1389 & 5.0278
& 0.1389 & 0.2778 & 27.8333 & \textbf{7.3634} & \textbf{1.2685} \\

\sysnameo
& 0.5000 & 0.7176 & 24.8611 & \textbf{13.3796} & 3.1898
& 0.2500 & 0.5000 & 26.4167 & 13.9306 & 3.6551 \\

\sysnameq
& \textbf{0.8056} & \textbf{0.9167} & \textbf{19.5556} & 13.3981 & \textbf{2.3148}
& \textbf{0.5833} & \textbf{0.7917} & \textbf{20.9722} & 10.6898 & 1.4213 \\ 

\rowcolor[gray]{.8}
\multicolumn{11}{c}{\textbf{T=5}} \\ 

COHERENT
& 0.2500 & 0.5648 & 28.1389 & 18.2778 & --
& 0.5000 & 0.6898 & 21.0278 & \textbf{9.8634} & -- \\ 

CoELA
& 0.3889 & 0.6481 & 26.0278 & 20.1574 & 4.4306
& 0.2222 & 0.4074 & 26.1389 & 14.9491 & \textbf{2.6644} \\

\sysnameo
& 0.7600 & \textbf{0.8933} & 24.4000 & 19.2867 & 6.3867
& 0.5750 & 0.7292 & \textbf{19.6250} & 12.9896 & 3.9542 \\

\sysnameq
& \textbf{0.8000} & 0.8788 & \textbf{20.1455} & \textbf{13.8796} & \textbf{2.1955}
& \textbf{0.6909} & \textbf{0.8364} & 23.4909 & 15.6333 & 2.8711 \\ 

\rowcolor[gray]{.8}
\multicolumn{11}{c}{\textbf{T=13}} \\ 

COHERENT
& 0.0000 & 0.3843 & 30.0000 & 19.3380 & --
& 0.1667 & 0.2500 & 26.5278 & \textbf{5.4074} & -- \\

CoELA
& 0.1389 & 0.5556 & 28.7500 & 20.8102 & 5.3287
& 0.1111 & 0.2037 & 28.1667 & 5.8079 & \textbf{0.9259} \\

\sysnameo
& 0.1389 & 0.4028 & 28.5000 & \textbf{12.6991} & 3.4537
& 0.5833 & 0.7731 & 21.3611 & 13.0903 & 2.7708 \\

\sysnameq
& \textbf{0.6389} & \textbf{0.8565} & \textbf{24.4722} & 16.1991 & \textbf{3.0370} 
& \textbf{0.7143} & \textbf{0.8524} & \textbf{18.2857} & 11.4762 & 1.5595 \\ 

\bottomrule

\end{tabular}} 
\end{table*}

%% file: table/llm-mcts.tex
\begin{table*}[htbp]
\centering
\scriptsize
\setlength{\tabcolsep}{3pt}
\renewcommand{\arraystretch}{1.2}
\caption{Evaluation Results on Sort Solids with Varying Robot Types and Task Difficulties ($T=0$)}
\label{tab:task3_sort_solids}
\resizebox{\linewidth}{!}{
\begin{tabular}{>{\centering\arraybackslash}p{3.0cm} *{15}{>{\centering\arraybackslash}p{0.8cm}}}
\toprule

\multirow{2}{*}{\textbf{Method}}
& \multicolumn{5}{c}{\textbf{Easy}}
& \multicolumn{5}{c}{\textbf{Medium}}
& \multicolumn{5}{c}{\textbf{Hard}} \\
\cmidrule(lr){2-6} \cmidrule(lr){7-11} \cmidrule(lr){12-16}
& SUCC$\uparrow$ & PS$\uparrow$ & TS$\downarrow$ & AS$\downarrow$ & CC$\downarrow$
& SUCC$\uparrow$ & PS$\uparrow$ & TS$\downarrow$ & AS$\downarrow$ & CC$\downarrow$
& SUCC$\uparrow$ & PS$\uparrow$ & TS$\downarrow$ & AS$\downarrow$ & CC$\downarrow$ \\
\midrule

\rowcolor[gray]{.85} \multicolumn{16}{c}{\textbf{2 Types of Robot}} \\
MCTS-NoINS & 0.3333 & 0.3333 & 15.9444 & - & - & 0.0556 & 0.2269 & 29.6111 & - & - & 0.0000 & 0.0556 & 50.0000 & - & - \\
MCTS & 0.5000 & 0.5000 & 13.0000 & - & - & 0.1667 & 0.2639 & 28.5000 & - & - & 0.5000 & 0.5556 & 31.8333 & - & - \\
LLM-MCTS & \textbf{1.0000} & \textbf{1.0000} & 5.6667 & - & - & \textbf{1.0000} & \textbf{1.0000} & 10.1667 & - & - & 0.8333 & 0.9444 & 19.0000 & - & - \\
CoELA & \textbf{1.0000} & \textbf{1.0000} & \textbf{3.1667} & \textbf{3.1667} & 0.1111 & 0.8333 & 0.9583 & 10.6111 & 9.6667 & 0.6667 & 0.9444 & 0.9815 & 14.6667 & 12.1667 & 1.3056 \\
\sysnameo & \textbf{1.0000} & \textbf{1.0000} & 4.0556 & 3.8056 & 0.4167 & 0.8333 & 0.9352 & 13.1111 & 11.7222 & 2.5278 & 0.9444 & 0.9722 & 9.6111 & 8.4722 & \textbf{1.3056} \\
\sysnameq & \textbf{1.0000} & \textbf{1.0000} & 3.2222 & 3.1945 & \textbf{0.0555} & \textbf{1.0000} & \textbf{1.0000} & \textbf{6.6111} & \textbf{6.6111} & \textbf{0.0556} & \textbf{0.9745} & \textbf{0.9859} & \textbf{7.6667} & \textbf{6.4825} & \textbf{0.8321} \\

\rowcolor[gray]{.85} \multicolumn{16}{c}{\textbf{3 Types of Robot}} \\
MCTS-NoINS & 0.3333 & 0.3704 & 16.9259 & - & - & 0.0000 & 0.1019 & 30.0000 & - & - & 0.0370 & 0.1543 & 48.8148 & - & - \\
MCTS & 0.6667 & 0.6667 & 13.2222 & - & - & 0.1111 & 0.2130 & 28.5556 & - & - & 0.3333 & 0.3981 & 44.2222 & - & - \\
LLM-MCTS & \textbf{1.0000} & \textbf{1.0000} & 5.1111 & - & - & 0.4444 & 0.6389 & 21.5556 & - & - & 0.5556 & 0.8148 & 40.1111 & - & - \\
CoELA & \textbf{1.0000} & \textbf{1.0000} & \textbf{3.3333} & \textbf{3.2593} & 0.2716 & \textbf{0.9630} & 0.9877 & 9.2593 & 8.6790 & 1.0123 & 0.8889 & 0.9630 & 16.8519 & 15.1235 & 2.0864 \\
\sysnameo & \textbf{1.0000} & \textbf{1.0000} & 5.0000 & 4.5556 & 0.8271 & \textbf{0.9630} & \textbf{0.9907} & 11.9630 & 11.0123 & 2.3704 & 0.9200 & 0.9815 & 16.4074 & 14.1852 & 4.0247 \\
\sysnameq & \textbf{1.0000} & \textbf{1.0000} & 4.5556 & 4.1481 & \textbf{0.1950} & 0.9615 & 0.9877 & \textbf{8.5185} & \textbf{8.0062} & \textbf{0.2778} & \textbf{0.9630} & \textbf{0.9907} & \textbf{14.9629} & \textbf{11.8642} & \textbf{1.4444} \\

\rowcolor[gray]{.85} \multicolumn{16}{c}{\textbf{4 Types of Robot}} \\
MCTS-NoINS & 0.0000 & 0.0556 & 20.0000 & - & - & 0.0000 & 0.0370 & 30.0000 & - & - & 0.0000 & 0.0815 & 50.0000 & - & - \\
MCTS & 0.0000 & 0.0000 & 20.0000 & - & - & 0.0000 & 0.0000 & 30.0000 & - & - & 0.0000 & 0.0667 & 50.0000 & - & - \\
LLM-MCTS & 0.6667 & 0.6667 & 13.0000 & - & - & 0.3333 & 0.6111 & 23.6667 & - & - & 0.0000 & 0.3722 & 50.0000 & - & - \\
CoELA & \textbf{1.0000} & \textbf{1.0000} & 4.5556 & 4.4167 & 0.5278 & 0.8889 & 0.9630 & \textbf{11.6667} & 10.9444 & 1.4444 & 0.7778 & 0.9407 & 21.3333 & 18.9722 & \textbf{2.6111} \\
\sysnameo & \textbf{1.0000} & \textbf{1.0000} & 4.8889 & 4.7778 & 0.5833 & \textbf{1.0000} & \textbf{1.0000} & 16.2222 & 14.2500 & 3.2778 & \textbf{1.0000} & \textbf{1.0000} & \textbf{20.2222} & \textbf{18.3333} & 4.4722 \\
\sysnameq & \textbf{1.0000} & \textbf{1.0000} & \textbf{4.0634} & \textbf{3.8626} & \textbf{0.2584} & \textbf{1.0000} & \textbf{1.0000} & 12.1111 & \textbf{10.4444} & \textbf{1.3889} & 0.9444 & 0.9630 & 28.3333 & 23.6667 & 3.4167 \\

\bottomrule
\end{tabular}
}
\end{table*}

%% file: table/conhernet_case.tex
\begin{table*}[h]
\centering
\footnotesize
\setlength{\tabcolsep}{6pt}
\renewcommand{\arraystretch}{1.18}

\caption{Representative failure case of the centralized coordinator.}
\label{tab:centralized_failure_case}

\begin{tabularx}{\textwidth}{
>{\centering\arraybackslash}p{2.0cm}
>{\RaggedRight\arraybackslash}p{7.0cm}
>{\RaggedRight\arraybackslash}X
}
\toprule
\rowcolor{gray!15}
\textbf{Stage} 
& \textbf{Key Observation} 
& \textbf{Failure Pattern} \\
\midrule

\textbf{Task Setup}
&
Team: \texttt{Ma-MoMa-UAV}. 
Targets: \texttt{book\_0}, \texttt{fork\_0}, and \texttt{soap}.
&
The coordinator needs to track only the target object's progress and exclude irrelevant objects.
\\

\midrule

\textbf{Step 9}
&
The centralized progress state contains \texttt{book\_0} and the non-target object \texttt{toothpaste\_0}.
&
A non-target object is incorrectly accepted as completed task progress.
\\

\midrule

\textbf{Steps 19--26}
&
The progress state becomes \texttt{[book\_0, toothpaste\_0, soap]}. 
The coordinator repeatedly outputs \texttt{wait} or claims that all required items are completed.
&
The missing target \texttt{fork\_0} is overlooked, causing global idling instead of correction or reallocation.
\\

\midrule

\textbf{Step 27}
&
The coordinator belatedly asks Bob to execute \texttt{pick(fork\_0)}.
Bob reports that \texttt{fork\_0} is not available in its local scene graph.
&
The recovery is delayed and not grounded in the robot's local observability.
\\

\midrule

\textbf{Steps 28--29}
&
Alice and Lucy are sent to additional exploratory locations, while Bob waits. 
Alice finds no object on \texttt{table\_1}; Lucy repeatedly reports obstacle-related failures.
&
The centralized policy reacts late with weak coordination, and \texttt{fork\_0} remains missing.
\\

\midrule

\textbf{Outcome}
&
\texttt{FinishedStep=30}, \texttt{PartSuccess=0.333}. 
Final State: \texttt{book\_0}, \texttt{soap}, and non-target \texttt{toothpaste\_0}; missing target: \texttt{fork\_0}.
&
The failure is caused by centralized state-tracking error and insufficient feedback-driven recovery.
\\

\bottomrule
\end{tabularx}

\begin{minipage}{0.96\textwidth}
\footnotesize
\emph{Note.} The true target objects are \texttt{book\_0}, \texttt{fork\_0}, and \texttt{soap}. 
This case shows that an incorrect centralized progress state can become a single point of failure, leading to stale global belief, delayed recovery, and eventual timeout.
\end{minipage}

\end{table*}

%% file: table/token_comparison.tex
\begin{table*}[t]
\centering
\scriptsize
\setlength{\tabcolsep}{4pt}
\renewcommand{\arraystretch}{1.0}
\caption{Comparison of average token consumption and API call counts between COHERENT and DynaHMRC-GPT4o
}
\label{tab:token_comparison}
\resizebox{\textwidth}{!}{
\begin{tabular}{
>{\centering\arraybackslash}p{2.4cm} 
>{\centering\arraybackslash}p{0.9cm} 
cc cc cc cc cc cc c}
\toprule
\textbf{Method} & \textbf{Task} 
& \multicolumn{2}{c}{\textbf{SD}} 
& \multicolumn{2}{c}{\textbf{TALB}}
& \multicolumn{2}{c}{\textbf{LE}}
& \multicolumn{2}{c}{\textbf{HMRC}}
& \multicolumn{2}{c}{\textbf{GD}}
& \multicolumn{2}{c}{\textbf{LP}} & \textbf{Total Tokens} \\
\cmidrule(lr){3-4} \cmidrule(lr){5-6} \cmidrule(lr){7-8} \cmidrule(lr){9-10} \cmidrule(lr){11-12} \cmidrule(lr){13-14}
& 
& \textbf{Calls} & \textbf{Tokens} 
& \textbf{Calls} & \textbf{Tokens} 
& \textbf{Calls} & \textbf{Tokens} 
& \textbf{Calls} & \textbf{Tokens} 
& \textbf{Calls} & \textbf{Tokens} 
& \textbf{Calls} & \textbf{Tokens} \\
\midrule

\multirow{5}{*}{\sysnameo}
& Static & 2.95 & 670.90 & 2.95 & 1687.08 & 3.67 & 2217.93 & 53.42 & 3532.95 & 3.25 & 3015.17 & 1.17 & 1273.50 & 205529.6 \\
& CTO   & 2.77 & 659.59 & 2.77 & 1607.12 & 3.34 & 2067.84 & 56.27 & 3505.75 & 3.58 & 3130.14 & 1.36 & 1283.46 & 210517.2 \\
& IRZ   & 3.29 & 667.14 & 3.29 & 1783.55 & 4.02 & 2416.39 & 60.60 & 3717.89 & 3.38 & 3222.91 & 1.07 & 1411.23 & 241190.9 \\
& ANC   & 2.53 & 971.45 & 2.53 & 2336.88 & 3.17 & 2493.01 & 68.41 & 3853.70 & 4.76 & 3175.95 & 1.37 & 1487.05 & 280459.6 \\
& REC   & 3.46 & 679.66 & 3.46 & 1823.02 & 4.03 & 2463.01 & 55.47 & 3551.53 & 3.01 & 2917.18 & 1.26 & 1197.33 & 216441.7 \\

\midrule

\multirow{5}{*}{COHERENT}
& Static & -- & -- & -- & -- & -- & -- & 17.63 & 6654.57 & -- & -- & -- & -- & 117345.1 \\
& CTO   & -- & -- & -- & -- & -- & -- & 17.56 & 7117.76 & -- & -- & -- & -- & 125004.0 \\
& IRZ   & -- & -- & -- & -- & -- & -- & 17.88 & 7572.02 & -- & -- & -- & -- & 135367.7 \\
& ANC   & -- & -- & -- & -- & -- & -- & 18.21 & 7449.58 & -- & -- & -- & -- & 135637.9 \\
& REC   & -- & -- & -- & -- & -- & -- & 19.18 & 6732.11 & -- & -- & -- & -- & 129122.9 \\

\bottomrule
\end{tabular}}
\end{table*}

%% file: table/token_robot_composition.tex
\begin{table*}[htbp]
\centering
\scriptsize
\setlength{\tabcolsep}{4pt}
\renewcommand{\arraystretch}{1.05}
\caption{Average Token Consumption Comparison Between \sysnameo and COHERENT across different tasks and robot compositions.}
\label{tab:token_robot_composition}
\resizebox{\textwidth}{!}{
\begin{tabular}{
>{\centering\arraybackslash}p{0.6cm}
>{\centering\arraybackslash}p{5.1cm}
ccccccc
c
}
\toprule
\textbf{Task} & \textbf{Robot Composition}
& \multicolumn{7}{c}{\textbf{\sysnameo}}
& \textbf{COHERENT} \\
\cmidrule(lr){3-9}
&
& \textbf{SD}
& \textbf{TALB}
& \textbf{LE}
& \textbf{HMRC}
& \textbf{GD}
& \textbf{LP}
& \textbf{Total}
& \textbf{Total} \\
\midrule

\multirow{6}{*}{Static}
& Ma-MoMa        & 669.33 & 1364.02 & 1600.67 & 3675.94 & 2705.32 & 1068.28 & 152967.9 & 96031.6 \\
& Ma-UAV         & 651.44 & 1409.19 & 1688.20 & 2711.61 & 2783.03 & 1063.21 & 117134.5 & 86303.3 \\
& Ma-MoMa-Mo     & 662.86 & 1681.98 & 2263.30 & 3745.79 & 3036.69 & 1296.48 & 234230.2 & 159865.0 \\
& Ma-MoMa-UAV    & 689.28 & 1728.33 & 2270.97 & 3802.63 & 3066.67 & 1362.50 & 240516.9 & 126566.7 \\
& Ma-Mo-UAV      & 657.74 & 1691.90 & 2239.49 & 3180.06 & 3275.23 & 1418.26 & 184653.3 & 152929.0 \\
& Ma-MoMa-Mo-UAV & 685.21 & 2023.07 & 2834.13 & 3862.17 & 3153.99 & 1368.29 & 299108.9 & 110383.6 \\

\midrule

\multirow{6}{*}{CTO}
& Ma-MoMa        & 670.10 & 1372.00 & 1607.70 & 3565.32 & 2956.42 & 1070.93 & 108491.3 & 78998.9 \\
& Ma-UAV         & 636.91 & 1370.30 & 1635.11 & 2774.89 & 2902.90 & 1074.11 & 156160.9 & 84887.1 \\
& Ma-MoMa-Mo     & 666.84 & 1658.73 & 2138.17 & 4016.30 & 3186.24 & 1324.75 & 311026.2 & 157610.6 \\
& Ma-MoMa-UAV    & 686.13 & 1729.00 & 2234.11 & 3759.57 & 3163.11 & 1304.95 & 215452.2 & 154555.1 \\
& Ma-Mo-UAV      & 648.29 & 1673.76 & 2185.43 & 3308.28 & 3197.66 & 1327.97 & 246380.1 & 161708.6 \\
& Ma-MoMa-Mo-UAV & 659.85 & 1949.43 & 2808.45 & 3951.23 & 3402.44 & 1599.75 & 328270.4 & 159953.5 \\

\midrule

\multirow{4}{*}{IRZ}
& Ma-MoMa-Mo     & 669.58 & 1624.92 & 2021.64 & 4056.62 & 3148.25 & 1322.77 & 315304.4 & 155016.0 \\
& Ma-MoMa-UAV    & 675.90 & 1729.59 & 2352.68 & 3796.47 & 3251.67 & 1324.17 & 223450.0 & 117690.8 \\
& Ma-Mo-UAV      & 649.79 & 1704.64 & 2301.27 & 3195.69 & 3270.77 & 1335.89 & 190138.5 & 141383.1 \\
& Ma-MoMa-Mo-UAV & 673.38 & 2015.42 & 2864.93 & 3856.91 & 3239.22 & 1632.55 & 281442.6 & 149193.8 \\

\midrule

\multirow{4}{*}{ANC}
& Ma-MoMa\boldmath$\mathbf{\;\rightarrow\;}$Ma-MoMa-UAV        & 1041.23 & 2272.47 & 2195.92 & 3769.22 & 3049.70 & 1367.32 & 193148.7 & 109408.7 \\
& Ma-UAV\boldmath$\mathbf{\;\rightarrow\;}$Ma-MoMa-UAV         & 1042.10 & 2320.50 & 2248.63 & 3887.07 & 3077.61 & 1354.90 & 292076.0 & 125527.7 \\
& Ma-MoMa-Mo\boldmath$\mathbf{\;\rightarrow\;}$Ma-MoMa-Mo-UAV     & 909.30 & 2360.60 & 2695.58 & 3894.81 & 3315.44 & 1617.44 & 306627.5 & 171243.8 \\
& Ma-Mo-UAV\boldmath$\mathbf{\;\rightarrow\;}$Ma-MoMa-Mo-UAV      & 907.07 & 2378.68 & 2769.84 & 3840.89 & 3246.20 & 1604.16 & 353722.3 & 189261.1 \\

\midrule

\multirow{4}{*}{REC}
& Ma-MoMa-UAV\boldmath$\mathbf{\;\rightarrow\;}$Ma-MoMa      & 691.95 & 1714.00 & 2204.24 & 3750.68 & 2747.22 & 1063.92 & 197580.5 & 115790.1 \\
& Ma-MoMa-UAV\boldmath$\mathbf{\;\rightarrow\;}$Ma-UAV     & 679.29 & 1668.67 & 2135.03 & 3194.26 & 2824.70 & 1094.57 & 156383.4 & 99027.4 \\
& Ma-MoMa-Mo-UAV\boldmath$\mathbf{\;\rightarrow\;}$Ma-MoMa-Mo   & 679.02 & 1994.05 & 2806.99 & 3947.61 & 3061.45 & 1344.10 & 326637.9 & 164289.3 \\
& Ma-MoMa-Mo-UAV\boldmath$\mathbf{\;\rightarrow\;}$Ma-Mo-UAV  & 668.02 & 1962.27 & 2809.52 & 3400.59 & 3054.30 & 1310.18 & 224959.2 & 136688.4 \\

\bottomrule
\end{tabular}}
\end{table*}

%% file: table/latency_scaling.tex
\begin{table}[htbp]
\centering
\small
\caption{Scalability Evaluation with Latency Breakdown for Different Robot Team Sizes on NVIDIA Jetson AGX Orin}
\label{tab:team-scalability-orin-breakdown}
\resizebox{\columnwidth}{!}{
\begin{tabular}{lccc}
\hline
Metric & 2 robots & 3 robots & 4 robots \\
\hline
\multicolumn{4}{l}{\textit{Workload}} \\
Input tokens & 4003.70 $\pm$ 1441.41 & 3419.20 $\pm$ 2394.78 & 4868.00 $\pm$ 1663.23 \\
Output tokens & 285.90 $\pm$ 118.72 & 234.20 $\pm$ 95.68 & 213.30 $\pm$ 43.43 \\
Total tokens & 4289.60 $\pm$ 1359.15 & 3653.40 $\pm$ 2367.71 & 5081.30 $\pm$ 1640.92 \\
\hline
\multicolumn{4}{l}{\textit{Latency and throughput}} \\
Preprocessing time (ms) & 2749.56 $\pm$ 1011.43 & 2394.65 $\pm$ 1706.93 & 3398.57 $\pm$ 1191.53 \\
Generation time (ms) & 10679.55 $\pm$ 4141.90 & 8899.68 $\pm$ 3322.53 & 8215.88 $\pm$ 1654.69 \\
Total time (ms) & 13429.12 $\pm$ 3426.82 & 11294.33 $\pm$ 3305.58 & 11614.44 $\pm$ 1608.00 \\
Preprocessing share (\%) & 22.67 $\pm$ 11.87 & 21.60 $\pm$ 13.57 & 29.25 $\pm$ 10.12 \\
Generation speed (t/s) & 26.52 $\pm$ 0.76 & 26.07 $\pm$ 1.26 & 25.95 $\pm$ 1.00 \\
\hline
\end{tabular}}
\end{table}

%% file: table/fail_scaling.tex
\begin{table}[htbp]
\centering
\scriptsize
\setlength{\tabcolsep}{4pt}
\renewcommand{\arraystretch}{1.05}
\caption{Statistics of Coordination and Alignment Failures}
\label{tab:fail_scaling}
\resizebox{\columnwidth}{!}{
\begin{tabular}{
>{\centering\arraybackslash}p{0.6cm}
>{\centering\arraybackslash}p{4.8cm}
ccccccc
c
}
\toprule
\textbf{Task} & \textbf{Robot Composition}
& \multicolumn{2}{c}{\textbf{\sysnameo}}
& \multicolumn{2}{c}{\textbf{COHERENT}} \\
\cmidrule(lr){3-4} \cmidrule(lr){5-6}
&
& \textbf{Redun.}
& \textbf{Conf.}
& \textbf{Redun.}
& \textbf{Conf.} \\
\midrule

\multirow{6}{*}{Static}
& Ma-MoMa        & 0.8667 & 0.0000 & 2.3704 & 0.0000 \\
& Ma-UAV         & 1.8889 & 0.0370 & 5.9630 & 0.0000 \\
& Ma-MoMa-Mo     & 2.5667 & 0.3667 & 11.8148 & 0.2222 \\
& Ma-MoMa-UAV    & 1.5333 & 0.6333 & 2.9630 & 0.8148 \\
& Ma-Mo-UAV      & 3.5000 & 0.4667 & 14.2222 & 1.1852 \\
& Ma-MoMa-Mo-UAV & 5.1458 & 1.4375 & 6.3704 & 0.9630 \\

\midrule

\multirow{6}{*}{CTO}
& Ma-MoMa        & 0.4667 & 0.0667 & 1.5185 & 0.0741 \\
& Ma-UAV         & 3.8182 & 0.0000 & 3.3704 & 0.0000 \\
& Ma-MoMa-Mo     & 5.2667 & 0.4000 & 8.5556 & 0.1111 \\
& Ma-MoMa-UAV    & 2.6000 & 0.4000 & 5.9630 & 0.5556 \\
& Ma-Mo-UAV      & 5.9333 & 0.6667 & 8.8148 & 1.0741 \\
& Ma-MoMa-Mo-UAV & 6.8000 & 1.8667 & 6.4074 & 0.9259 \\

\midrule

\multirow{4}{*}{IRZ}
& Ma-MoMa-Mo     & 6.6667 & 0.2500 & 9.5185 & 0.1852 \\
& Ma-MoMa-UAV    & 2.1538 & 0.4615 & 4.0000 & 0.7407 \\
& Ma-Mo-UAV      & 3.9286 & 0.5000 & 10.7778 & 0.7407 \\
& Ma-MoMa-Mo-UAV & 4.8750 & 1.0000 & 7.6296 & 0.9259 \\

\midrule

\multirow{4}{*}{ANC}
& Ma-MoMa\boldmath$\mathbf{\;\rightarrow\;}$Ma-MoMa-UAV        & 1.4667 & 0.2000 & 3.4074 & 0.4444 \\
& Ma-UAV\boldmath$\mathbf{\;\rightarrow\;}$Ma-MoMa-UAV         & 2.1429 & 0.2857 & 11.0000 & 1.2222 \\
& Ma-MoMa-Mo\boldmath$\mathbf{\;\rightarrow\;}$Ma-MoMa-Mo-UAV  & 4.9000 & 0.8500 & 15.4074 & 0.8889 \\
& Ma-Mo-UAV\boldmath$\mathbf{\;\rightarrow\;}$Ma-MoMa-Mo-UAV   & 6.6000 & 0.8500 & 16.9259 & 1.5926 \\

\midrule

\multirow{4}{*}{REC}
& Ma-MoMa-UAV\boldmath$\mathbf{\;\rightarrow\;}$Ma-MoMa         & 1.6364 & 0.8182 & 2.0741 & 0.3333 \\
& Ma-MoMa-UAV\boldmath$\mathbf{\;\rightarrow\;}$Ma-UAV          & 1.4667 & 0.2333 & 5.1481 & 0.4074 \\
& Ma-MoMa-Mo-UAV\boldmath$\mathbf{\;\rightarrow\;}$Ma-MoMa-Mo   & 5.7500 & 0.7083 & 8.2593 & 0.6296 \\
& Ma-MoMa-Mo-UAV\boldmath$\mathbf{\;\rightarrow\;}$Ma-Mo-UAV    & 5.3333 & 1.1429 & 11.6667 & 1.1852 \\

\bottomrule
\end{tabular}}
\end{table}